\documentclass[10pt,twocolumn,letterpaper]{article}

\usepackage[pagenumbers]{wacv}

\usepackage{graphicx}
\usepackage{amsmath}
\usepackage{amssymb}
\usepackage{booktabs}

\usepackage{xcolor}
\usepackage{listings}
\usepackage{pgf}
\usepackage{subcaption}
\usepackage{enumitem}
\usepackage{amsmath,amsfonts,bm}

\newcommand{\bx}{\mathbf{x}}
\newcommand{\by}{\mathbf{y}}

\usepackage{color, colortbl}
\definecolor{Gray}{gray}{0.9}
\usepackage{tikz}
\usepackage[final]{changes}
\usepackage[pagebackref,breaklinks,colorlinks]{hyperref}

\usepackage[capitalize]{cleveref}
\crefname{section}{Sec.}{Secs.}
\Crefname{section}{Section}{Sections}
\Crefname{table}{Table}{Tables}
\crefname{table}{Tab.}{Tabs.}

\newcommand{\ours}{LFG-Diffusion}

\def\wacvPaperID{2284} %
\def\confName{WACV}
\def\confYear{2024}

\begin{document}

\title{Latent Feature-Guided Diffusion Models for Shadow Removal}

\author{Kangfu Mei\thanks{This work was done during an internship at Adobe Research.}$^{\,\,\,\,1,2}$,  Luis Figueroa$^{2}$, Zhe Lin$^{2}$, Zhihong Ding$^{2}$,\\ Scott Cohen$^{2}$, Vishal M. Patel\thanks{Vishal M. Patel was supported by NSF CAREER award 2045489.}$^{\,\,\,\,1}$\\
$^{1}$ Johns Hopkins University, $^{2}$ Adobe Research \\
\url{https://kfmei.com/shadow-diffusion/}
}

\maketitle

\begin{abstract}
Recovering textures under shadows has remained a challenging problem due to the difficulty of inferring shadow-free scenes from shadow images. In this paper, we propose the use of diffusion models as they offer a promising approach to gradually refine the details of shadow regions during the diffusion process. Our method improves this process by conditioning on a learned latent feature space that inherits the characteristics of shadow-free images, thus avoiding the limitation of conventional methods that condition on degraded images only. Additionally, we propose to alleviate potential local optima during training by fusing noise features with the diffusion network. We demonstrate the effectiveness of our approach which outperforms the previous best method by 13\% in terms of RMSE on the AISTD dataset. Further, we explore instance-level shadow removal, where our model outperforms the previous best method by 82\% in terms of RMSE on the DESOBA dataset.
\end{abstract}

\section{Introduction}
Images captured in natural illumination often contain shadows caused by objects blocking the light from the illumination source. Shadows can degrade the performance of many computer vision algorithms, such as detection, segmentation, and recognition~\cite{ma2008shadow, zhang2018improving}. Furthermore, removing shadows is essential for photo-editing applications such as distractor removal~\cite{zhang2021no} and relighting~\cite{hou2022face}; which may rely on instance-level shadow removal. Therefore, it is critical to develop methods that can automatically remove shadows from captured images as works explored in literature.

\begin{figure}
    \centering
    \begin{subfigure}[b]{.32\linewidth}
    \includegraphics[width=\linewidth]{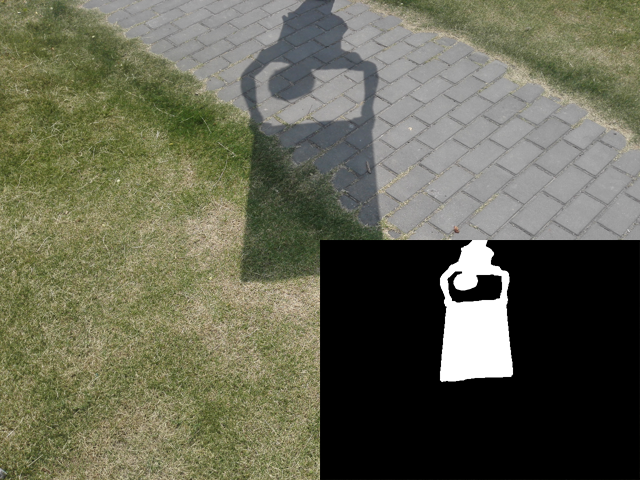}
    \caption*{Shadow \& Mask}
    \end{subfigure}
    \begin{subfigure}[b]{.32\linewidth}
    \includegraphics[width=\linewidth]{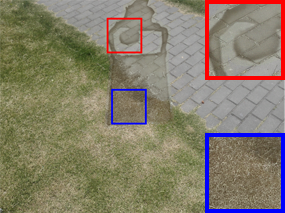}
    \caption*{Baseline Result}
    \end{subfigure}
    \begin{subfigure}[b]{.32\linewidth}
    \includegraphics[width=\linewidth]{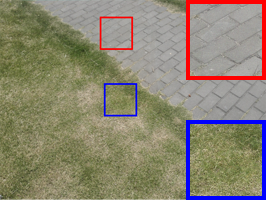}
    \caption*{Our Result}
    \end{subfigure}
    \hfill
    \caption{Our baseline method, which conditions diffusion models solely on shadow and mask images, produces incorrect results such as color mixing in highlight areas. In contrast, our proposed method generates results with consistent and reasonable colors that match the surrounding area.}
    \label{fig:intensity}
\end{figure}

Recently, diffusion models~\cite{sohl2015deep} with hierarchical denoising autoencoders~\cite{ho_denoising_2020} have shown to achieve impressive synthesis performance in terms of sample quality and diversity.
The conditional generation ability further allows for iterative refinement and fine-grained control according to certain conditions.
Motivated by the success of diffusion-based image restoration models~\cite{saharia2022image, rombach2022high}, we adapt diffusion models for the task of shadow removal by conditioning on the input shadow image and corresponding shadow mask as a baseline approach to generate shadow-free images.
However, preserving and generating high-fidelity textures and colors in the shadow region after removal is non-trivial.
The baseline model appears to favor borrowing textures from the surrounding non-shadow areas rather than focusing on restoring the original details underneath the shadow, which results in incorrect color mixtures and loss of detail in the shadow region.
In Fig.~\ref{fig:intensity}, we show one of the representative issues of image-mask conditioning, \ie, the model synthesizes results containing an incorrect color mixture.

Intuitively, the intensity drop in shadow regions means that diffusion models are typically guided more strongly by the surrounding non-shadow areas. However, this guidance can harm the fidelity of the result if the texture and color under the shadow differs significantly from the surrounding areas. In addition, the multi-head attention module~\cite{vaswani2017attention} used in diffusion models can exacerbate this issue by extracting global information. This motivates us to consider guiding the conditioned diffusion models with an additional latent feature space that captures external perceptual shadow-free information as the shadow removal priors.

\begin{figure*}[ht]
\centering
\setlength\lineskip{1pt}
\setlength{\fboxsep}{0pt}%
\setlength{\fboxrule}{.5pt}%
\begin{subfigure}[t]{\linewidth}
\centering
\fbox{\includegraphics[width=.195\linewidth, height=.11\linewidth]{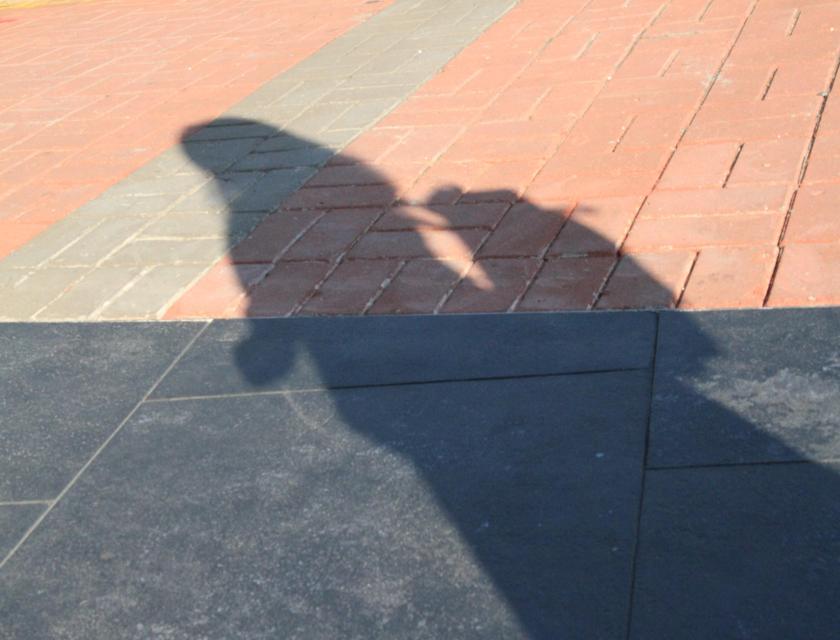}}%
\fbox{\includegraphics[width=.195\linewidth, height=.11\linewidth]{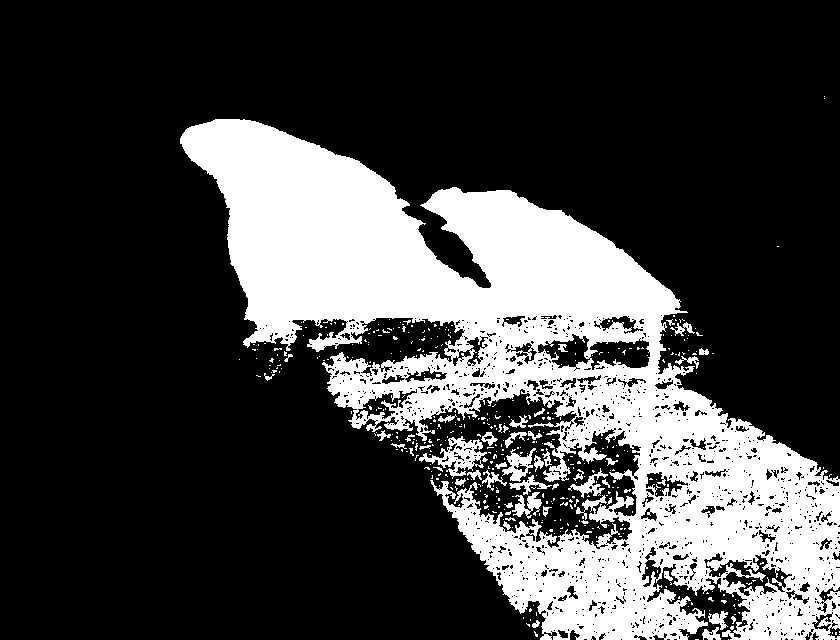}}%
\fbox{\includegraphics[width=.195\linewidth, height=.11\linewidth]{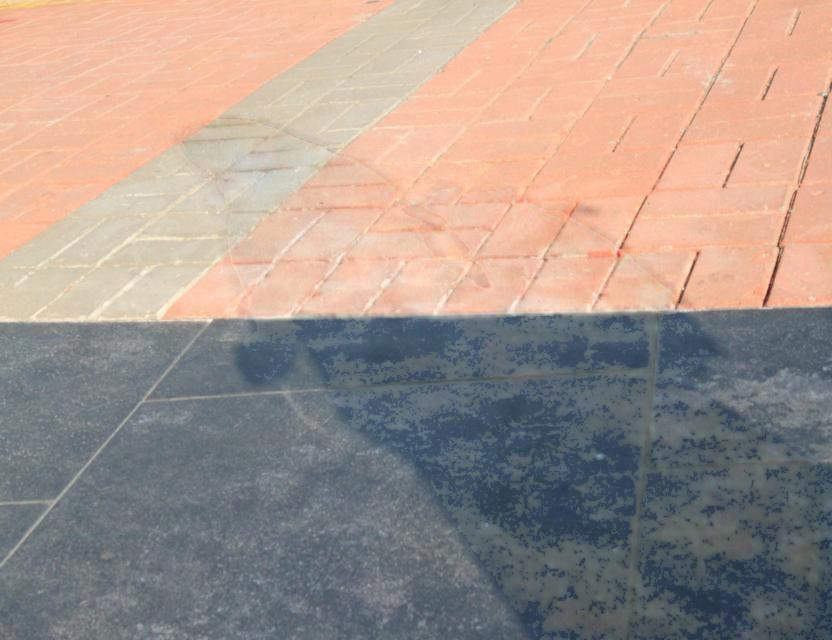}}%
\fbox{\includegraphics[width=.195\linewidth, height=.11\linewidth]{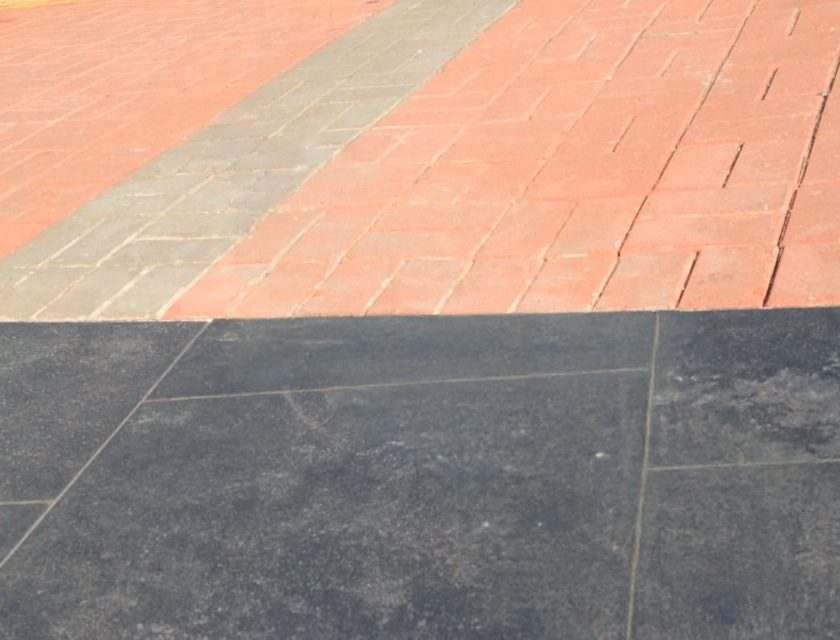}}%
\fbox{\includegraphics[width=.195\linewidth, height=.11\linewidth]{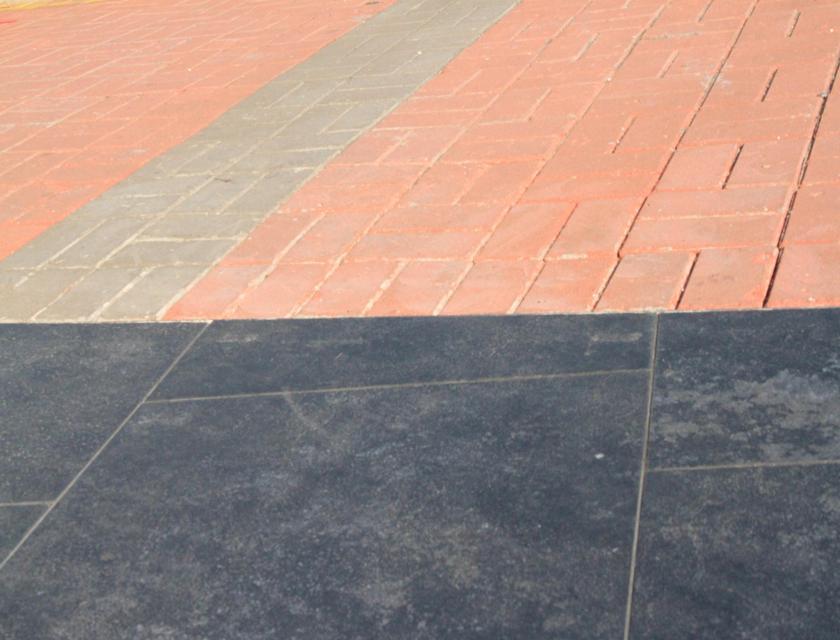}}%
\hfill
\end{subfigure}
\begin{subfigure}[t]{\linewidth}
\centering
\fbox{\includegraphics[width=.195\linewidth, height=.11\linewidth]{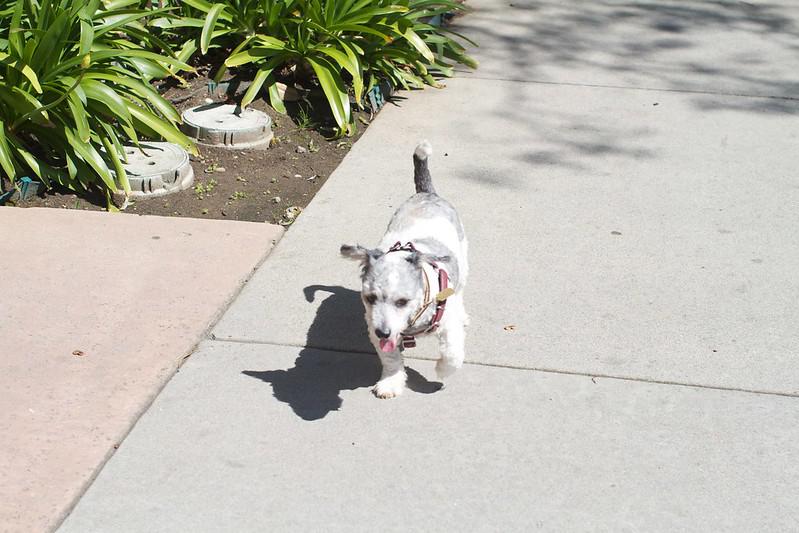}}%
\fbox{\includegraphics[width=.195\linewidth, height=.11\linewidth]{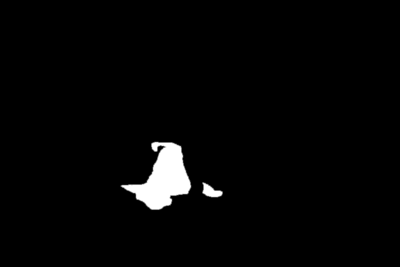}}%
\fbox{\includegraphics[width=.195\linewidth, height=.11\linewidth]{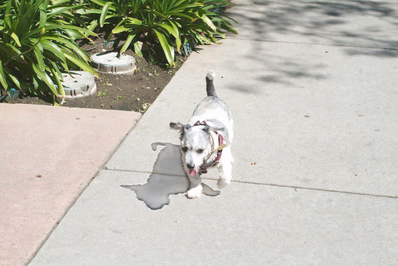}}%
\fbox{\includegraphics[width=.195\linewidth, height=.11\linewidth]{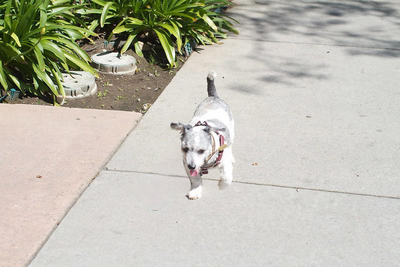}}%
\fbox{\includegraphics[width=.195\linewidth, height=.11\linewidth]{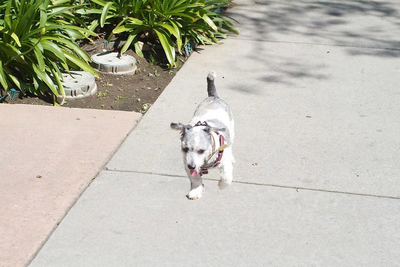}}%
\end{subfigure}
\begin{subfigure}[t]{\linewidth}
\centering
\fbox{\includegraphics[width=.195\linewidth, height=.11\linewidth]{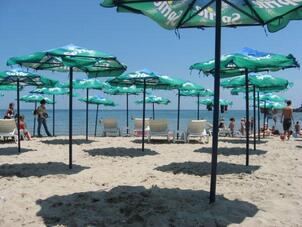}}%
\fbox{\includegraphics[width=.195\linewidth, height=.11\linewidth]{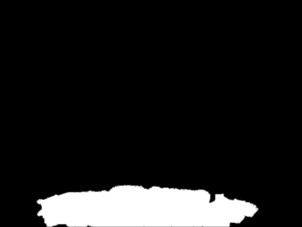}}%
\fbox{\includegraphics[width=.195\linewidth, height=.11\linewidth]{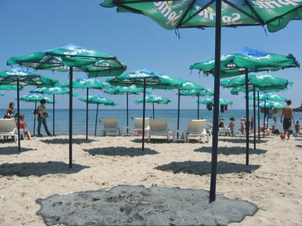}}%
\fbox{\includegraphics[width=.195\linewidth, height=.11\linewidth]{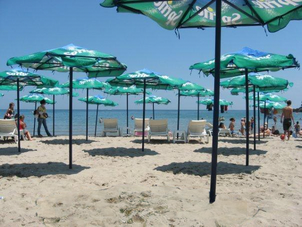}}%
\fbox{\includegraphics[width=.195\linewidth, height=.11\linewidth]{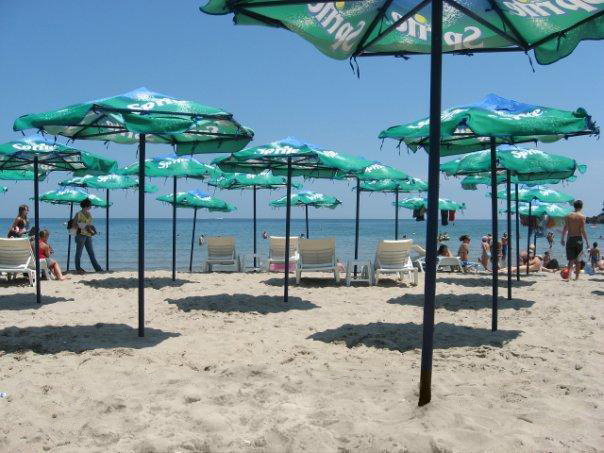}}%
\end{subfigure}
\caption{Given a shadow mask, our method effectively removes shadows and recovers the underlying details for shadows at the general level (top two rows) or instance level (bottom two rows). From left to right, we show the input image, shadow mask, SG-ShadowNet~\cite{wan2022} result, our method result, and shadow-free images for comparisons.}
\end{figure*}

Our proposed method differs from latent diffusion models (LDMs)~\cite{rombach2022high} in that we incorporate a learnable feature encoder to discover a latent feature space. To optimize the latent feature encoder, we minimize the difference between the feature space of shadow images and that of shadow-free images, using it as the loss function. Through experimentation, we have found that optimizing the encoder together with the diffusion models leads to a compact and perceptual latent feature space. Additionally, we demonstrate that pretraining the diffusion model on shadow-free images simplifies the optimization process and is crucial for achieving high-fidelity synthesis. By guiding the diffusion models on the latent feature space, instead of just conditioning on the shadow image and mask, we observe significant improvements in shadow removal capability.

In addition to the proposed latent feature space guidance, we propose an improved diffusion network that addresses the issue of \emph{posterior collapse}~\cite{zhao2017infovae, he2019lagging, dieng2019avoiding}, which refers to the local optima of diffusion models.
We identify the local optimum as the degrading effect of the noise variable and introduce a Dense Latent Variable Fusion (DLVF) module that includes dense skip connections between the embedding of the noise and diffusion network. DLVF significantly improves shadow removal results without introducing additional parameters or running complexity. In summary, this paper makes the following contributions:

\begin{itemize}[noitemsep]
    \item A new shadow removal model that addresses the challenging task of general and instance-level shadow removal. This is the first work, to the best of our knowledge, to demonstrate the applicability of diffusion models for instance shadow removal.
    \item We show that it is possible to acquire compact and perceptual guidance in a learned feature space that is optimized together with the diffusion models, without relying on handcrafted features or physical quantities.
    \item We identify the local optimum of diffusion models that degrades the model results and introduce a dense latent variable fusion module to alleviate it, leading to significant performance improvement.  
\end{itemize}

\begin{figure*}[t]
    \centering
    \includegraphics[width=.72\linewidth]{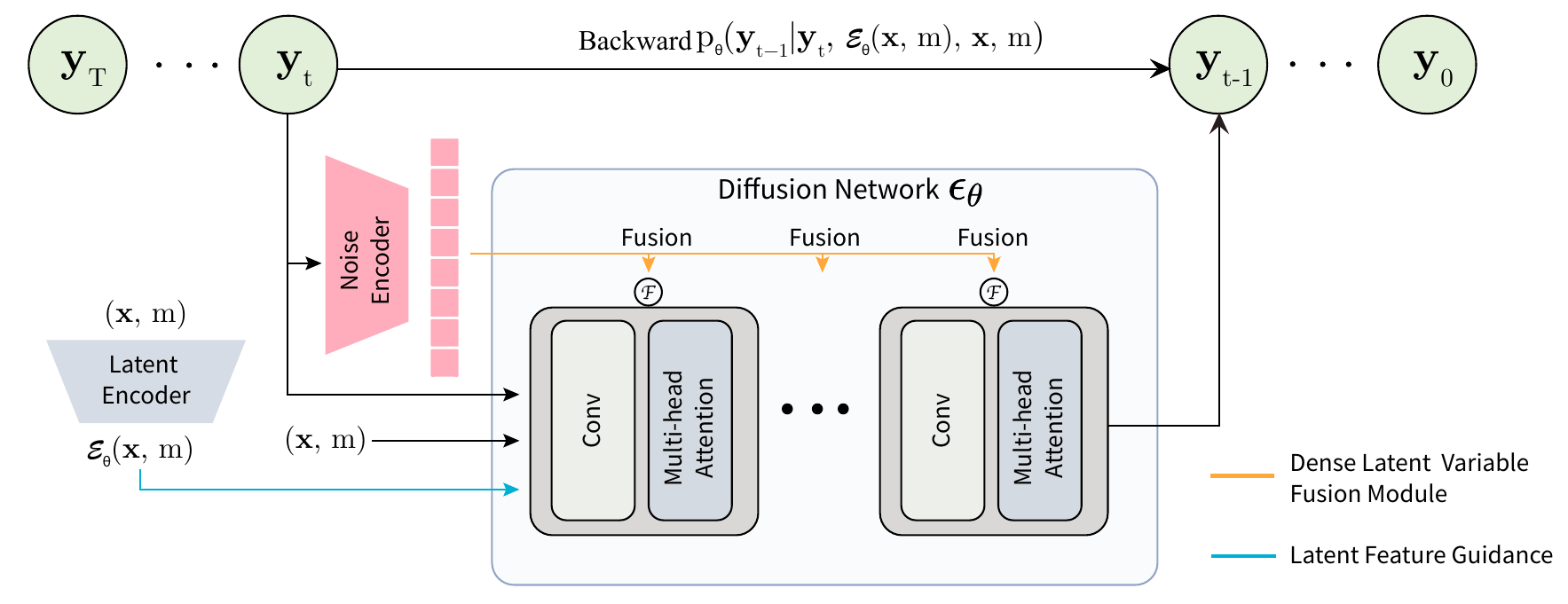}
    \caption{Our diffusion model architecture is illustrated in this backward diffusion diagram. The latent feature encoder $\mathcal{E}_\theta(\cdot)$ takes the shadow image $\bx\in\mathbb{R}^{3 \times H \times W}$ and shadow mask $m \in \mathbb{R}^{1 \times H \times W}$ as input, with a resolution of $H\times W$, and acquires the latent feature in a compressed dimension of $1 \times H \times W$. The diffusion network $\epsilon_\theta(\cdot)$ conditioned on $(\bx, m)$ takes the latent feature concatenated with the noisy image $\by_t\in \mathbb{R}^{3 \times H \times W}$ as input, and estimates the noiseless image $\by_{t-1}\in \mathbb{R}^{3 \times H \times W}$ at each diffusion process $p_\theta(\cdot)$. In this process, the noise encoder takes the noise image $\by_t$ as input and acquires a 1-D vector as the noise embedding, which is fused with the diffusion network features by modulation for escaping the local optima.}
    \label{fig:backbone}
\end{figure*}

\section{Related Work}
\noindent\textbf{Shadow Removal.} The major challenge of modern learning-based shadow removal approaches comes from the large diversity of real-world shadow scenes.  The performance of recent shadow removal methods degrades significantly on out-of-distribution scenes~\cite{arbel2010shadow}.
Various approaches have been explored for addressing this issue, such as using physical illumination models, handcrafted priors, and image gradients~\cite{finlayson2005removal, finlayson2009entropy, guo2012paired}.
The recent trend has been in developing learning-based methods that can predict shadow-free scenes~\cite{hu2019mask, chen2021canet, fu2021auto, jin2021dc, liu2021shadow} or intermediate factors~\cite{le2019shadow, le2020shadow} for restoration.  
These methods have improved from previous methods in learning data~\cite{liu2021shadow}, shadow effects ~\cite{jin2021dc}, network architecture~\cite{chen2021canet, wan2022crformer, guo2023shadowformer}, and learning target decomposition~\cite{le2019shadow, zhang2020ris, le2020shadow}.
In particular, generative models have gained some traction for shadow removal.
ARGAN~\cite{ding2019argan} removes the effect of shadow in a progressive manner determined by a discriminator.
Nevertheless, these end-to-end GAN-based methods lack generalizability on the out-of-distribution shadow images without significant modifications.
Recent diffusion models have shown promising performance in general image restoration tasks but are rarely explored in shadow removal~\cite{meng2021sdedit, saharia2022palette, lugmayr2022repaint}.
In this work, we first propose to apply diffusion models for removing shadows, to leverage their impressive capacity of perceptual synthesis, which is shown to be capable of gradually preserving details in denoising sequences.

\noindent\textbf{Latent Feature Space Guidance.}
Guidance has become an essential component of diffusion models and powers spectacular image generation results in recent works.
Typical guidance for diffusion models includes class information~\cite{dhariwal2021diffusion}, text description~\cite{ramesh2022hierarchical, saharia2022photorealistic}, and even gradients~\cite{ho2022classifier}.
Nevertheless, these features cannot be easily adopted in shadow removal to provide more guidance than images.
In literature, physical quantities and handcrafted features have been heavily explored for guiding the restoration network. Zhu et al.~\cite{zhu_bijective_2022} propose to guide the network with an estimated shadow-invariant color map, and Wan et al.~\cite{wan2022} propose to guide the network with coarse de-shadowed images.
Illumination invariant representations~\cite{funt1995color, stricker1995similarity, finlayson1996color} are another related approach that aims to decompose intrinsic images by finding quantities invariant to color, density, or shading.
In our approach, we define a new latent feature space for guiding diffusion models.
By maximizing the similarity between the shadow and shadow-free latents,
we empirically demonstrate that it better guides the diffusion model to remove shadows by encapsulating essential perceptual information as a shadow-free prior.

\noindent\textbf{Posterior Collapse.}
The problem of posterior collapse refers to undesirable local optima first observed in the training of VAE models~\cite{kingma2013auto}.
Efforts to address it have included aggressive optimization of the inference network proposed by He et al.~\cite{he2019lagging}, weakening the generator by Fu et al.~\cite{fu2019cyclical}, and changing the objective function by Tolstikhin et al.~\cite{tolstikhin2017wasserstein}.
In this work, we show that although this issue has primarily been investigated in VAE models, conditional diffusion models can also suffer from similar issues.
Specifically, the conditions used in diffusion models usually provide stronger guidance compared to the latent noise variable.
Inspired by previous efforts to address the issue, we propose a new Dense Latent Variable Fusion (DLVF) module for diffusion models and experimentally demonstrate that this design improvement improves shadow removal results without introducing additional costs or modifications.
\added{Different from the other latent-based diffusion methods~\cite{preechakul2022diffusion, rombach2022high}, ours uses simpler pixel space and models shadow-free image distribution.}

\section{Proposed Method}
\deleted{
Our proposed shadow removal method employs diffusion models to produce high-quality and accurate results, as shown in Fig.~\ref{fig:backbone}. The diffusion network takes the shadow image, mask, and latent feature as inputs to guide the diffusion process, preserving the underlying scene structure while refining the details of the shadow regions. In Sec.~\ref{sec:cdn}, we provide a concise overview of our methodology, followed by a detailed discussion of the learned latent space features in Sec.~\ref{sec:ifsg}. Finally, in Sec.~\ref{sec:dpffm}, we discuss our approach to address the local optimum of diffusion models.
}

\subsection{Conditional Diffusion Models}
\label{sec:cdn}
\noindent\textbf{Diffusion Forward Process.}
The denoising diffusion models have been shown to be effective for modeling complex data distributions by reversing a gradual noising process.
For the shadow-free image distribution, we define the forward diffusion process that destroys a shadow-free image $\by \sim q(\by)$ with $T$ successive standard noises:
\setlength{\belowdisplayskip}{2pt} \setlength{\belowdisplayshortskip}{2pt}
\setlength{\abovedisplayskip}{2pt} \setlength{\abovedisplayshortskip}{2pt}
\begin{align}
\begin{split}    
  q(\by_t| \by_{t-1}) &= \mathcal{N}\left(\by_t ; \sqrt{{\beta}_{t}} \by_0,\left(1-{\beta}_{t}\right) \mathbf{I}\right).
\end{split}
\end{align}
Alternatively, we can use the reparameterization trick~\cite{kingma2013auto} to express this as:
\begin{align}
\label{eq:alpha}
\begin{split}
  q(\by_t | \by_{0}) &= \mathcal{N}\left(\by_t ; \sqrt{\bar{\alpha}_{t}} \by_0,\left(1-\bar{\alpha}_{t}\right) \mathbf{I}\right) \\
  &= \sqrt{\bar{\alpha}_{t}} \by_0 + \epsilon \sqrt{1-\bar{\alpha}_{t}}, \epsilon \sim \mathcal{N}(0, \mathbf{I}),
\end{split}
\end{align}
where the variance schedule $\{\beta_1, \dots, \beta_T\}$ linear increases and has a closed form $\alpha_t =1 - \beta_t$ and $\bar \alpha_t := \prod^t_{s=1} \alpha_s$.

\noindent\textbf{Diffusion Backward Process.}
The reversion of $q(\by_t| \by_{t-1})$ is tractable by conditioning on image $\by_0$, and it results in sampling arbitrary shadow-free images  from noise $\by_T \sim \mathcal{N}(0, I)$ for removal as:
\begin{align}
q\left(\by_{t-1} | \by_t, \by_0\right) =\mathcal{N}\left(\by_{t-1} ; \tilde{\mu}\left(\by_t, \by_0\right), \tilde{\beta}_t \mathbf{I}\right).
\end{align}
According to Bayes' rule and Eq.~\eqref{eq:alpha}, we represent $\tilde{\mu}_t$ as:
\begin{align}
    \tilde{\mu}_t\left(\by_t, \by_0\right):=\frac{1}{\sqrt{\alpha_t}}(\by_t - \frac{1-\alpha_t}{\sqrt{1-\bar \alpha_t}}\epsilon_t).
\end{align}
Ho et al.~\cite{ho_denoising_2020} suggests modeling the process with $p_\theta$ by optimizing the \emph{variational lower bound} ($L_{VLB}$) as:
\begin{align}
L_\text{VLB} &= L_T + L_{T-1} + \dots + L_0,
\end{align}
which is defined with Kullback–Leibler (KL) divergence as:
\begin{align}
\begin{split}
\label{eq:kl}
L_T &= D_\text{KL}(q(\mathbf{y}_T \vert \mathbf{y}_0) \parallel p_\theta(\mathbf{y}_T)), \\
L_t &= D_\text{KL}(q(\mathbf{y}_t \vert \mathbf{y}_{t+1}, \mathbf{y}_0) \parallel p_\theta(\mathbf{y}_t \vert\mathbf{y}_{t+1})), \\
L_0 &= - \log p_\theta(\mathbf{y}_0 \vert \mathbf{y}_1),
\end{split}
\end{align}
and the effective simplification of $L_t$ is
\begin{align}
    \label{eq:sloss}
    L_{\text {simple }}:=E\left[\left\|\epsilon-\epsilon_\theta\left(\by_t, t\right)\right\|^2\right].
\end{align}

\noindent\textbf{Diffusion Conditioning.}
A straightforward approach to producing shadow-free results is to condition diffusion models on the shadow image $\bx$ and shadow mask $m$ by concatenating them with noise $\by_t$ along the channel dimension: 
\begin{align}
\label{eq:base}
p_\theta\left(\by_{t-1} | \by_t\right) & := p_\theta\left(\by_{t-1} | \by_t,\bx, m, t\right).
\end{align}
In the following sections, we will discuss our improvement based on the baseline following Eq.~\eqref{eq:base} that takes image $\bx$, $\by_t$, and mask $m$ as input and predicts the shadow-free noise $\by_{t-1}$ for effective shadow removal as $p_\theta\left(\by_{t-1} | \by_t, t\right)$.

\begin{figure}[t!]
    \centering
    \includegraphics[width=1\linewidth]{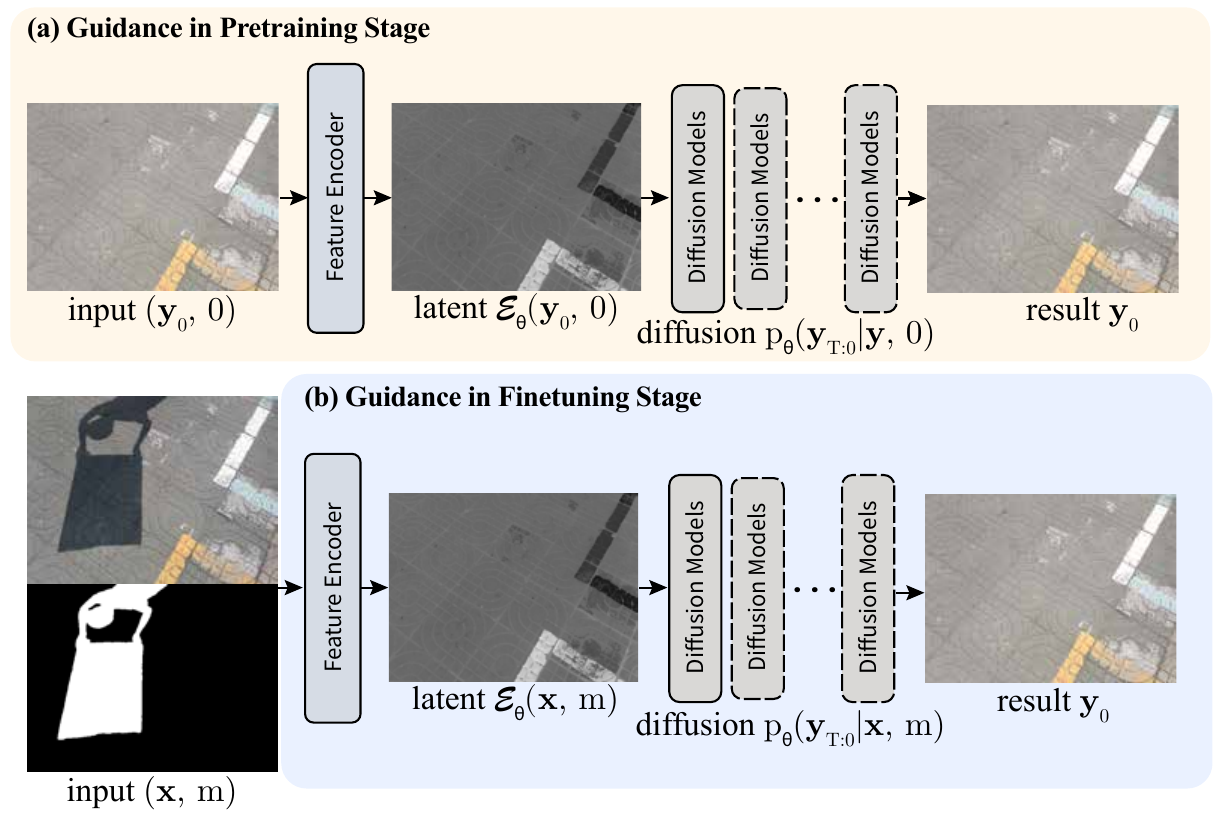}
    \caption{The diagram illustrates the two-stage learning approach used in our proposed method. In the pretraining stage (top row), the diffusion network is trained on shadow-free images to learn a latent feature space that captures informative shadow-free priors as guidance. In the finetuning stage (bottom row), we initialize the diffusion network with the pretraining weights from (a) for shadow removal under the latent feature guidance.}
    \label{fig:stage}
\end{figure}

\begin{figure}[htbp]
\centering
\begin{subfigure}[t]{.24\linewidth}
\captionsetup{justification=centering, labelformat=empty}
\includegraphics[width=1\linewidth]{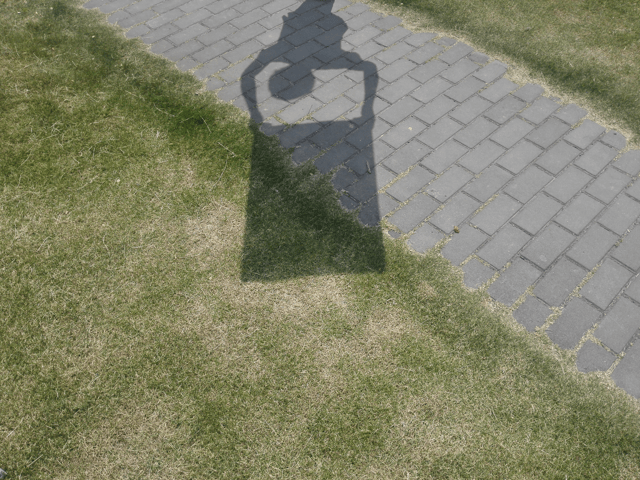}
\end{subfigure}
\begin{subfigure}[t]{.24\linewidth}
\captionsetup{justification=centering, labelformat=empty}
\includegraphics[width=1\linewidth, height=.75\linewidth]{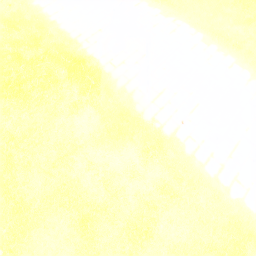}
\end{subfigure}
\begin{subfigure}[t]{.24\linewidth}
\captionsetup{justification=centering, labelformat=empty}
\includegraphics[width=1\linewidth]{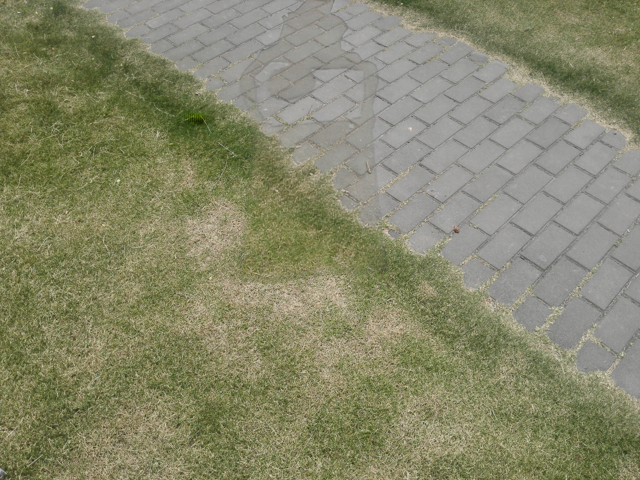}
\end{subfigure}
\begin{subfigure}[t]{.24\linewidth}
\captionsetup{justification=centering, labelformat=empty}
\includegraphics[width=1\linewidth, height=.75\linewidth]{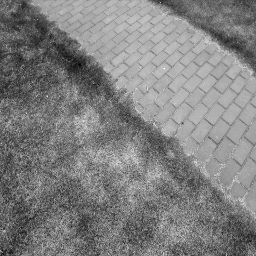}
\end{subfigure}
\hfill

\begin{subfigure}[t]{.24\linewidth}
\captionsetup{justification=centering, labelformat=empty}
\includegraphics[width=1\linewidth]{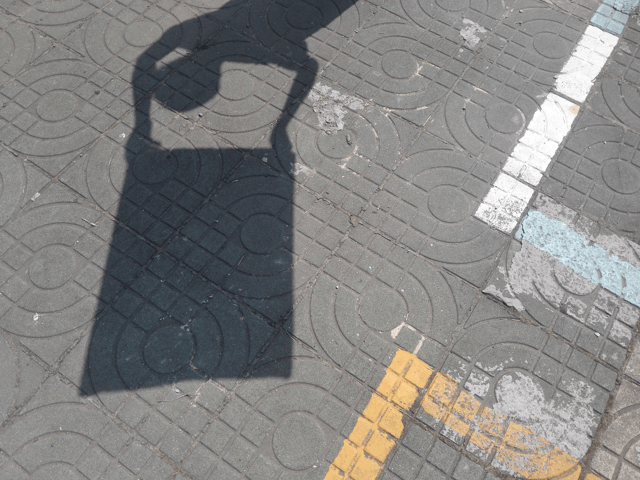}
\caption{(a)}
\end{subfigure}
\begin{subfigure}[t]{.24\linewidth}
\captionsetup{justification=centering, labelformat=empty}
\includegraphics[width=1\linewidth, height=.75\linewidth]{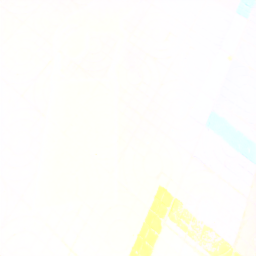}
\caption{(b)}
\end{subfigure}
\begin{subfigure}[t]{.24\linewidth}
\captionsetup{justification=centering, labelformat=empty}
\includegraphics[width=1\linewidth]{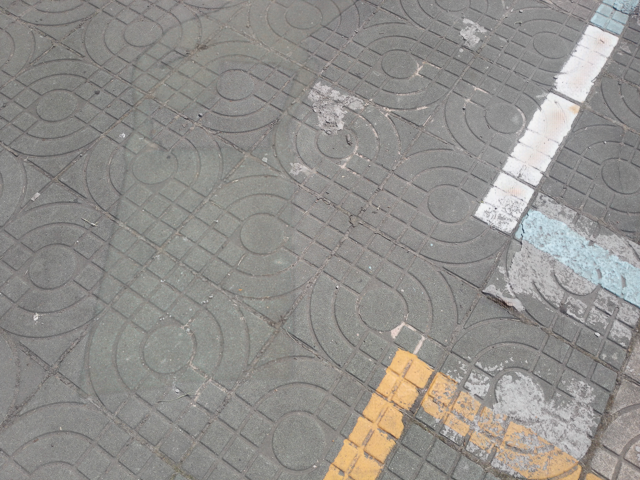}
\caption{(c)}
\end{subfigure}
\begin{subfigure}[t]{.24\linewidth}
\captionsetup{justification=centering, labelformat=empty}
\includegraphics[width=1\linewidth, height=.75\linewidth]{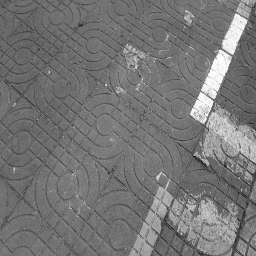}
\caption{(d)}
\end{subfigure}
\hfill

\vspace{-.5\baselineskip}
\caption{Visual comparisons of different guidance strategies in shadow removal literature. \textbf{(a)} to \textbf{(d)}: shadow image, invariant color map~\cite{zhu_bijective_2022}, coarse deshadowed image~\cite{wan2022}, and our learned latent feature. Our approach provides more perceptual information than \textbf{(b)} and contains fewer shadow features than \textbf{(c)}, which still retains a shadow boundary. 
}
\label{fig:latent}
\vspace{-1\baselineskip}
\end{figure}

\subsection{Latent Feature Guidance}
\label{sec:ifsg}

The proposed latent feature encoder $\mathcal{E}_\theta(\cdot)$ uses the same network architecture as the diffusion network $\epsilon_\theta(\cdot)$ with the exception of a timestep embedding and predicts a single-channel feature map that has the same spatial dimension as the shadow image $x$.
It guides the diffusion process Eq.~\eqref{eq:base} by concatenating the guidance with conditions:
\begin{align}
\label{eq:final}
p_\theta\left(\by_{t-1} | \by_t\right) & := p_\theta\left(\by_{t-1} | \by_t, \mathcal{E}_\theta(\bx, m), t \right).
\end{align}
We propose to learn to extract shadow-free priors using the latent feature space by minimizing the invariant loss between the encoded shadow-free images and shadow images with shadow masks as:
\begin{align}
\label{eq:invariant}
\arg \min_{\theta} \left\| \mathcal{E}_\theta(\by_0, \mathbf{0})  -  \mathcal{E}_\theta(\bx, m) \right\|^2.
\end{align}
In order to extract a compact and perceptual feature space to guide the diffusion model, we optimize the encoder together with the whole network during training based on Eq.~\eqref{eq:sloss}:
\begin{align}
    \label{eq:loss}
    \begin{split}
    L_{\text {simple }}:=&E\left[\left\|\epsilon-\epsilon_\theta\left(\by_t, \mathcal{E}_\theta(\bx, m), \bx, m, t\right)\right\|^2\right]\\
    &+\left\| \mathcal{E}_\theta(\by_0, \mathbf{0})  -  \mathcal{E}_\theta(\bx, m) \right\|^2.
    \end{split}
\end{align}

Moreover, we empirically find that pretraining the diffusion model $\epsilon_\theta(\cdot)$ and then finetuning it accelerates the optimization of Eq.~\eqref{eq:loss}. Intuitively, the pretraining strategy provides a good starting point to finetune the diffusion model, such that the encoder has already learned to model the important characteristics of shadow-free images such as shadow-free textures and colors. This feature space provides strong guidance during finetuning for minimizing shadow features with the invariant loss, allowing the model to achieve higher-quality results.

\begin{figure}[htbp]
    \centering
    \includegraphics[width=\linewidth]{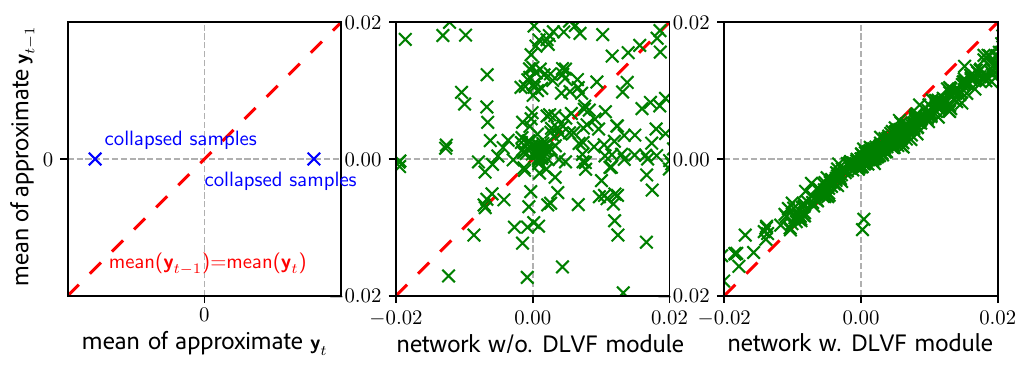}
    \caption{We visualize the mean space of variables to show the collapse and our effects. The horizontal and vertical axis represent the mean of predicted $\textbf{y}_{t}$ and $\textbf{y}_{t-1}$, respectively. The dashed diagonal line represents when the approximate noise is relevant. By projecting denoised samples, the results show the network with our DLVF (third) successfully moves points onto the diagonal line and away from collapses compared to without it (second).}
    \label{fig:dpffv}
\end{figure}

Subsequently, we propose a two-stage learning approach for guiding the diffusion models including pretraining and finetuning as shown in Fig.~\ref{fig:stage} as:
\begin{itemize}[noitemsep]
    \item Optimize the diffusion network $\epsilon_\theta$ together with the latent encoder $\mathcal{E}_\theta$ for modeling the characteristics of shadow-free images by minimizing the loss:
    \begin{align}
    E\left[\left\|\epsilon-\epsilon_\theta\left(\by_t, \mathcal{E}_\theta(\by_0, 0), \by_0, m, t\right)\right\|^2\right].
    \end{align}    
    \item Finetune the encoder $\mathcal{E}_\theta$ and diffusion network $\epsilon_\theta$ by optimizing Eq.~\eqref{eq:loss} to effectively remove shadows and preserve the underlying texture.
\end{itemize}

To demonstrate the effectiveness of our proposed latent feature guidance, Fig.~\ref{fig:latent} compares it with existing guidance strategies in shadow removal literature, including ~\cite{wan2022}, which conditions on estimated coarse de-shadowed images, and ~\cite{zhu_bijective_2022}, which conditions on estimated invariant color maps for restoration. Our approach preserves more shadow-free perceptual details compared to the estimated invariant color map, which only consists of large color blocks. Similarly, our approach retains fewer shadow features compared to the estimated coarse de-shadowed image, which still retains shadow boundaries that may lead to incorrect results.

\subsection{Dense Latent Variable Fusion Module}
\label{sec:dpffm}
The phenomenon known as posterior collapse occurs when the training procedure of generative models falls into a trivial local optimum of $L_{VLB}$, causing the model to ignore the latent variable and collapse the model posterior to the prior, which has only been discussed in VAE~\cite{he2019lagging}.
Given the intrinsic similarity between diffusion models and VAE, we first determine the collapse issue of diffusion models under guidance and then address it with a new module.

In our proposed diffusion models, we parameterize the variational distribution $p_\theta(\by_{t-1}|\by_t ,\by_0)$ with the latent variable $\by_t$ under the guidance $\mathcal{E}_\theta(\mathbf{x}, m)$ in Eq.~\eqref{eq:final}. In this case, the local optima are characterized by:
\begin{equation}
    \begin{aligned}
    p_\theta(\mathbf{y}_{t-1}) &=p_\theta\left(\mathbf{y}_{t-1} \mid \mathbf{y}_t, \mathcal{E}_\theta(\mathbf{x}, m), t\right) \\
    &= p_\theta\left(\mathbf{y}_{t-1} \mid \mathbf{y}_t\right) p_\theta\left(\mathbf{y}_{t-1} \mid \mathcal{E}_\theta(\mathbf{x}, m), t\right) \\
    &:= p_\theta\left(\mathbf{y}_{t-1} \mid \mathcal{E}_\theta(\mathbf{x}, m), t\right).
    \end{aligned}
\end{equation}
This is undesirable since a crucial goal of diffusion models is to produce diverse outputs. This is particularly important for shadow removal, where complex shadow distributions exist that cannot be easily represented by guidance alone.

Much attention has been devoted to remedying the posterior collapse of VAE models. However, some of these methods weaken the encoder or modeling capability of posterior-related components, as observed in \cite{he2019lagging, fu2019cyclical}. Other approaches, such as those proposed in \cite{yang2017improved, tolstikhin2017wasserstein}, significantly complicate the optimization.

In this work, we introduce a new Dense Latent Variable Fusion (DLVF) module that works in tandem with the diffusion network to establish strong links between the latent variable and the generated results.
To elaborate, for each block $\mathcal{G}(\cdot)^n$ of the diffusion network~\cite{dhariwal2021diffusion} at level $n$, the feature $h^{n-1}$ and the embedding of $\mathrm{emb}(\by_t)$ generated by a three-layer MLP are both inputted as:
\begin{align}
    h^{n} = \mathcal{G}(h^{n-1}, \mathrm{emb}(\by_t) \downarrow_{2n}, t)^n,
\end{align}
where $\downarrow_{2n}$ denotes the pooling operation with a scaling factor of $2n$ to match the dimension of the features $h^{n-1}$.
To achieve a larger receptive field for the latent variable embedding, we employ fully-connected layers as an additional encoder before inputting them into the network and use adaptive pooling operations to transform the noise into vectors:
\begin{align}
    \by_t' = \mathcal{P}_{ooling}(\mathcal{E}_{\mathrm{noise}}(\by_t)), \by_t' \in \mathbb{R}^{1\times N},
\end{align}
where $N$ is the size of the vector noise.
The connection between the embedding $\mathrm{emb}(\by_t) \downarrow_{2n}$ and the network features $h^{n-1}$ is conducted by point-wise summing.

To demonstrate the collapse and effectiveness of our method, we visualize the correspondence between the latent variable $\by_t$ and approximated $\by_{t-1}$, which are randomly selected from $T$ denoising processes, shown in Fig.~\ref{fig:dpffv}.
In comparison to the baseline without our fusion strategies, our method shows a stronger correspondence between the two variables, indicating a better optimum in the training dynamics. This ultimately results in more effective removal.

\begin{table*}[t]
    \caption{\textbf{Quantitative result comparisons of our methods and the state-of-the-art methods on \emph{AISTD}}. The best and second-best performance is indicated with \textbf{bold} and \emph{\underline{italic}} respectively. We use $\uparrow$ and $\downarrow$ to suggest better high/lower score.}
    \vspace{-.5\baselineskip}
    \label{tab:aistd}
    \centering\small
    \resizebox{.9\linewidth}{!}{
    \begin{tabular}{lc|ccc|ccc|ccc}
    \toprule
    & & \multicolumn{3}{c|}{\emph{shadow region}} & \multicolumn{3}{c|}{\emph{non-shadow region}} &  \multicolumn{3}{c}{\emph{all} image} \\
    \midrule
    Method &  & RMSE $\downarrow$ & PSNR $\uparrow$ & SSIM $\uparrow$  &  RMSE $\downarrow$ & PSNR $\uparrow$ & SSIM $\uparrow$ & RMSE  $\downarrow$ & PSNR $\uparrow$ & SSIM $\uparrow$ \\
    \midrule
    SP+M-Net~\cite{le2019shadow} & ICCV-19 & 5.93 & 37.96 & 0.990 & 3.05 & 35.77 & 0.973 & 3.51 & 32.90 & 0.957  \\
    DHAN~\cite{cun_towards_2020} & AAAI-20 & 11.38 & 33.18 & 0.987 & 7.15 & 27.10 & 0.972 & 7.81 & 25.65 & 0.954 \\
    Param+M+D-Net~\cite{le2020shadow} & ECCV-20 & 9.67 & 33.46 & 0.985 & 2.91 & 34.85 & 0.974 & 3.98 & 30.13 & 0.945 \\
    G2R-ShadowNet~\cite{liu2021shadow} & CVPR-21 & 7.38 & 36.24 & 0.988 & 3.00 & 35.26 & 0.975 & 3.69 & 31.90 & 0.953  \\
    Auto-Exposure~\cite{fu2021auto} & CVPR-21 & 6.57 & 36.30 & 0.976 & 3.83 & 31.10 & 0.874 & 4.27 & 29.44 & 0.838  \\
    DC-ShadowNet~\cite{jin2021dc} & ICCV-21 & 10.57 & 32.15 & 0.976 & 3.82 & 34.99 & 0.969 & 4.80 & 28.75 & 0.925  \\
    EMDN~\cite{zhu2022efficient} & AAAI-22 & 7.94 & 36.44 & 0.986 & 4.78 & 31.80 & 0.962 & 5.28 & 29.98 & 0.940 \\
    SG-ShadowNet~\cite{wan2022} & ECCV-22 & 5.93 & 37.25 & 0.989 & 3.00 & 35.27 & 0.975 & 3.46 & 32.42 & 0.956  \\
    BMN~\cite{zhu_bijective_2022} & CVPR-22 & \emph{\underline{5.69}} & \emph{\underline{38.00}} & \textbf{0.991} & \emph{\underline{2.52}} & \emph{\underline{37.35}} & \textbf{0.981} & \emph{\underline{3.02}} & \emph{\underline{33.93}} & \emph{\underline{0.966}} \\
    \midrule
    Palette Diffusion~\cite{saharia2022palette} & SIGGRAPH-22 & 15.40 & - & - & 7.82 & - & - & 6.41 & - & - \\
    Repaint Diffusion~\cite{lugmayr2022repaint} & CVPR-22 & 12.90 & - & - & 10.66 & - & - & 24.90 & - & - \\
    \rowcolor{Gray} \ours & Ours & \textbf{5.15} & \textbf{39.36} & \textbf{0.991} & \textbf{2.47} & \textbf{37.69} & \textbf{0.981} & \textbf{2.90} & \textbf{34.69} & \textbf{0.968} \\
    \midrule
    \emph{shadow image} & & 39.72 & 20.87 & 0.944 & 2.51 & 36.63 & 0.980 & 8.38 & 20.45 & 0.908 \\
    \bottomrule
    \end{tabular}}
    \vspace{-.5\baselineskip}
\end{table*}

\begin{figure*}[htbp]
\centering
  \begin{subfigure}[t]{.135\linewidth}
    \captionsetup{justification=centering, labelformat=empty, font=scriptsize}
    \includegraphics[width=1\linewidth,height=.75\linewidth]{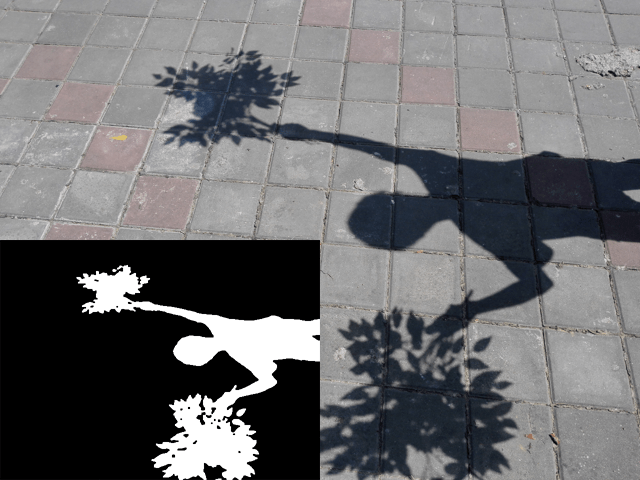}
    \includegraphics[width=1\linewidth,height=.75\linewidth]{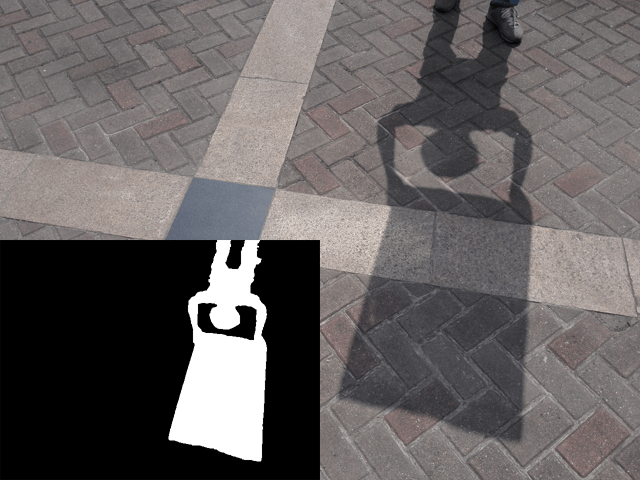}
    \caption{Shadow \& Mask}
  \end{subfigure}
  \begin{subfigure}[t]{.135\linewidth}
    \captionsetup{justification=centering, labelformat=empty, font=scriptsize}
    \includegraphics[width=1\linewidth,height=.75\linewidth]{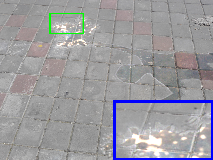}
    \includegraphics[width=1\linewidth,height=.75\linewidth]{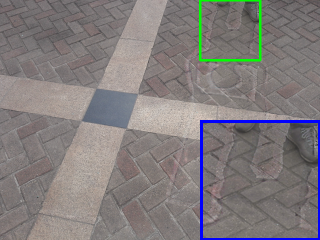}
    \caption{\tiny Param+M+D-Net~\cite{le2020shadow}}
  \end{subfigure}
  \begin{subfigure}[t]{.135\linewidth}
    \captionsetup{justification=centering, labelformat=empty, font=scriptsize}
    \includegraphics[width=1\linewidth,height=.75\linewidth]{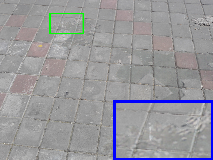}
    \includegraphics[width=1\linewidth,height=.75\linewidth]{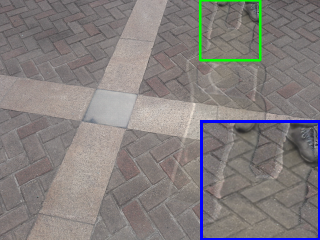}
    \caption{\tiny G2R-ShadowNet~\cite{liu2021shadow}}
  \end{subfigure}
  \begin{subfigure}[t]{.135\linewidth}
    \captionsetup{justification=centering, labelformat=empty, font=scriptsize}
    \includegraphics[width=1\linewidth,height=.75\linewidth]{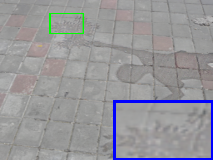}
    \includegraphics[width=1\linewidth,height=.75\linewidth]{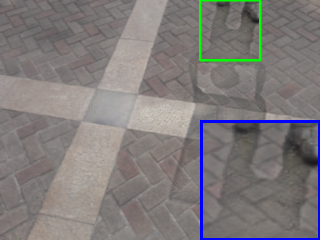}
    \caption{DC-ShadowNet~\cite{jin2021dc}}
  \end{subfigure}
  \begin{subfigure}[t]{.135\linewidth}
    \captionsetup{justification=centering, labelformat=empty, font=scriptsize}
    \includegraphics[width=1\linewidth,height=.75\linewidth]{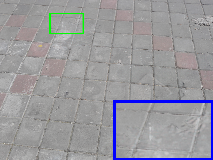}
    \includegraphics[width=1\linewidth,height=.75\linewidth]{resources/major-comparison/aistd_103-4/comparisons_DC-ShadowNet.png}
    \caption{SG-ShadowNet~\cite{wan2022}}
  \end{subfigure}
  \begin{subfigure}[t]{.135\linewidth}
    \captionsetup{justification=centering, labelformat=empty, font=scriptsize}
    \includegraphics[width=1\linewidth,height=.75\linewidth]{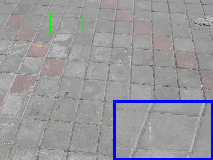}
    \includegraphics[width=1\linewidth,height=.75\linewidth]{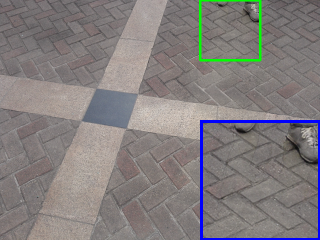}
    \caption{\tiny \ours~(Ours)}
  \end{subfigure}
  \begin{subfigure}[t]{.135\linewidth}
    \captionsetup{justification=centering, labelformat=empty, font=scriptsize}
    \includegraphics[width=1\linewidth,height=.75\linewidth]{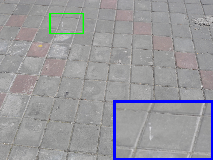}
    \includegraphics[width=1\linewidth,height=.75\linewidth]{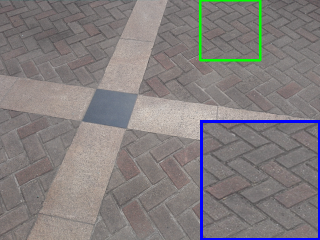}
    \caption{Ground Truth}
  \end{subfigure}
  \vspace{-.5\baselineskip}
  \caption{\textbf{Visual comparisons of the representative hard shadow removal results on \emph{AISTD} dataset.} Here we highlight the details of shadow regions that are marked with green box in the blue box area, where ours best perseveres details and removes shadow effects. Please see the supplement for additional visual results.}
  \vspace{-1\baselineskip}
  \label{fig:main_aistd}
\end{figure*}

\begin{table*}[ht]
    \caption{\textbf{Quantitative comparison results of our methods and the state-of-the-art methods on the \emph{ISTD dataset} and \emph{SRD dataset}}. We want to remark on a slight performance drop in the non-shadow region of our method. The reason is that the two benchmarks are un-adjusted, which means the shadow and shadow-free image pairs were captured at different lighting environments. The color inconsistency would result in inaccurate non-shadow region and all image measurement.}
    \label{tab:istd}
    \vspace{-.5\baselineskip}
    \centering\small
    \subfloat[\textbf{ISTD dataset} results.]{
    \resizebox{.67\linewidth}{!}{
    \begin{tabular}{l|ccc|ccc|ccc}
    \toprule
    & \multicolumn{3}{c|}{\emph{shadow region}} & \multicolumn{3}{c|}{\emph{non-shadow region}} &  \multicolumn{3}{c}{\emph{all image}} \\
    \midrule
    Method  & RMSE $\downarrow$ & PSNR $\uparrow$ & SSIM $\uparrow$  &  RMSE $\downarrow$ & PSNR $\uparrow$ & SSIM $\uparrow$ & RMSE  $\downarrow$ & PSNR $\uparrow$ & SSIM $\uparrow$ \\
    \midrule
    DHAN~\cite{cun_towards_2020} & 7.53 & 35.82 & \emph{\underline{0.989}} & 5.33 & 30.95 & 0.971 & 5.68 & 29.09 & 0.953 \\
    G2R-ShadowNet~\cite{liu2021shadow} & 10.72 & 31.63 & 0.975 & - & - & - & - & - & -  \\
    Auto-Exposure~\cite{fu2021auto} & 7.82 & 34.94 & 0.973 & 5.59 & 28.57 & 0.862 & 5.94 & 27.19 & 0.824  \\
    DC-ShadowNet~\cite{jin2021dc} & 11.43 & 31.69 & 0.976 & 5.86 & 28.92 & 0.956 & 6.62 & 26.38 & 0.917 \\
    EMDN~\cite{zhu2022efficient} & 7.94 & \emph{\underline{36.44}} & 0.986 & 4.78 & 31.80 & 0.962 & 5.28 & 29.98 & 0.940  \\
     SG-ShadowNet~\cite{wan2022} & - & - & - & - & - & - & - & - & - \\
     BMN~\cite{zhu_bijective_2022} & \emph{\underline{7.44}} & 35.73 & \emph{\underline{0.989}} & \textbf{4.61} & \textbf{32.73} & \emph{\underline{0.976}} & \emph{\underline{5.06}} & \emph{\underline{30.26}} & \emph{\underline{0.957}}  \\
    \rowcolor{Gray} \ours~(Ours) & \textbf{6.41} & \textbf{37.19} & \textbf{0.990} & \emph{\underline{4.65}} & \emph{\underline{32.60}} & \textbf{0.977} & \textbf{4.93} & \textbf{30.64} & \textbf{0.963} \\
    \midrule
    \emph{shadow image} & 32.67 & 22.43 & 0.953 & 6.77 & 27.27 & 0.974 & 10.86 & 20.56 & 0.908 \\
    \bottomrule
    \end{tabular}
    }
    }
    \subfloat[\textbf{SRD dataset} result.]{
    \resizebox{.305\linewidth}{!}{
    \begin{tabular}{l|ccc}
    \toprule
    & \multicolumn{3}{c}{\emph{shadow region}} \\
    \midrule
    Method & RMSE $\downarrow$ & PSNR $\uparrow$ & SSIM \\
    \midrule
    DHAN~\cite{cun_towards_2020} & 8.94 & 33.67 & 0.978\\
    G2R-ShadowNet~\cite{liu2021shadow} & - & - & - \\
    Auto-exposure~\cite{fu2021auto} & 8.86 & 34.93 & 0.963  \\
    DC-ShadowNet~\cite{jin2021dc} & 7.86 & 36.34 & 0.970  \\
    EMDN~\cite{zhu2022efficient} & 9.83 & 32.48 & 0.928  \\
    SG-ShadowNet~\cite{wan2022} & 8.00 & 35.53 & 0.974  \\
   BMN~\cite{zhu_bijective_2022} & \emph{\underline{7.40}} & \emph{\underline{36.81}} & \textbf{0.979} \\
    \rowcolor{Gray} \ours~(Ours) & \textbf{6.81} & \textbf{37.42} & \textbf{0.979} \\
    \midrule
    \emph{shadow image} & 46.24 & 19.95 & 0.889 \\
    \bottomrule
    \end{tabular}
    }
    }
\end{table*}

\begin{figure*}[ht]
\centering
  \begin{subfigure}[t]{.12\linewidth}
    \captionsetup{justification=centering, labelformat=empty, font=scriptsize}
    \includegraphics[width=1\linewidth,height=1\linewidth]{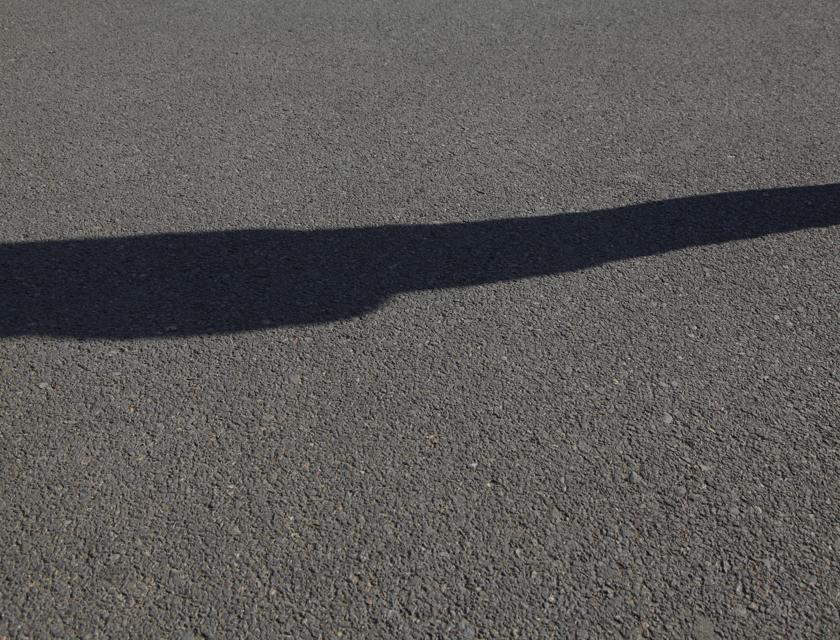}
    \includegraphics[width=1\linewidth,height=1\linewidth]{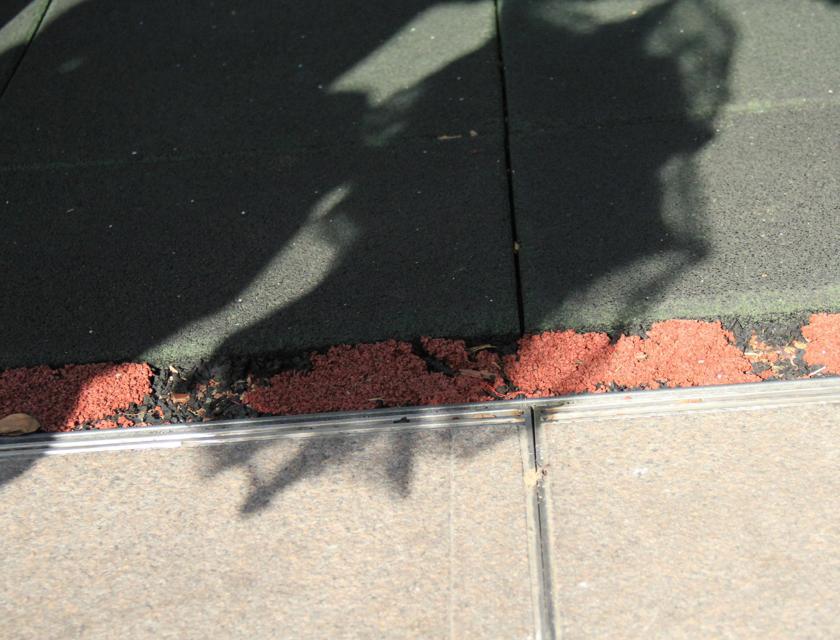}
    \caption{Shadows}
  \end{subfigure}
  \begin{subfigure}[t]{.12\linewidth}
    \captionsetup{justification=centering, labelformat=empty, font=scriptsize}
    \includegraphics[width=1\linewidth,height=1\linewidth]{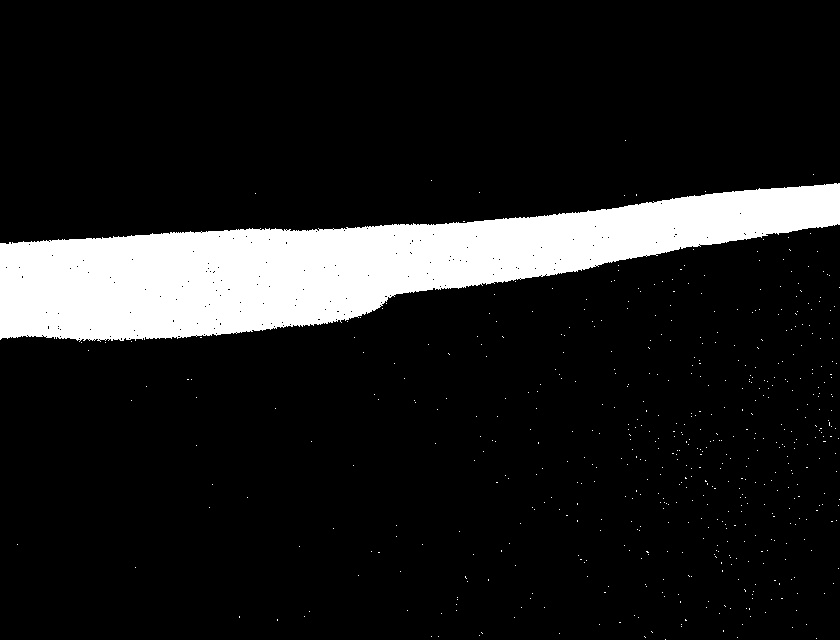}
    \includegraphics[width=1\linewidth,height=1\linewidth]{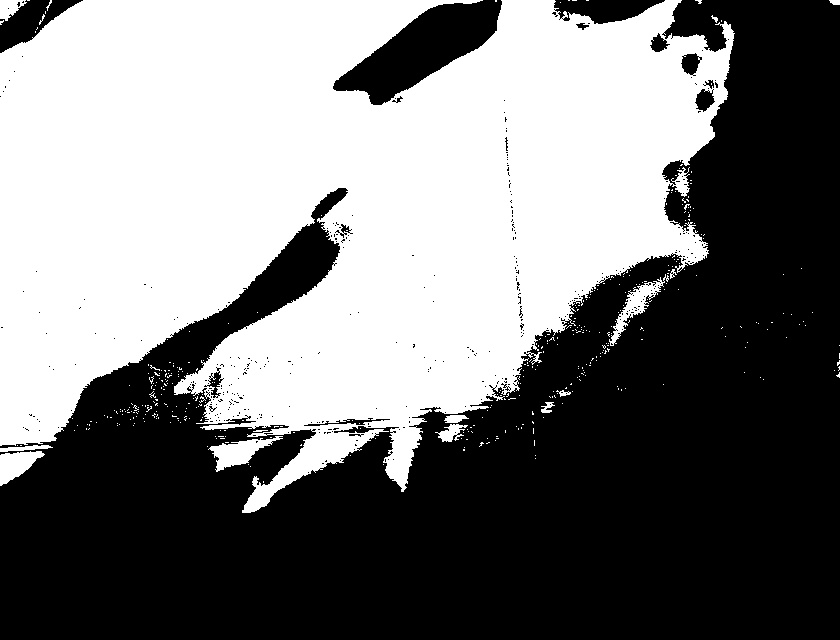}
    \caption{Shadow Mask}
  \end{subfigure}
  \begin{subfigure}[t]{.12\linewidth}
    \captionsetup{justification=centering, labelformat=empty, font=scriptsize}
    \includegraphics[width=1\linewidth,height=1\linewidth]{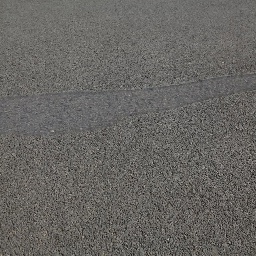}
    \includegraphics[width=1\linewidth,height=1\linewidth]{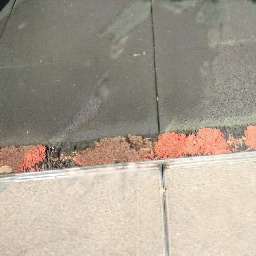}
    \caption{Auto-Exposure~\cite{fu2021auto}}
  \end{subfigure}
  \begin{subfigure}[t]{.12\linewidth}
    \captionsetup{justification=centering, labelformat=empty, font=scriptsize}
    \includegraphics[width=1\linewidth,height=1\linewidth]{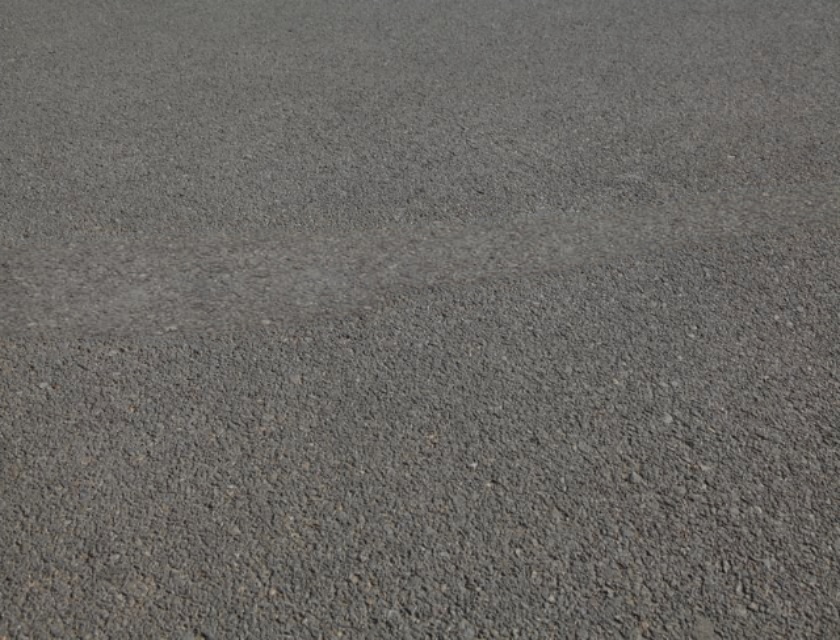}
    \includegraphics[width=1\linewidth,height=1\linewidth]{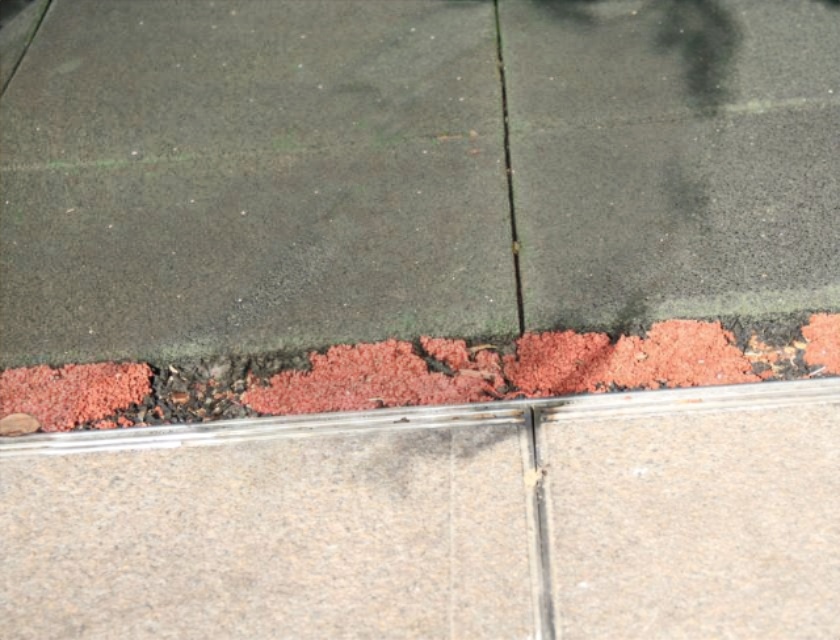}
    \caption{DHAN~\cite{cun_towards_2020}}
  \end{subfigure}
  \begin{subfigure}[t]{.12\linewidth}
    \captionsetup{justification=centering, labelformat=empty, font=scriptsize}
    \includegraphics[width=1\linewidth,height=1\linewidth]{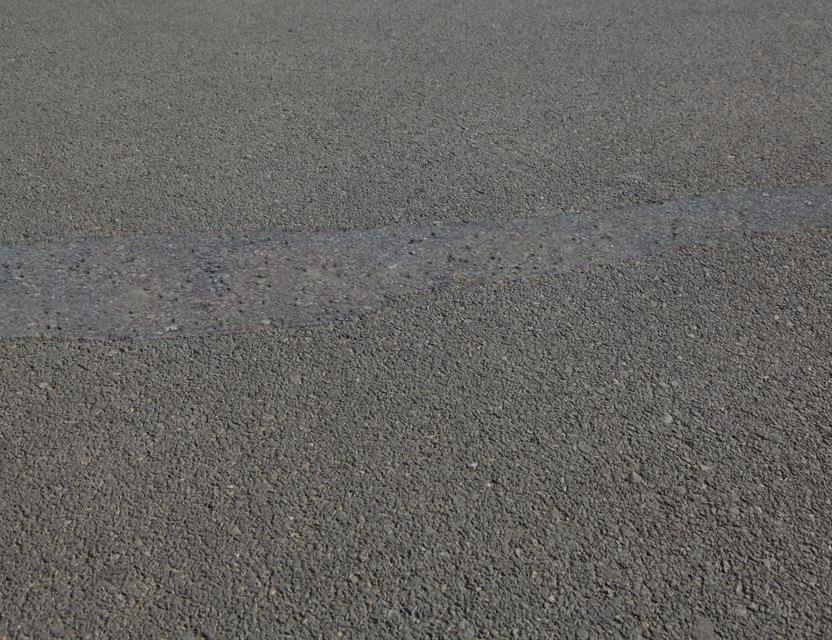}
    \includegraphics[width=1\linewidth,height=1\linewidth]{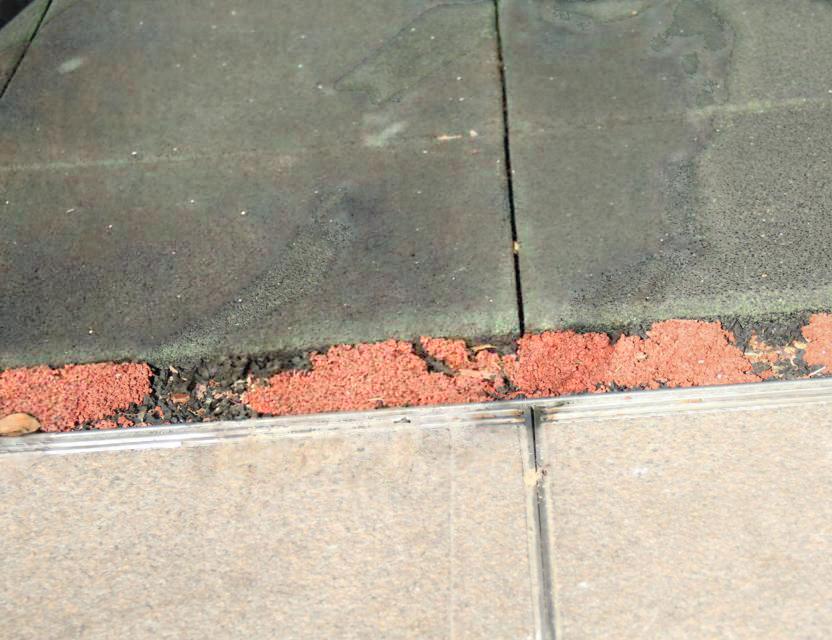}
    \caption{EMDN~\cite{zhu2022efficient}}
  \end{subfigure}
  \begin{subfigure}[t]{.12\linewidth}
    \captionsetup{justification=centering, labelformat=empty, font=scriptsize}
    \includegraphics[width=1\linewidth,height=1\linewidth]{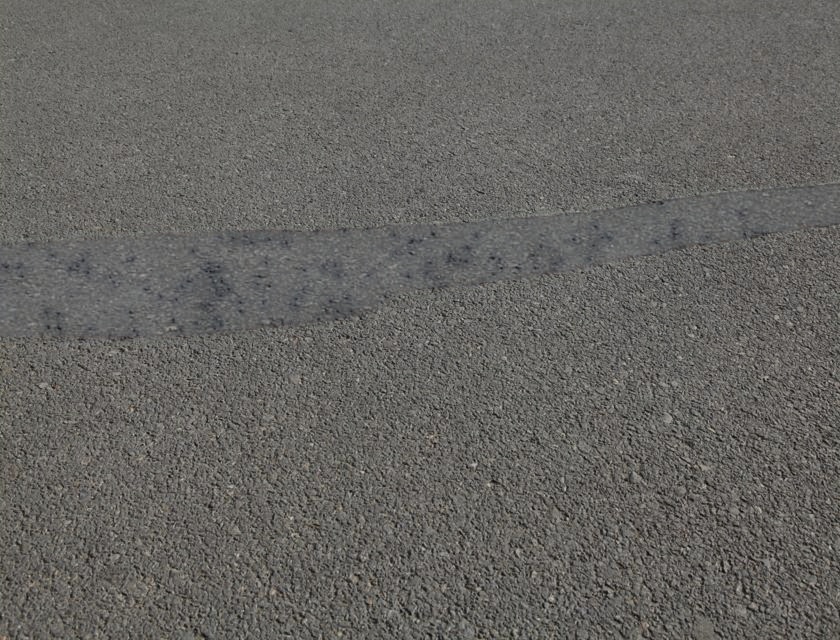}
    \includegraphics[width=1\linewidth,height=1\linewidth]{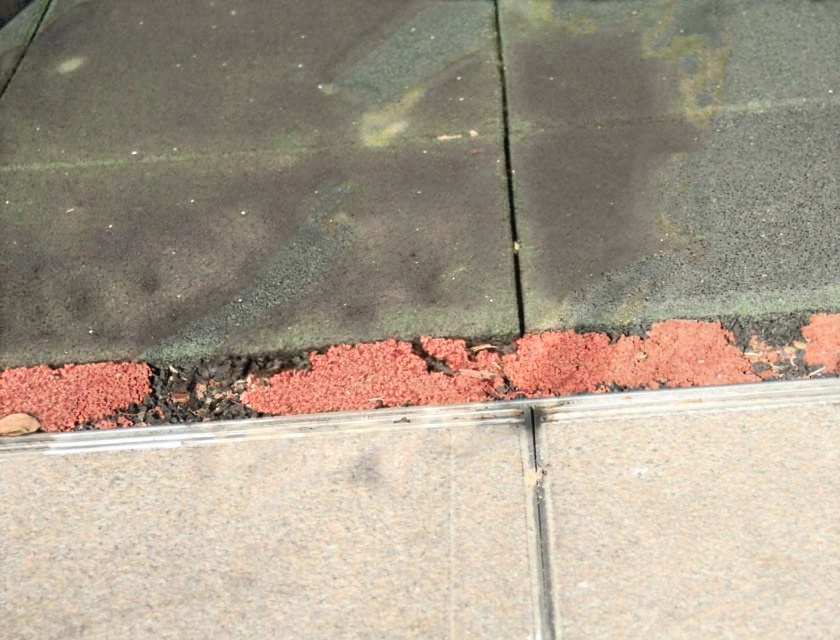}
    \caption{BMN~\cite{zhu_bijective_2022}}
  \end{subfigure}
  \begin{subfigure}[t]{.12\linewidth}
    \captionsetup{justification=centering, labelformat=empty, font=scriptsize}
    \includegraphics[width=1\linewidth,height=1\linewidth]{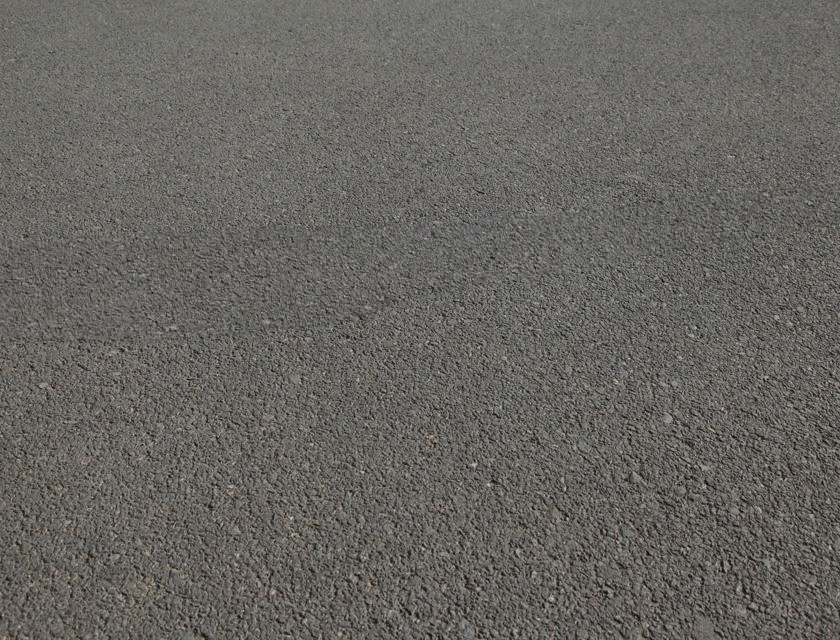}
    \includegraphics[width=1\linewidth,height=1\linewidth]{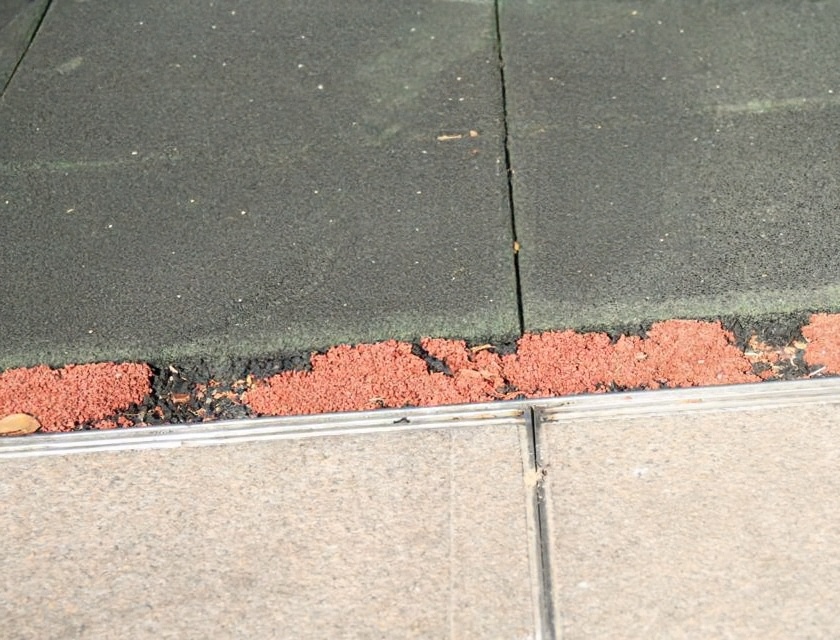}
    \caption{\tiny \ours~(Ours)}
  \end{subfigure}
  \begin{subfigure}[t]{.12\linewidth}
    \captionsetup{justification=centering, labelformat=empty, font=scriptsize}
    \includegraphics[width=1\linewidth,height=1\linewidth]{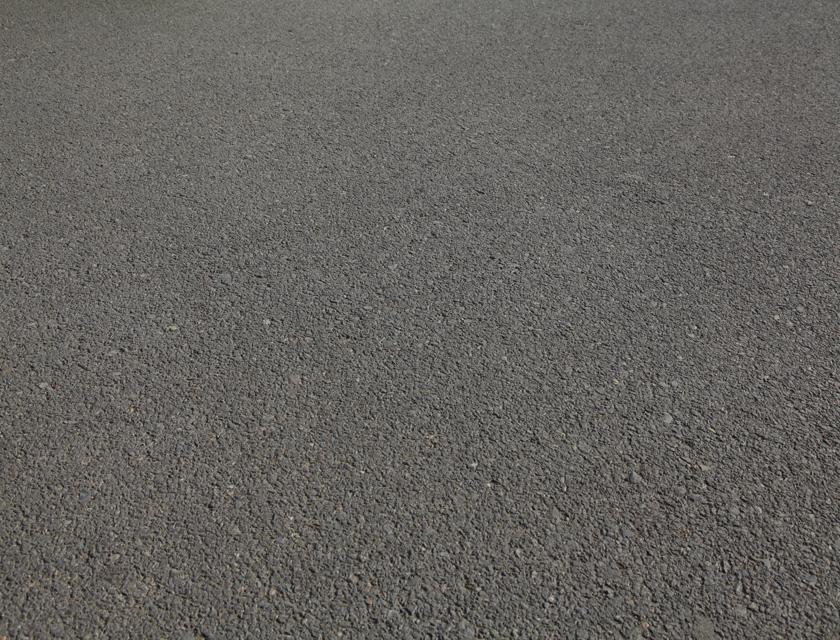}
    \includegraphics[width=1\linewidth,height=1\linewidth]{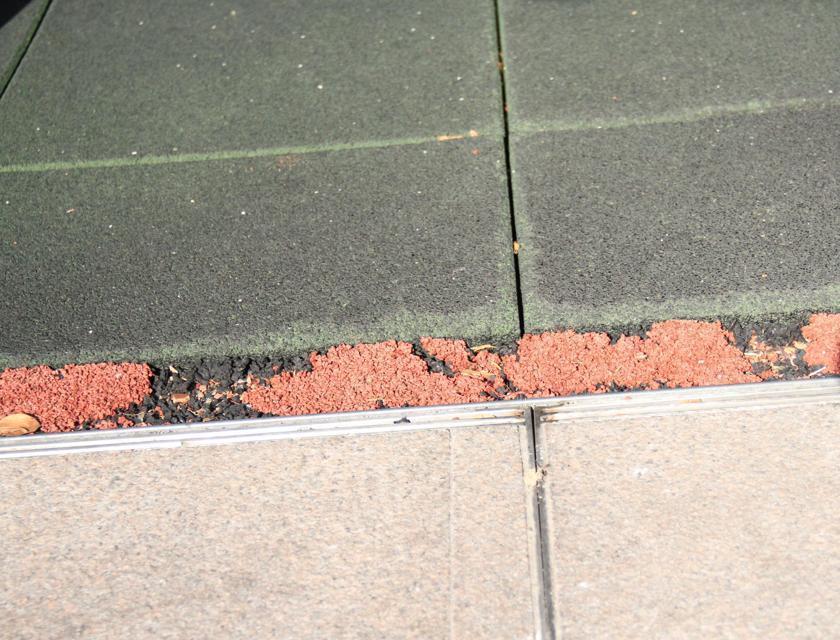}
    \caption{Ground Truth}
  \end{subfigure}
  \hfill
  \vspace{-.5\baselineskip}
  \caption{\textbf{Visual comparisons of the representative hard shadow cases from the SRD dataset.}}
  \vspace{-1\baselineskip}
  \label{fig:main_srd}
\end{figure*}

\begin{figure*}[h]
    \vspace{-1\baselineskip}
    \centering
    \begin{subfigure}[b]{.28\linewidth}
    \captionsetup{justification=centering, labelformat=empty}
    \includegraphics[width=\linewidth]{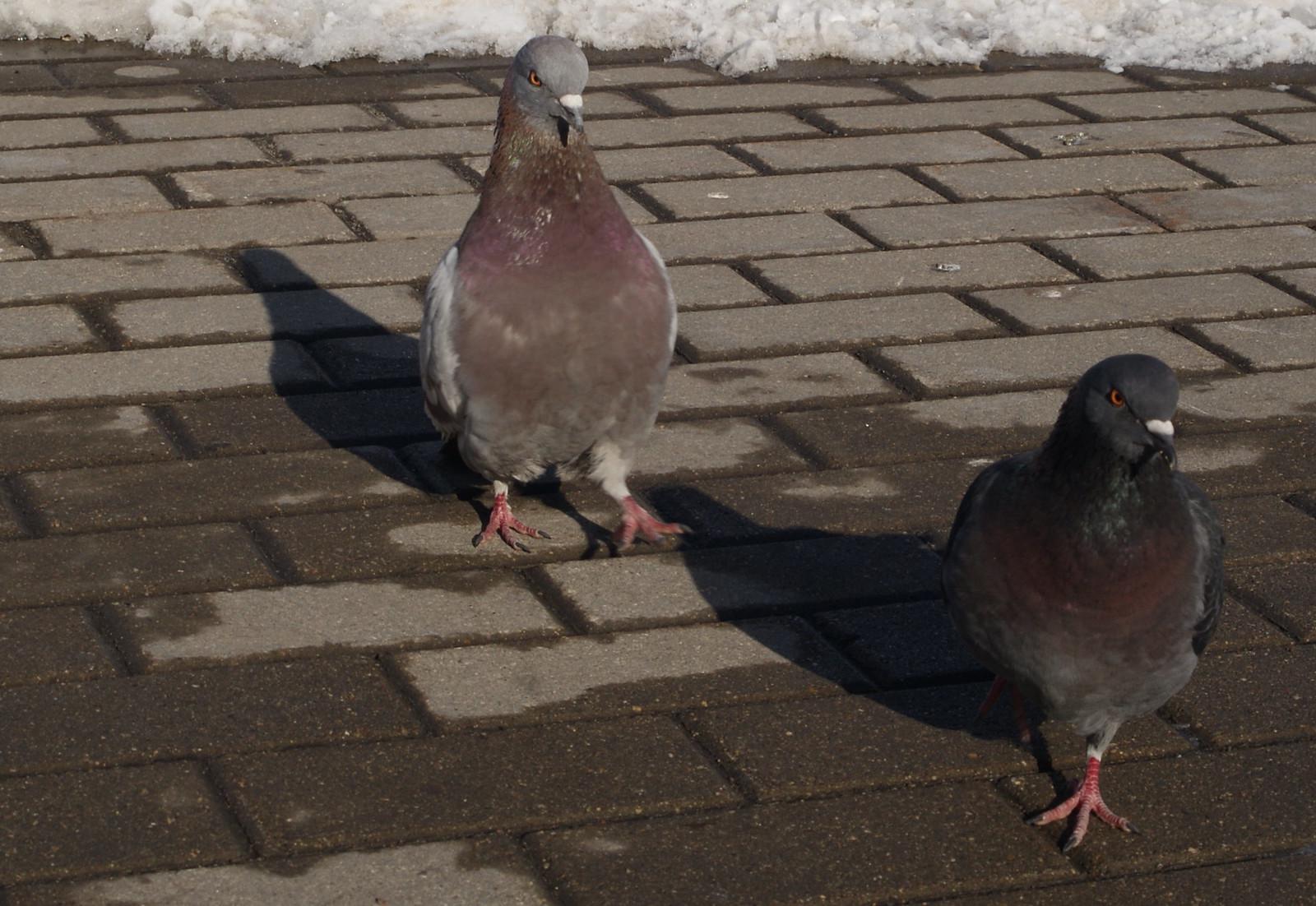}
    \caption{shadow image}
    \end{subfigure}
    \begin{subfigure}[b]{.58\linewidth}
    \centering
    \captionsetup{justification=centering, labelformat=empty}
    \includegraphics[width=.24\linewidth ]{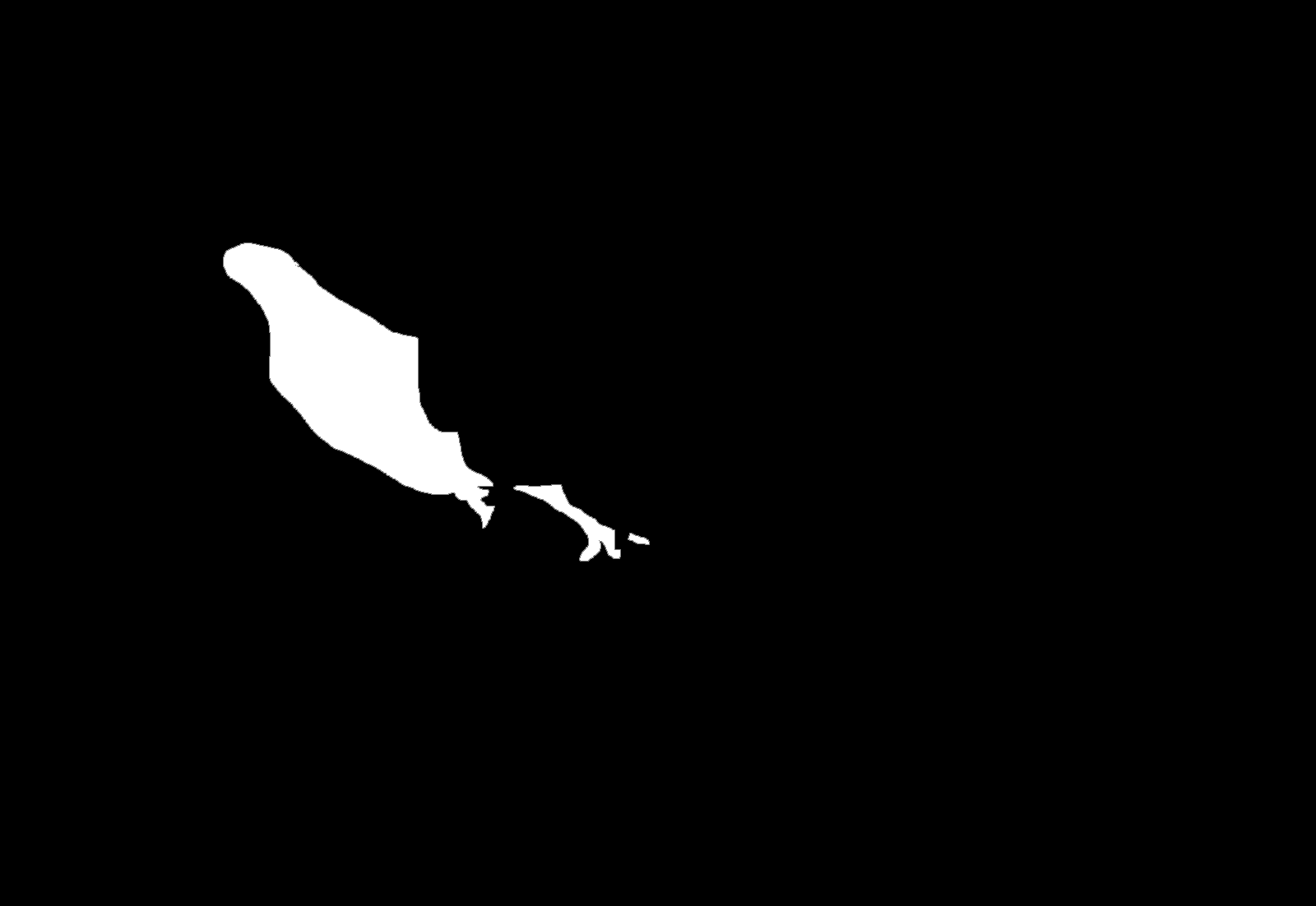}
    \includegraphics[width=.24\linewidth]{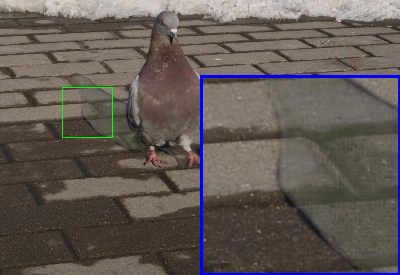}
    \includegraphics[width=.24\linewidth]{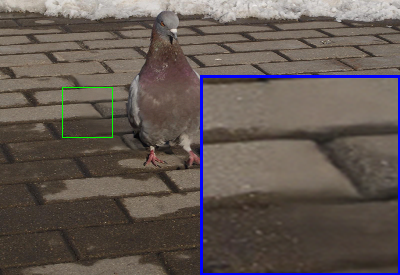}
    \includegraphics[width=.24\linewidth]{resources/real/ours_web-shadow0610_00.png}
    \hfill
    
    \includegraphics[width=.24\linewidth]{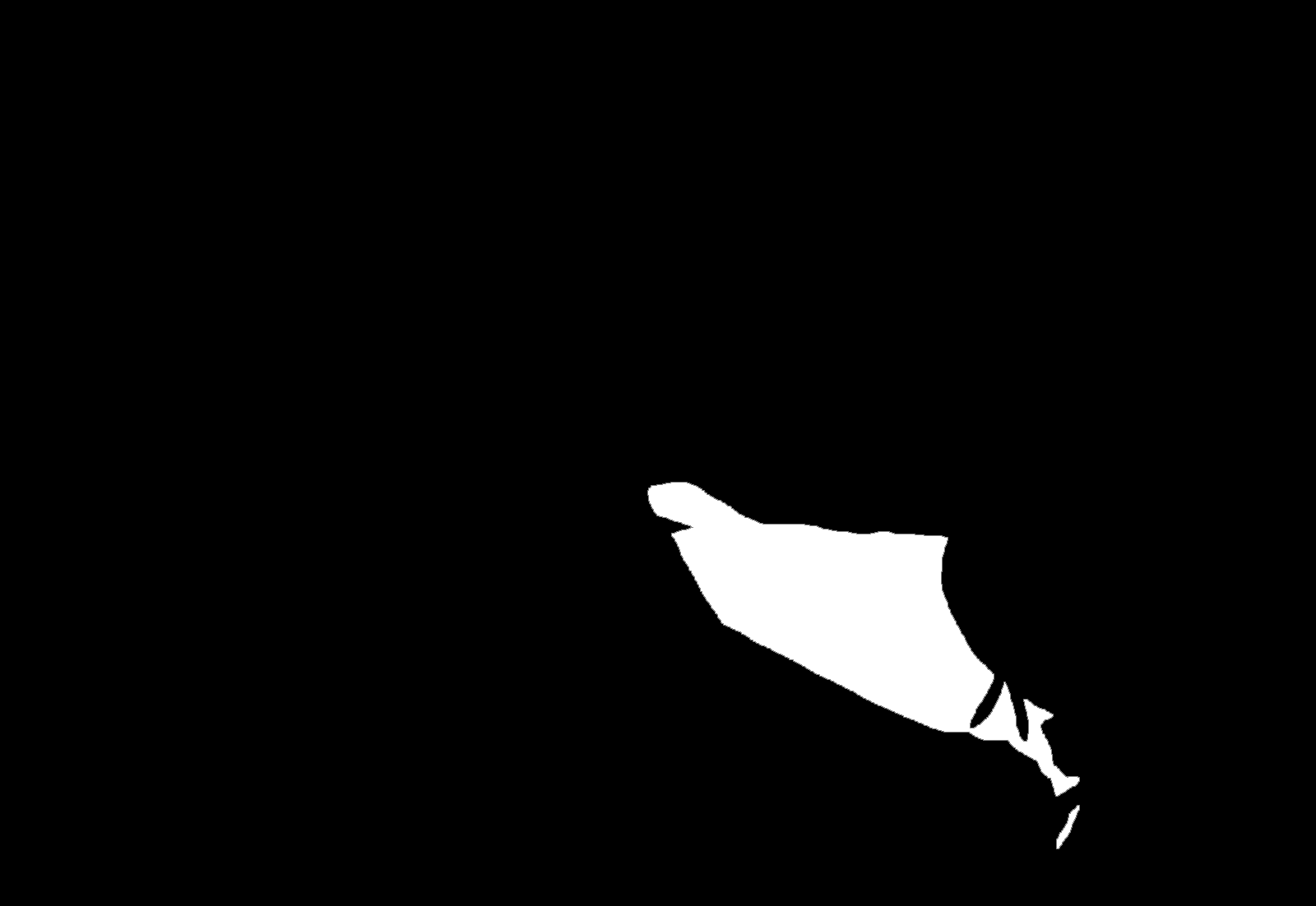}
    \includegraphics[width=.24\linewidth]{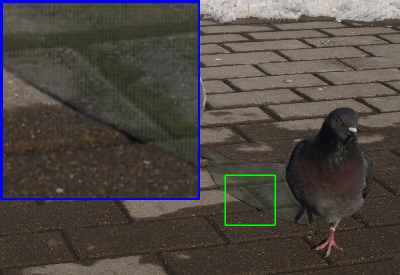}
    \includegraphics[width=.24\linewidth]{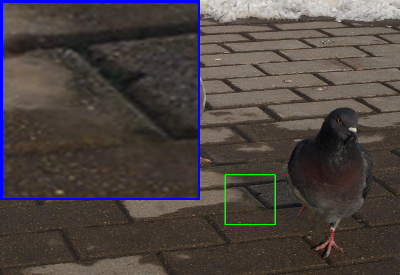}
    \includegraphics[width=.24\linewidth]{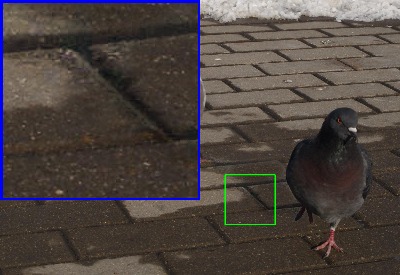}
    \hfill 
    \caption{\emph{shadow mask}, \emph{SG-ShadowNet result}, \emph{ours}, and \emph{shadow-free image} from L to R}
    \end{subfigure}
    \vspace{-.5\baselineskip}
    \caption{\textbf{Visual comparisons of the real instance shadow removal results on the DeSOBA dataset.}}
    \vspace{-1\baselineskip}
    \label{fig:real}
\end{figure*}

\section{Experiments}
We provide further implementation details, including the settings of the network and optimizer, in the supplemental.

\noindent\textbf{Shadow Removal Benchmarks.}
We conduct both quantitative and qualitative comparisons on three benchmarks: ISTD~\cite{wang2018stacked}, AISTD~\cite{le2019shadow}, and SRD~\cite{qu2017deshadownet}.
The ISTD dataset is a real-world shadow-removal benchmark that consists of 1,330 image triplets for training and 540 image triplets for testing. The image triplet includes the shadow image, shadow mask, and the corresponding shadow-free image.
The shadow mask is extracted from the binary difference between the shadow image and the shadow-free image.
The AISTD dataset uses the same scene as the ISTD dataset but avoids inconsistent color between the shadow and shadow-free image for accurate comparisons.
SRD contains different scenes and consists of 2,680 image pairs for training and 408 image pairs for testing.
Since SRD does not contain binary masks for the shadow regions, we follow the common practice and use the masks generated by Cu et al.~\cite{cun_towards_2020}.
For data processing, we empirically dilate all shadow masks in a kernel size of $k=21$ to address incomplete shadow masks.

\noindent\textbf{Instance-level Shadow Removal Benchmark.} We conduct various experiments with visual comparisons on shadow images collected from the internet.
The major difference between the above benchmarks and instance-shadow images is the number of shadows in the image, whereas the latter usually has more than one shadow instances.
The major collections of our instance-shadow images come from the shadow object association (SOBA) dataset~\cite{wang2020instance}.
We use the manually manipulated shadow-free images of the DESOBA dataset~\cite{hong2022shadow} as ground truths for removing shadows at the instance level.
For the network training, we synthesize shadow image triplets following the method proposed by Inoue et al.~\cite{inoue2020learning}.
Please see the supplement for a deep analysis of the synthesized data.

\subsection{Performance Evaluation}
We evaluate our proposed algorithm against state-of-the-art shadow-removal methods, including SP+M-Net~\cite{le2019shadow}, DHAN~\cite{cun_towards_2020}, Param+M+D-Net~\cite{le2020shadow}, G2R-ShadowNet~\cite{liu2021shadow}, Auto-Exposure~\cite{fu2021auto}, DC-ShadowNet~\cite{jin2021dc}, EMDN~\cite{zhu2022efficient}, BMN~\cite{zhu_bijective_2022}, and SG-ShadowNet~\cite{wan2022}, as well as two representative image restoration diffusion models, Palette Diffusion~\cite{lugmayr2022repaint}, and Repaint Diffusion~\cite{saharia2022palette}.
The evaluation metrics include the Root Mean Square Error (RMSE) between the shadow-free results and the ground truth in the LAB color space as well as the Peak Signal-to-Noise Ratio (PSNR) and structural similarity (SSIM) in the RGB space.
We also provide the metrics measured on the whole image and non-shadow region for reference.
Following previous methods~\cite{le2020shadow, fu2021auto, liu2021shadow}, we interpolate the results with a resolution of  $256\times 256$ for evaluation.
We also present the metrics evaluated on the shadow images for reference.

Tab.~\ref{tab:aistd} shows the quantitative results on the AISTD dataset.
Compared with the representative end-to-end learning-based methods, including EMDN~\cite{zhu2022efficient}, Auto-Exposure~\cite{fu2021auto} and DHAN~\cite{cun_towards_2020}, ours significantly outperforms them in all regions.
The performance gap between them and ours in the \emph{non-shadow} and \emph{full} regions further indicates the superiority of our model in generating high-quality textures of backgrounds.
As expected, the comparison between ours and the other generative methods, including BMN~\cite{zhu_bijective_2022}, DC-ShadowNet~\cite{jin2021dc}, and G2R-ShadowNet~\cite{liu2021shadow} demonstrate that our method achieves equal performance improvement in different regions. In contrast, the other methods fail in regions with specific textures.
The difference suggests the guidance effectiveness of our modeled latent feature, which is capable of balancing the unbalanced guidance from the surrounding non-shadow areas and shadow regions via the invariant loss function to aid the model in preserving texture and color.
The results shown in Tab.~\ref{tab:istd} on the SRD and ISTD datasets further demonstrate the superiority of our method over the others.

Visual comparison from the AISTD dataset in Fig.~\ref{fig:main_aistd} and SRD dataset in Fig.~\ref{fig:main_srd} further validates the effectiveness of our method.
As shown in Fig.~\ref{fig:main_aistd}, our method demonstrates robustness to imperfect shadow mask inputs and preserves the textures as well as removing other subtle shadow effects.

\subsection{Instance Shadow Removal Evaluation}
For real-world applications, shadows cast by objects in the scene are usually instance-level; thus, preserving the other shadows while accurately removing the target instance shadow is crucial.
Here, we compare our method with the most recent shadow removal work SG-ShadowNet~\cite{wan2022} to demonstrate the generalizability of our method, where we finetuned it with the same dataset synthesized for our experiments.
Sample results are shown in Fig.~\ref{fig:real}.
Compared with the SG-ShadowNet, ours thoroughly removes the shadow from the images.
As far as we know, this is the first work to demonstrate the applicability of instance shadow removal.

\subsection{Ablation Study and Analysis}
\label{sec:ab}

\begin{table}[htbp!]
    \caption{\textbf{Effects of different types of strategies for addressing the posterior collapse in diffusion models.} We only show shadow region results that are distinguishing.}
    \label{tab:abposterior}
    \centering\small
    \resizebox{1\linewidth}{!}{
    \begin{tabular}{l|c|ccc}
    \toprule
    & & \multicolumn{3}{c}{AISTD dataset} \\
    \midrule
    Settings & Region & RMSE $\downarrow$ & PSNR $\uparrow$ & SSIM $\uparrow$ \\
    \midrule
    $\mathrm{w/o.~fusion}$ & \emph{shadow} & 6.75 & 36.34 & \textbf{0.990} \\
    $\mathrm{lagged~posterior}$ & \emph{shadow} & \emph{\underline{6.65}} & \emph{\underline{36.27}} & \textbf{0.990} \\
    \rowcolor{Gray} $\mathrm{dense~fusion}$ (Ours) & \emph{shadow} & \textbf{5.92} & \textbf{37.70} & \textbf{0.990} \\
    \bottomrule
    \end{tabular}
    }
\end{table}

\noindent\textbf{Effects of the DLVF module.}
In Tab.~\ref{tab:abposterior}, we investigate the effectiveness of the proposed DLVF module.
We use two alternative methods for comparison: a $\mathrm{lagged~posterior}$ approach~\cite{he2019lagging} for addressing the posterior collapse, which aggressively optimizes the diffusion network before optimizing the latent feature encoder, and the baseline approach that uses a diffusion network without the fusion strategy.  
The results show that $\mathrm{lagged~posterior}$ is less effective, with only a slight improvement margin over the baseline, which could be due to the large complexity of diffusion models and difficulty in training. 
In contrast, our proposed $\mathrm{dense~fusion}$ scheme outperforms the baseline by a margin of 0.83 RMSE.
Moreover, we visually demonstrate its effectiveness by showing the correspondence of the denoised results $\mathbf{y}_{t-1}$ and $\mathbf{y}_t$ in Fig.~\ref{fig:dpffv}.
These results validate the idea proposed in our DLVF, \ie, fusing more noise features into each block of the diffusion network is a promising approach for alleviating the local optima of training diffusion models.

\noindent\textbf{Effects of Latent Feature Space Guidance.}
\label{sec:lfsp}
Tab.~\ref{tab:abcondition} compares different types of diffusion model guidance for removing shadows, including \textbf{(a)} estimated invariant color map, \textbf{(b)} estimated coarse de-shadowed image, \textbf{(c)} learned latent feature space without invariant loss, and \textbf{(d)} learn latent feature space with our two-stage learning.
Our proposed setting achieves a significantly better numerical performance compared to the others.
Interestingly, the guidance (\ie $\mathrm{coarse~deshadowed}$) that provides the most pixel information performs worse than the guidance (\ie $\mathrm{invariant~color~map}$) that only provides a simple color map.
After deeply looking at their visualization in Fig.~\ref{fig:latent}, we observe that even coarse de-shadowed image still contains shadow boundary that may mislead the diffusion models, while the color map omits most shadow features, which demonstrates that only encapsulating shadow-free features is crucial for improving the performance.
Correspondingly, our latent feature is acquired by minimizing the difference between the encoded features of shadow and shadow-free images, which implicitly omits shadow features, and it contains more perceptual features because we optimize it together with diffusion models for learning denoising. Therefore, it guides diffusion models with more shadow-free features and outperforms the compared methods.

\begin{table}[ht]
    \caption{\textbf{Effects of different types of diffusion model guidance that provides shadow-free priors.}}
    \label{tab:abcondition}
    \centering\small
    \resizebox{\linewidth}{!}{
    \begin{tabular}{l|c|ccc}
    \toprule
    & & \multicolumn{3}{c}{AISTD dataset} \\
    \midrule
    Settings & Region & RMSE $\downarrow$ & PSNR $\uparrow$ & SSIM $\uparrow$ \\
    \midrule
    $\mathrm{invariant~color~map}$~\cite{zhu_bijective_2022} & \emph{shadow} & \emph{\underline{7.72}} & 36.24 & 0.986  \\
    $\mathrm{coarse~deshadowed}$~\cite{wan2022} & \emph{shadow}  & 8.03 & 35.74 & \emph{\underline{0.988}}  \\
    \midrule
    $\bar{\mathcal{E}}_\theta(\bx, m)$ & \emph{shadow} & 7.59 & \emph{\underline{36.65}}  & 0.984  \\
    \rowcolor{Gray}  $\mathcal{E}_\theta(\bx, m)$ (Ours) & \emph{shadow} & \textbf{5.92} & \textbf{37.70} & \textbf{0.990} \\
    \bottomrule
    \end{tabular}
    }
\end{table}

\begin{table}[htbp]
\caption{\textbf{Complexity comparisons of our distilled lighter model with the accelerated diffusion solver}.}
\resizebox{\linewidth}{!}{
\begin{tabular}{l | c c | c c c} 
\toprule
& & & \multicolumn{3}{c}{AISTD dataset (RMSE)} \\
\midrule
Method & params & time & shadow & non-shadow & all  \\
\midrule
BMN & \textbf{0.4M} & 1.69s  & 5.69 & 2.52 & 3.02 \\
G2R-ShadowNet & 22.8M & 0.36s & 7.38 & 3.00 & 3.69  \\
\rowcolor{Gray} \ours~(Ours) & 82.6M & 2.76s & \textbf{5.15} & 2.47 & \textbf{2.90}  \\
\ours~(Distilled) & 25.5M & \textbf{0.24s} & 5.21 & \textbf{2.34} & 2.94   \\
\bottomrule
\end{tabular}     
}
\label{tab:ef}
\end{table}

\noindent\textbf{Model Complexity Analysis.}
Our work focuses on adapting diffusion models to address shadow removal, and therefore we prioritize exploration over analysis of model complexity and inference time. However, we demonstrate the feasibility of our approach in terms of model complexity and inference time using advanced technologies such as those proposed in \cite{lu2022knowledge} for reducing model parameters without sacrificing performance, and \cite{lu2022dpm} for accelerating diffusion sampling in Tab.~\ref{tab:ef}.
We find that even with similar settings, our lighter model outperforms compared methods with better restoration performance and is also faster.

\noindent\textbf{ShadowDiffusion Comparison.}
Given the similarity between the recent ShadowDiffusion~\cite{guo2023shadowdiffusion} (SD) and our method, which both characterize the shadow-free image distribution by conditioning diffusion models, ours further explores shadow removal at the instance level without any modifications to the model.
The other difference is majorly in the method complexity, \ie, tackling challenges such as color-mixing and collapse, often arising from direct conditioning on shadow images.
SD integrates a pre-trained shadow removal network.
In contrast, ours models the shadow-free priors through two-stage learning and mitigates collapse using dense fusion modules.
Tab.~\ref{tab:sd} demonstrates our efficiency in the shadow region of AISTD.
($\ddag$We use their evaluation settings that give different numbers.)\\
\begin{table}[htbp]
\vspace{-1.5\baselineskip}
\caption{\textbf{Quantitative comparison with ShadowDiffusion.}}
\label{tab:sd}
\resizebox{\linewidth}{!}{
\begin{tabular}{l|ccc}
    \toprule
    Model & params$(\downarrow)$ & time$(\downarrow)$ & RMSE$(\downarrow)$ \\
    \midrule
    ShadowDiffusion~\cite{guo2023shadowdiffusion} & 602.6M* & 7.54s* & 4.9$\dag$ \\
    LFG-Diffusion (Ours) & 82.6M & 2.76s & 5.0$\ddag$  \\
    \bottomrule
\end{tabular}
}
\end{table}

\section{Conclusion}
In this work, we introduced a novel class of diffusion models that significantly outperform existing shadow removal methods at the general and instance level. By incorporating a latent feature space that captures perceptual shadow-free priors, we have shown that this guidance can mitigate the unbalanced guidance issue between shadow and non-shadow areas during restoration. Furthermore, we have proposed the DLVF module, which strengthens the connections between latent variable of noise and the diffusion network to prevent local optimum. Our comprehensive evaluations and analyses have demonstrated the superior effectiveness of our method compared to existing state-of-the-art shadow removal methods. We believe that our proposed diffusion model-based technique has the potential to be applied to other similar ill-posed low-level problems.

{\small
\bibliographystyle{ieee_fullname}
\bibliography{shadow_removal, diffusion}

\begin{thebibliography}{10}\itemsep=-1pt

\bibitem{arbel2010shadow}
Eli Arbel and Hagit Hel-Or.
\newblock Shadow removal using intensity surfaces and texture anchor points.
\newblock {\em IEEE TPAMI}, 2010.

\bibitem{chen2021canet}
Zipei Chen, Chengjiang Long, Ling Zhang, and Chunxia Xiao.
\newblock Canet: A context-aware network for shadow removal.
\newblock In {\em ICCV}, 2021.

\bibitem{choi2022perception}
Jooyoung Choi, Jungbeom Lee, Chaehun Shin, Sungwon Kim, Hyunwoo Kim, and
  Sungroh Yoon.
\newblock Perception prioritized training of diffusion models.
\newblock In {\em CVPR}, 2022.

\bibitem{cun_towards_2020}
Xiaodong Cun, Chi-Man Pun, and Cheng Shi.
\newblock Towards ghost-free shadow removal via dual hierarchical aggregation
  network and shadow matting {GAN}.
\newblock In {\em {AAAI}}, 2020.

\bibitem{dhariwal2021diffusion}
Prafulla Dhariwal and Alexander Nichol.
\newblock Diffusion models beat gans on image synthesis.
\newblock In {\em NeurIPS}, 2021.

\bibitem{dieng2019avoiding}
Adji~B Dieng, Yoon Kim, Alexander~M Rush, and David~M Blei.
\newblock Avoiding latent variable collapse with generative skip models.
\newblock In {\em International Conference on Artificial Intelligence and
  Statistics}, 2019.

\bibitem{ding2019argan}
Bin Ding, Chengjiang Long, Ling Zhang, and Chunxia Xiao.
\newblock Argan: Attentive recurrent generative adversarial network for shadow
  detection and removal.
\newblock In {\em ICCV}, 2019.

\bibitem{finlayson1996color}
Graham~D Finlayson, Subho~S Chatterjee, and Brian~V Funt.
\newblock Color angular indexing.
\newblock In {\em ECCV}, 1996.

\bibitem{finlayson2009entropy}
Graham~D Finlayson, Mark~S Drew, and Cheng Lu.
\newblock Entropy minimization for shadow removal.
\newblock {\em IJCV}, 2009.

\bibitem{finlayson2005removal}
Graham~D Finlayson, Steven~D Hordley, Cheng Lu, and Mark~S Drew.
\newblock On the removal of shadows from images.
\newblock {\em IEEE TPAMI}, 2005.

\bibitem{fu2019cyclical}
Hao Fu, Chunyuan Li, Xiaodong Liu, Jianfeng Gao, Asli Celikyilmaz, and Lawrence
  Carin.
\newblock Cyclical annealing schedule: A simple approach to mitigating kl
  vanishing.
\newblock {\em arXiv preprint arXiv:1903.10145}, 2019.

\bibitem{fu2021auto}
Lan Fu, Changqing Zhou, Qing Guo, Felix Juefei-Xu, Hongkai Yu, Wei Feng, Yang
  Liu, and Song Wang.
\newblock Auto-exposure fusion for single-image shadow removal.
\newblock In {\em CVPR}, 2021.

\bibitem{funt1995color}
Brian~V. Funt and Graham~D. Finlayson.
\newblock Color constant color indexing.
\newblock {\em IEEE TPAMI}, 1995.

\bibitem{guo2023shadowformer}
Lanqing Guo, Siyu Huang, Ding Liu, Hao Cheng, and Bihan Wen.
\newblock Shadowformer: Global context helps image shadow removal.
\newblock {\em arXiv preprint arXiv:2302.01650}, 2023.

\bibitem{guo2023shadowdiffusion}
Lanqing Guo, Chong Wang, Wenhan Yang, Siyu Huang, Yufei Wang, Hanspeter
  Pfister, and Bihan Wen.
\newblock Shadowdiffusion: When degradation prior meets diffusion model for
  shadow removal.
\newblock In {\em Proceedings of the IEEE/CVF Conference on Computer Vision and
  Pattern Recognition}, pages 14049--14058, 2023.

\bibitem{guo2012paired}
Ruiqi Guo, Qieyun Dai, and Derek Hoiem.
\newblock Paired regions for shadow detection and removal.
\newblock {\em IEEE TPMAI}, 2012.

\bibitem{he2019lagging}
Junxian He, Daniel Spokoyny, Graham Neubig, and Taylor Berg-Kirkpatrick.
\newblock Lagging inference networks and posterior collapse in variational
  autoencoders.
\newblock {\em arXiv preprint arXiv:1901.05534}, 2019.

\bibitem{ho_denoising_2020}
Jonathan Ho, Ajay Jain, and Pieter Abbeel.
\newblock Denoising diffusion probabilistic models.
\newblock {\em NeurIPS}, 2020.

\bibitem{ho2022classifier}
Jonathan Ho and Tim Salimans.
\newblock Classifier-free diffusion guidance.
\newblock {\em arXiv preprint arXiv:2207.12598}, 2022.

\bibitem{hong2022shadow}
Yan Hong, Li Niu, and Jianfu Zhang.
\newblock Shadow generation for composite image in real-world scenes.
\newblock In {\em AAAI}, 2022.

\bibitem{hou2022face}
Andrew Hou, Michel Sarkis, Ning Bi, Yiying Tong, and Xiaoming Liu.
\newblock Face relighting with geometrically consistent shadows.
\newblock In {\em Proceedings of the IEEE/CVF Conference on Computer Vision and
  Pattern Recognition}, pages 4217--4226, 2022.

\bibitem{hu2019mask}
Xiaowei Hu, Yitong Jiang, Chi-Wing Fu, and Pheng-Ann Heng.
\newblock Mask-shadowgan: Learning to remove shadows from unpaired data.
\newblock In {\em ICCV}, 2019.

\bibitem{inoue2020learning}
Naoto Inoue and Toshihiko Yamasaki.
\newblock Learning from synthetic shadows for shadow detection and removal.
\newblock {\em IEEE TCSVT}, 2020.

\bibitem{jin2021dc}
Yeying Jin, Aashish Sharma, and Robby~T Tan.
\newblock Dc-shadownet: Single-image hard and soft shadow removal using
  unsupervised domain-classifier guided network.
\newblock In {\em ICCV}, 2021.

\bibitem{kingma2013auto}
Diederik~P Kingma and Max Welling.
\newblock Auto-encoding variational bayes.
\newblock {\em arXiv preprint arXiv:1312.6114}, 2013.

\bibitem{le2019shadow}
Hieu Le and Dimitris Samaras.
\newblock Shadow removal via shadow image decomposition.
\newblock In {\em ICCV}, 2019.

\bibitem{le2020shadow}
Hieu Le and Dimitris Samaras.
\newblock From shadow segmentation to shadow removal.
\newblock In {\em {ECCV}}, 2020.

\bibitem{liu2021shadow}
Zhihao Liu, Hui Yin, Xinyi Wu, Zhenyao Wu, Yang Mi, and Song Wang.
\newblock From shadow generation to shadow removal.
\newblock In {\em CVPR}, 2021.

\bibitem{liu_shadow_2021}
Zhihao Liu, Hui Yin, Xinyi Wu, Zhenyao Wu, Yang Mi, and Song Wang.
\newblock From shadow generation to shadow removal.
\newblock In {\em {CVPR}}, 2021.

\bibitem{lu2022knowledge}
Chengqiang Lu, Jianwei Zhang, Yunfei Chu, Zhengyu Chen, Jingren Zhou, Fei Wu,
  Haiqing Chen, and Hongxia Yang.
\newblock Knowledge distillation of transformer-based language models
  revisited.
\newblock {\em arXiv preprint arXiv:2206.14366}, 2022.

\bibitem{lu2022dpm}
Cheng Lu, Yuhao Zhou, Fan Bao, Jianfei Chen, Chongxuan Li, and Jun Zhu.
\newblock Dpm-solver: A fast ode solver for diffusion probabilistic model
  sampling in around 10 steps.
\newblock {\em arXiv preprint arXiv:2206.00927}, 2022.

\bibitem{lugmayr2022repaint}
Andreas Lugmayr, Martin Danelljan, Andres Romero, Fisher Yu, Radu Timofte, and
  Luc Van~Gool.
\newblock Repaint: Inpainting using denoising diffusion probabilistic models.
\newblock In {\em Proceedings of the IEEE/CVF Conference on Computer Vision and
  Pattern Recognition}, pages 11461--11471, 2022.

\bibitem{ma2008shadow}
Haijian Ma, Qiming Qin, and Xinyi Shen.
\newblock Shadow segmentation and compensation in high resolution satellite
  images.
\newblock In {\em IGARSS 2008-2008 IEEE International Geoscience and Remote
  Sensing Symposium}, volume~2, pages II--1036, 2008.

\bibitem{meng2021sdedit}
Chenlin Meng, Yutong He, Yang Song, Jiaming Song, Jiajun Wu, Jun-Yan Zhu, and
  Stefano Ermon.
\newblock Sdedit: Guided image synthesis and editing with stochastic
  differential equations.
\newblock In {\em International Conference on Learning Representations}, 2021.

\bibitem{preechakul2022diffusion}
Konpat Preechakul, Nattanat Chatthee, Suttisak Wizadwongsa, and Supasorn
  Suwajanakorn.
\newblock Diffusion autoencoders: Toward a meaningful and decodable
  representation.
\newblock In {\em Proceedings of the IEEE/CVF Conference on Computer Vision and
  Pattern Recognition}, pages 10619--10629, 2022.

\bibitem{qu2017deshadownet}
Liangqiong Qu, Jiandong Tian, Shengfeng He, Yandong Tang, and Rynson~WH Lau.
\newblock Deshadownet: A multi-context embedding deep network for shadow
  removal.
\newblock In {\em CVPR}, 2017.

\bibitem{ramesh2022hierarchical}
Aditya Ramesh, Prafulla Dhariwal, Alex Nichol, Casey Chu, and Mark Chen.
\newblock Hierarchical text-conditional image generation with clip latents.
\newblock {\em arXiv preprint arXiv:2204.06125}, 2022.

\bibitem{rombach2022high}
Robin Rombach, Andreas Blattmann, Dominik Lorenz, Patrick Esser, and Bj{\"o}rn
  Ommer.
\newblock High-resolution image synthesis with latent diffusion models.
\newblock In {\em CVPR}, 2022.

\bibitem{saharia2022palette}
Chitwan Saharia, William Chan, Huiwen Chang, Chris Lee, Jonathan Ho, Tim
  Salimans, David Fleet, and Mohammad Norouzi.
\newblock Palette: Image-to-image diffusion models.
\newblock In {\em ACM SIGGRAPH 2022 Conference Proceedings}, pages 1--10, 2022.

\bibitem{saharia2022photorealistic}
Chitwan Saharia, William Chan, Saurabh Saxena, Lala Li, Jay Whang, Emily
  Denton, Seyed Kamyar~Seyed Ghasemipour, Burcu~Karagol Ayan, S~Sara Mahdavi,
  Rapha~Gontijo Lopes, et~al.
\newblock Photorealistic text-to-image diffusion models with deep language
  understanding.
\newblock {\em arXiv preprint arXiv:2205.11487}, 2022.

\bibitem{saharia2022image}
Chitwan Saharia, Jonathan Ho, William Chan, Tim Salimans, David~J Fleet, and
  Mohammad Norouzi.
\newblock Image super-resolution via iterative refinement.
\newblock {\em IEEE TPAMI}, 2022.

\bibitem{sohl2015deep}
Jascha Sohl-Dickstein, Eric Weiss, Niru Maheswaranathan, and Surya Ganguli.
\newblock Deep unsupervised learning using nonequilibrium thermodynamics.
\newblock In {\em ICML}, 2015.

\bibitem{song2020denoising}
Jiaming Song, Chenlin Meng, and Stefano Ermon.
\newblock Denoising diffusion implicit models.
\newblock {\em arXiv preprint arXiv:2010.02502}, 2020.

\bibitem{stricker1995similarity}
Markus~Andreas Stricker and Markus Orengo.
\newblock Similarity of color images.
\newblock In {\em Storage and retrieval for image and video databases III},
  volume 2420, pages 381--392. SPiE, 1995.

\bibitem{tolstikhin2017wasserstein}
Ilya Tolstikhin, Olivier Bousquet, Sylvain Gelly, and Bernhard Schoelkopf.
\newblock Wasserstein auto-encoders.
\newblock {\em arXiv preprint arXiv:1711.01558}, 2017.

\bibitem{vaswani2017attention}
Ashish Vaswani, Noam Shazeer, Niki Parmar, Jakob Uszkoreit, Llion Jones,
  Aidan~N Gomez, {\L}ukasz Kaiser, and Illia Polosukhin.
\newblock Attention is all you need.
\newblock In {\em NeurIPS}, 2017.

\bibitem{wan2022}
Jin Wan, Hui Yin, Zhenyao Wu, Xinyi Wu, Yanting Liu, and Song Wang.
\newblock Style-guided shadow removal.
\newblock In {\em ECCV}, 2022.

\bibitem{wan2022crformer}
Jin Wan, Hui Yin, Zhenyao Wu, Xinyi Wu, Zhihao Liu, and Song Wang.
\newblock Crformer: A cross-region transformer for shadow removal.
\newblock {\em arXiv preprint arXiv:2207.01600}, 2022.

\bibitem{wang2018stacked}
Jifeng Wang, Xiang Li, and Jian Yang.
\newblock Stacked conditional generative adversarial networks for jointly
  learning shadow detection and shadow removal.
\newblock In {\em CVPR}, 2018.

\bibitem{wang2020instance}
Tianyu Wang, Xiaowei Hu, Qiong Wang, Pheng-Ann Heng, and Chi-Wing Fu.
\newblock Instance shadow detection.
\newblock In {\em CVPR}, 2020.

\bibitem{yang2017improved}
Zichao Yang, Zhiting Hu, Ruslan Salakhutdinov, and Taylor Berg-Kirkpatrick.
\newblock Improved variational autoencoders for text modeling using dilated
  convolutions.
\newblock In {\em International conference on machine learning}, pages
  3881--3890. PMLR, 2017.

\bibitem{zhang2021no}
Edward Zhang, Ricardo Martin-Brualla, Janne Kontkanen, and Brian~L Curless.
\newblock No shadow left behind: Removing objects and their shadows using
  approximate lighting and geometry.
\newblock In {\em CVPR}, 2021.

\bibitem{zhang2020ris}
Ling Zhang, Chengjiang Long, Xiaolong Zhang, and Chunxia Xiao.
\newblock Ris-gan: Explore residual and illumination with generative
  adversarial networks for shadow removal.
\newblock In {\em AAAI}, 2020.

\bibitem{zhang2018improving}
Wuming Zhang, Xi Zhao, Jean-Marie Morvan, and Liming Chen.
\newblock Improving shadow suppression for illumination robust face
  recognition.
\newblock {\em IEEE TPAMI}, 2018.

\bibitem{zhao2017infovae}
Shengjia Zhao, Jiaming Song, and Stefano Ermon.
\newblock Infovae: Information maximizing variational autoencoders.
\newblock {\em arXiv preprint arXiv:1706.02262}, 2017.

\bibitem{zhu_bijective_2022}
Yurui Zhu, Jie Huang, Xueyang Fu, Feng Zhao, Qibin Sun, and Zheng-Jun Zha.
\newblock Bijective {Mapping} {Network} for {Shadow} {Removal}.
\newblock In {\em {CVPR}}, 2022.

\bibitem{zhu2022efficient}
Yurui Zhu, Zeyu Xiao, Yanchi Fang, Xueyang Fu, Zhiwei Xiong, and Zheng-Jun Zha.
\newblock Efficient model-driven network for shadow removal.
\newblock In {\em AAAI}, 2022.

\end{thebibliography}
}

\newpage
\appendix
\onecolumn

\section{Demo}
In this supplemental, we have provided a recording of our demo in use, named ``screenshot-demo.mp4''.
We strongly encourage reviewers to watch the recording to observe the results of our model on instance shadow removal.
In our demo, we use two different types of inference, \ie, {\color{red} \emph{Removal}} and {\color{blue} \emph{Quick Removal}}, to process shadow images.
The {\color{red} \emph{Removal}} method removes shadows in a $256\times 256$ sliding window manner, which preserves most of the details under the shadow.
The  {\color{blue} \emph{Quick Removal}} method first downsamples the shadow image into $512\times 512$ resolution and then removes shadows by denoising, which is significantly faster but blurrier than the first method.
Our demo allows for incomplete shadow masks by pre-processing masks with dilation kernels in different sizes.

\begin{figure}[htbp]
    \centering
    \includegraphics[width=.7\linewidth]{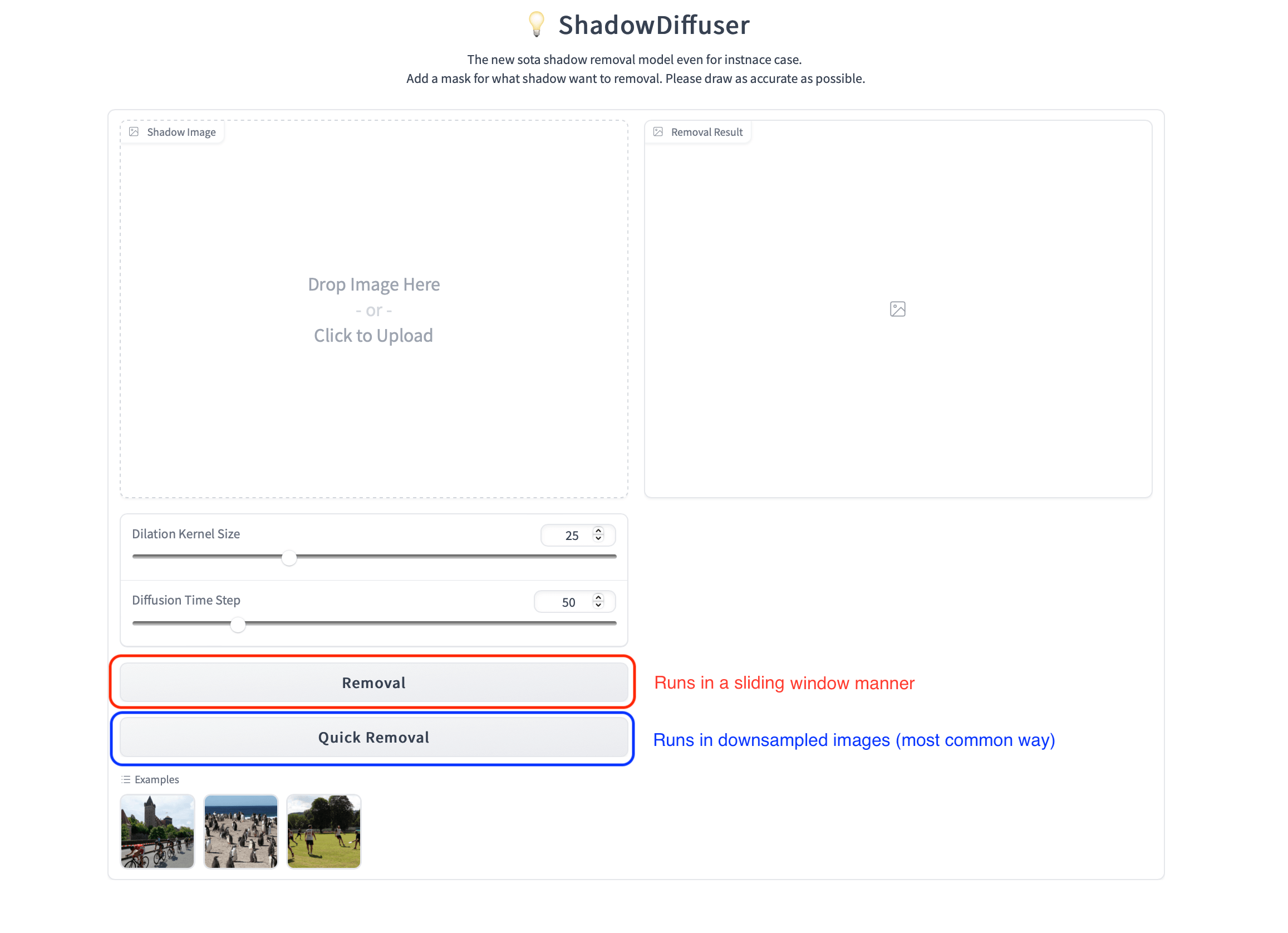}
    \includegraphics[width=.64\linewidth]{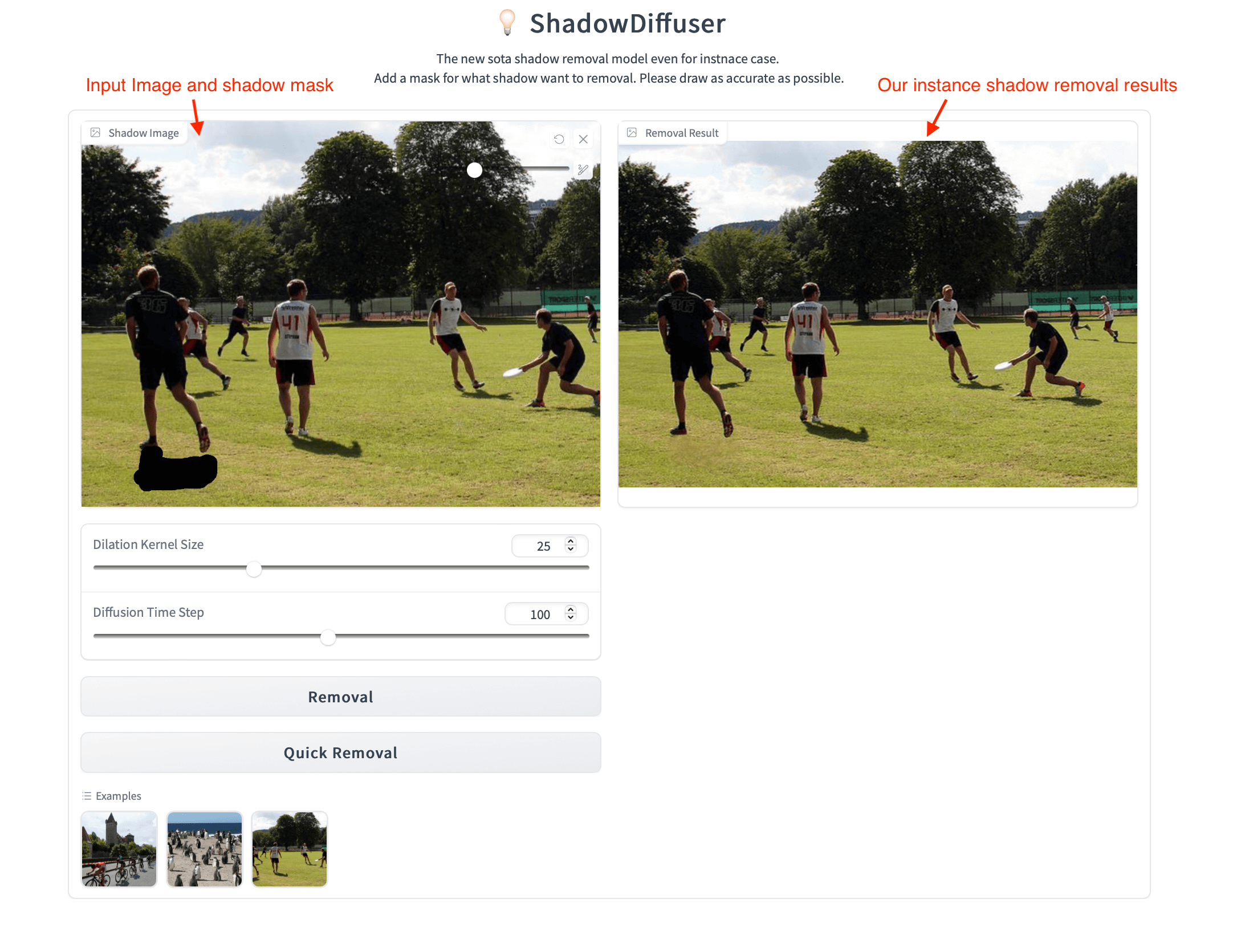}
    \caption{Screenshot of our demo for instance shadow removal.}
\end{figure}

\section{Implementation}
\begin{figure}[htbp]
    \centering
    \includegraphics[width=.5\linewidth]{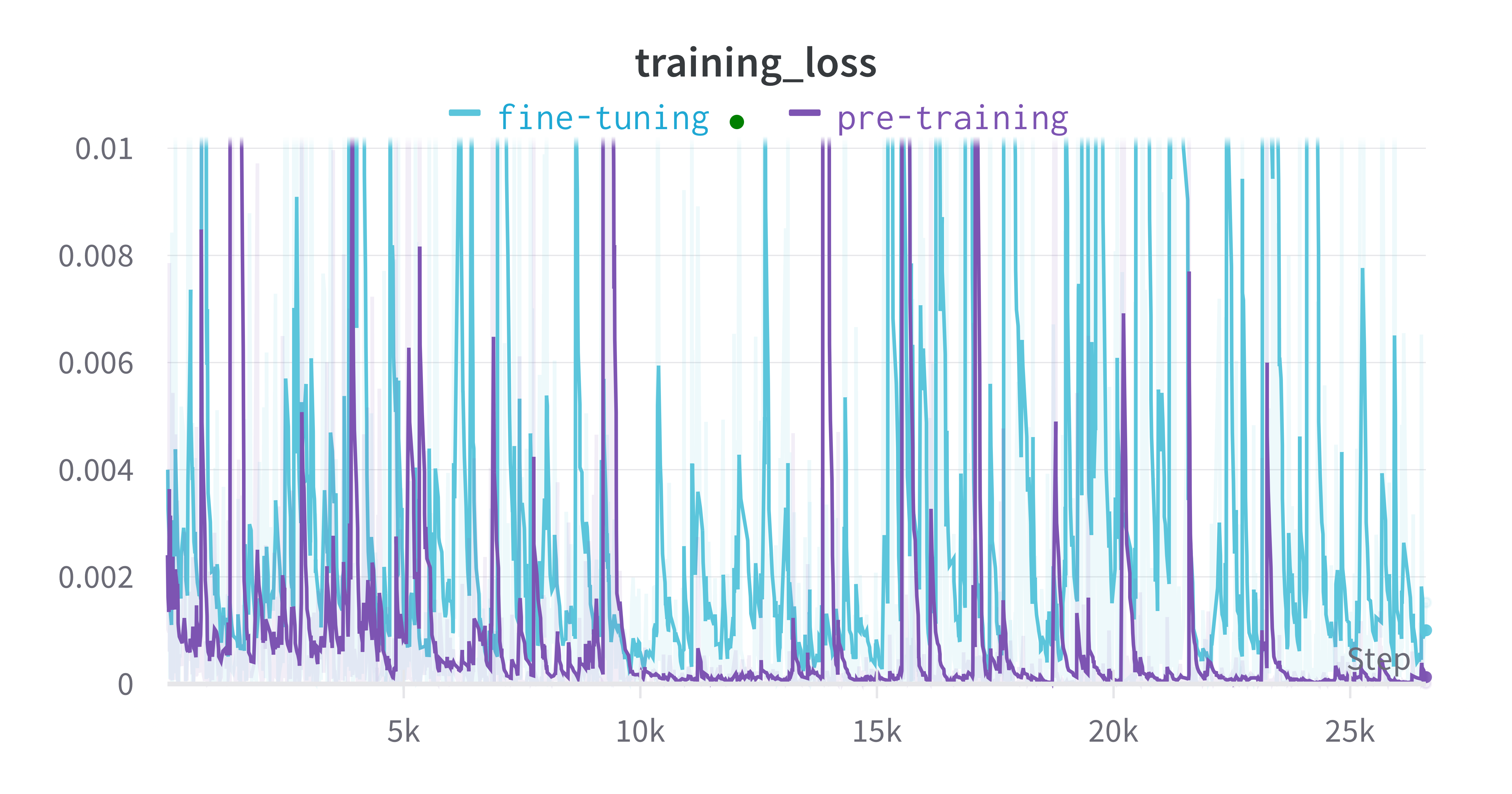}
    \vspace{-1\baselineskip}
    \caption{Loss curves of the diffusion network}
    \label{fig:lossd}
\end{figure}

\begin{figure}[htbp]
    \centering
    \includegraphics[width=.5\linewidth]{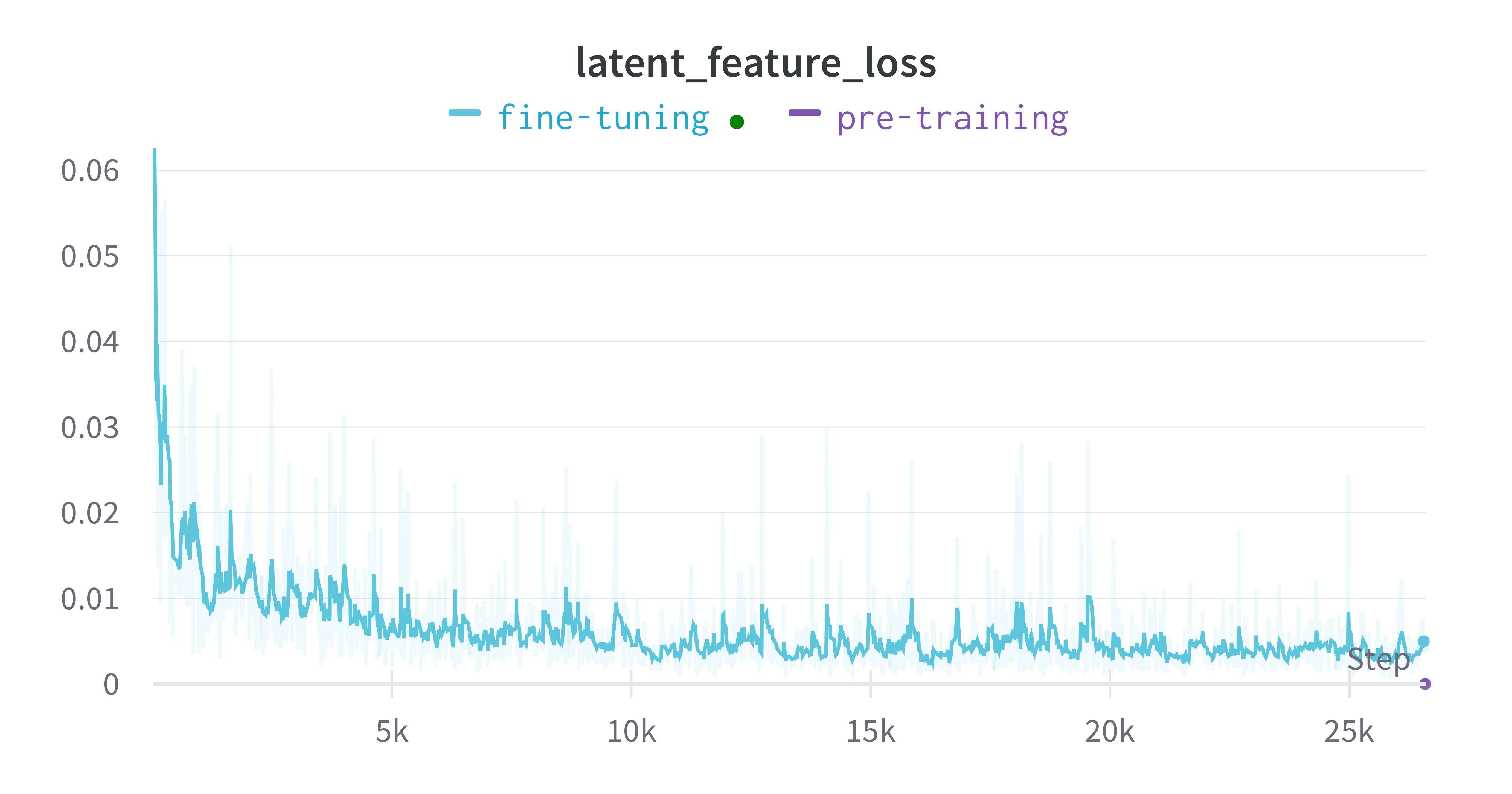}
    \vspace{-1\baselineskip}
    \caption{Loss curves of the latent feature encoder}
    \label{fig:lossl}
\end{figure}

\begin{figure}[htbp]
    \centering
    \includegraphics[width=.5\linewidth]{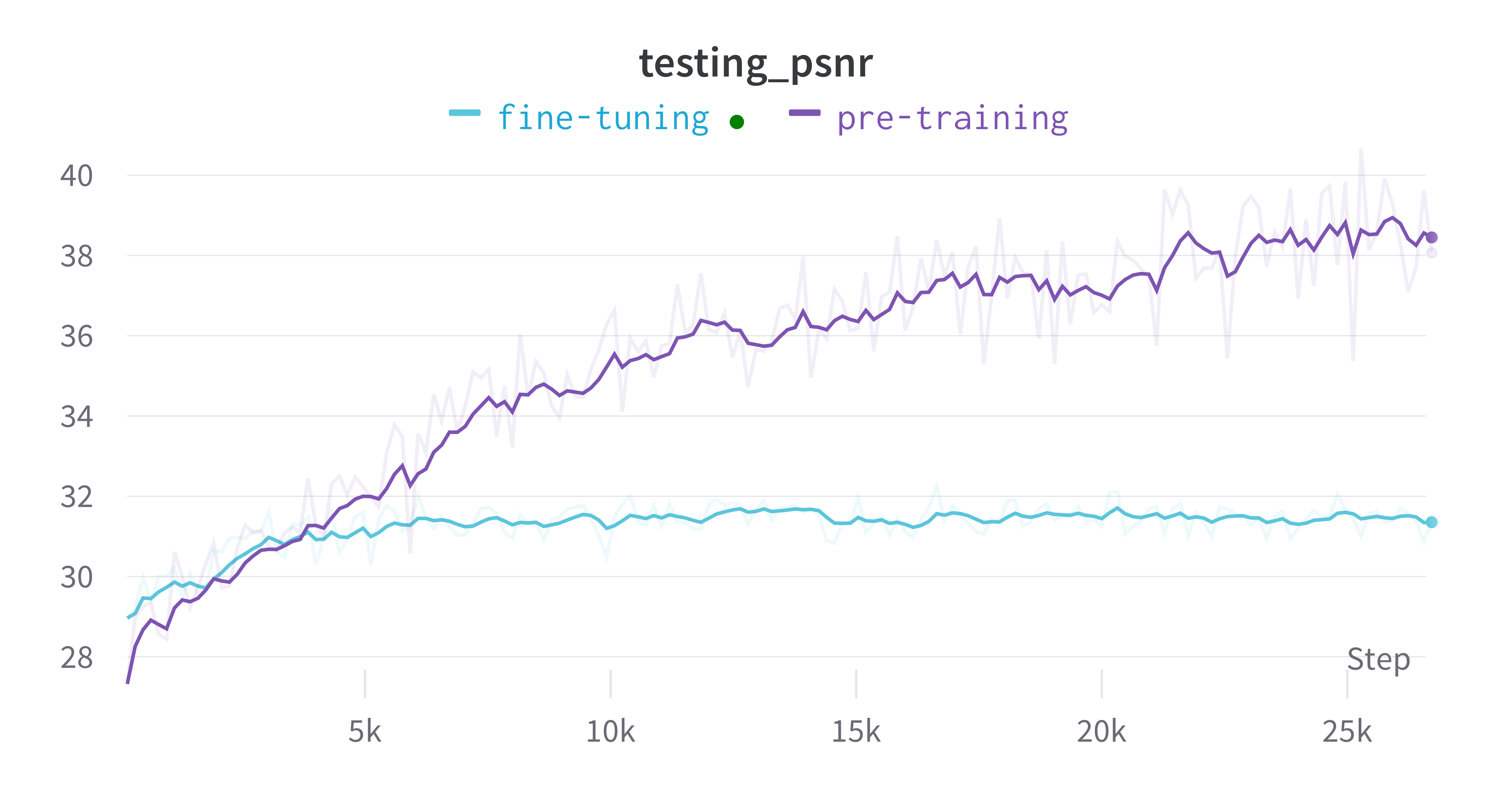}
    \vspace{-1\baselineskip}
    \caption{Loss curves of the PSNR value of a partial testing set (randomly selected 10 images).}
    \label{fig:psnr}
\end{figure}

The primary diffusion network architecture contains a multi-head attention U-Net~\cite{dhariwal2021diffusion}.
In the training process, we utilize a perception prioritized weighting scheme~\cite{choi2022perception} with $\gamma=1, k=1$ to accelerate the diffusion network learning.
Our diffusion reversion process utilizes an implicit diffusion model (\ie DDIM~\cite{song2020denoising}) for sampling acceleration, which is shown to be effective with 50 timesteps only.
The experiments are conducted using the PyTorch framework with 8 NVIDIA A100 GPUs (4 days training) and are reproducible with V100 GPUs with longer running times.
We use a constant learning rate $2.5e-5$ and find that the network converges after 200k iterations.
Fig.~\ref{fig:lossd}, Fig.~\ref{fig:lossl}, and Fig.~\ref{fig:psnr} show the curves related to the implementation details, respectively.

\newpage

\section{Detailed Diffusion Backward Process}
In the submission, we have omitted the details between $L_{VLB}$ defined in Eq.~\ref{eq:kl} and $L_{simple}$ defined in Eq.~\ref{eq:sloss} due to the space limitation.
Here, we provide a detailed derivation between these two losses. For each KL term in $L_{VLB}$, where:
\begin{align}
\begin{split}
L_T &= D_\text{KL}(q(\mathbf{y}_T \vert \mathbf{y}_0) \parallel p_\theta(\mathbf{y}_T)), \\
L_t &= D_\text{KL}(q(\mathbf{y}_{t-1} \vert \mathbf{y}_t, \mathbf{y}_0) \parallel p_\theta(\mathbf{y}_{t-1} \vert\mathbf{y}_t)), \\
L_0 &= - \log p_\theta(\mathbf{y}_0 \vert \mathbf{y}_1).
\end{split}
\end{align}
The first term of $L_T$ can be ignored because itself does not contain any parameters and $\by_T$ is just a Gaussian noise.
The third term of $L_0$ can be parameterized by a discrete decoder as $\mathcal{N}(\by_0|\boldsymbol{\mu}_\theta(\by_1, 1), \boldsymbol{\Sigma}_\theta(\by_1, 1))$~\cite{ho_denoising_2020}.
The second term of $L_t$ parameterized $q(\mathbf{y}_{t-1} \vert \mathbf{y}_t, \mathbf{y}_0)$ can be implemented by the mean $\tilde{\mu}_t (\by_t, \by_0)$ and variance $\tilde{\beta}_t$ of the standard Gaussian density function.
Specifically, we can represent the probability of $q(\mathbf{y}_{t-1} \vert \mathbf{y}_t, \mathbf{y}_0) $ by using Bayes' rule as:
\begin{align}
\tilde{\beta}_t 
&= 1/(\frac{\alpha_t}{\beta_t} + \frac{1}{1 - \bar{\alpha}_{t-1}})\\
\tilde{\mu}_t (\mathbf{y}_t, \mathbf{y}_0)
&= (\frac{\sqrt{\alpha_t}}{\beta_t} \mathbf{y}_t + \frac{\sqrt{\bar{\alpha}_{t-1} }}{1 - \bar{\alpha}_{t-1}} \mathbf{y}_0)/(\frac{\alpha_t}{\beta_t} + \frac{1}{1 - \bar{\alpha}_{t-1}}).
\end{align}
Here, we parameterize $\tilde{\mu}_t$ with $\boldsymbol{\mu}_\theta$ as:
\begin{align}
\boldsymbol{\mu}_\theta(\by_t, t) &= \frac{1}{\sqrt{\alpha_t}} \Big(\by_t - \frac{1 - \alpha_t}{\sqrt{1 - \bar{\alpha}_t}} \boldsymbol{\epsilon}_\theta(\by_t, t) \Big),
\end{align}
and thus the loss term of $L_t$ could be wrote as minimizing the difference between $\boldsymbol{\mu}_\theta$ and $\tilde{\mu}_t$ with the weight $\boldsymbol{\Sigma}_\theta$ as:
\begin{align}
\begin{split}
L_t &= \mathbb{E}_{\by_0, \boldsymbol{\epsilon}} 
\Big[
\frac{1}{2 \|\boldsymbol{\Sigma}_\theta \|^2_2} \|\frac{1}{\sqrt{\alpha_t}} \Big( \by_t - \frac{1 - \alpha_t}{\sqrt{1 - \bar{\alpha}_t}} \boldsymbol{\epsilon}_t \Big)
- \frac{1}{\sqrt{\alpha_t}} 
\Big(
\by_t - \frac{1 - \alpha_t}{\sqrt{1 - \bar{\alpha}_t}}
\boldsymbol{\epsilon}_\theta(\by_t, t)
\Big) \|^2 
\Big] \\
&= \mathbb{E}_{\mathbf{y}_0, \boldsymbol{\epsilon}} \Big[\frac{ (1 - \alpha_t)^2 }{2 \alpha_t (1 - \bar{\alpha}_t) \| \boldsymbol{\Sigma}_\theta \|^2_2} \|\boldsymbol{\epsilon}_t - \boldsymbol{\epsilon}_\theta(\by_t, t)\|^2 \Big],
\end{split}
\end{align}
which can be simplified together with $L_{0}$ and $L_{T}$ by ignoring the weights as:
\begin{align}
    L_{simple} = \mathbb{E}_{t \sim [1, T], \mathbf{y}_0, \boldsymbol{\epsilon}_t} \Big[\|\boldsymbol{\epsilon}_t - \boldsymbol{\epsilon}_\theta(\mathbf{y}_t, t)\|^2 \Big].
\end{align}
Moreover, in this paper, we set the variance $\tilde{\beta}_t$ as a sequence of linearly increasing constants suggested by Ho et al.~\cite{ho_denoising_2020}.

\newpage

\section{Synthetic Shadow Image Triplet Settings}
\begin{figure}[htbp]
    \centering
    \includegraphics[width=.3\linewidth]{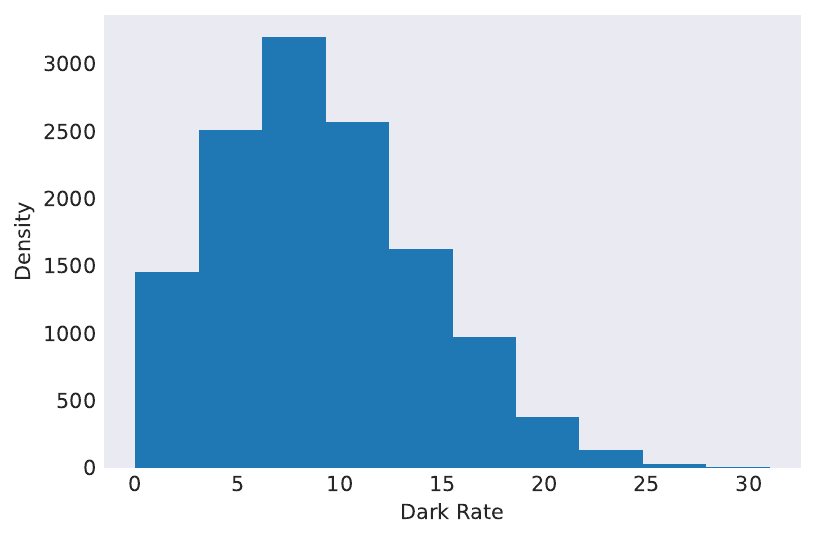}
    \includegraphics[width=.3\linewidth]{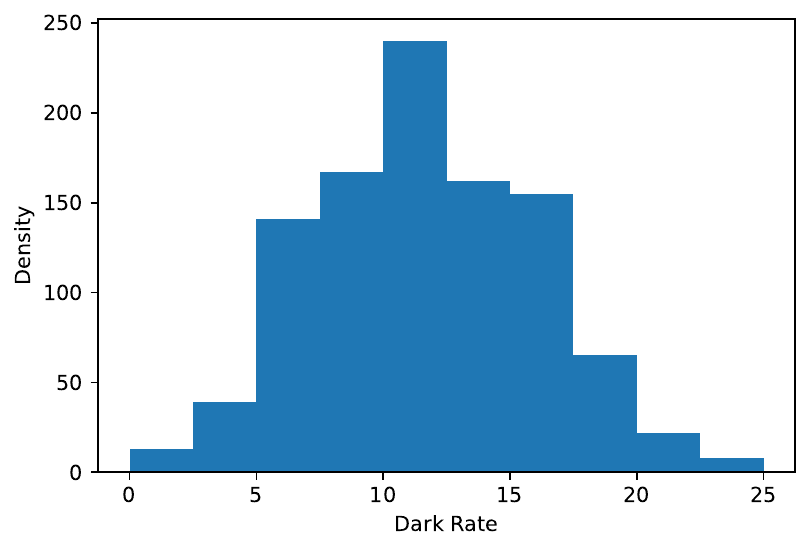}
    \vspace{-.5\baselineskip}
    \caption{\textbf{Dark rate distribution of the training set.} The left one is the synthetic dark shadow distribution according to recent work by~\cite{inoue2020learning}. The right one is the DeSOBA dataset dark shadow distribution.}
    \vspace{-.5\baselineskip}
    \label{fig:darkrate}
\end{figure}

Figure~\ref{fig:darkrate} visualize the dark rate of the synthetic training image triplets for instance-shadow removal.
We also visualize the synthetic image triplets under different dark rates to provide a straightforward understanding of the dataset constitution. 

\begin{figure}[htbp]
    \centering
    \includegraphics[width=.36\linewidth]{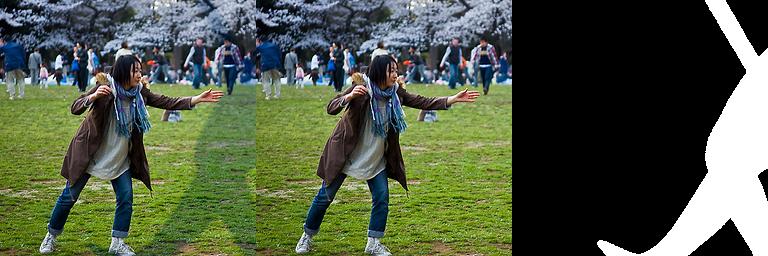}
    \includegraphics[width=.36\linewidth]{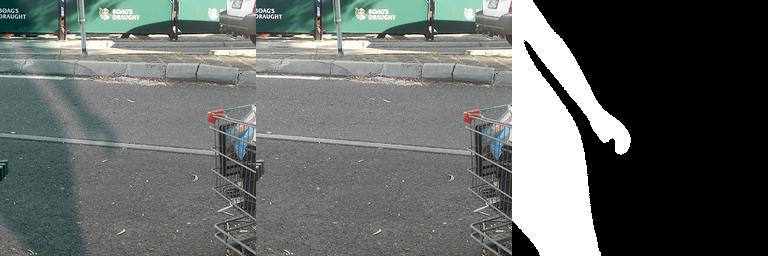}
    \hfill
    \includegraphics[width=.36\linewidth]{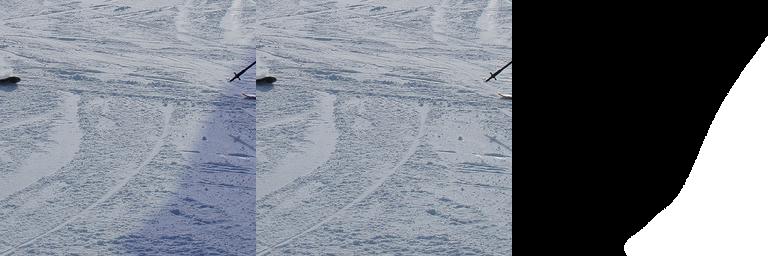}
    \includegraphics[width=.36\linewidth]{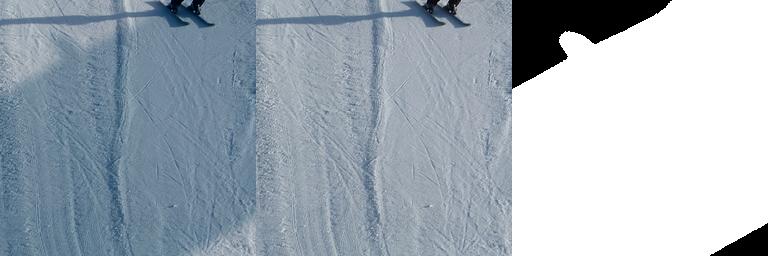}
    \vspace{-.5\baselineskip}
    \caption{\textbf{Synthetic Images with a dark rate of $5$.}}
    \vspace{-1\baselineskip}
\end{figure}
\begin{figure}[htbp]
    \centering
    \includegraphics[width=.36\linewidth]{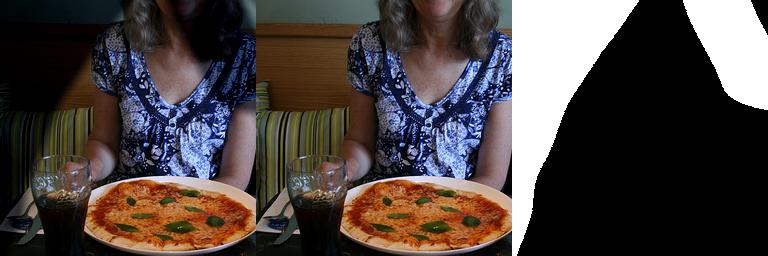}
    \includegraphics[width=.36\linewidth]{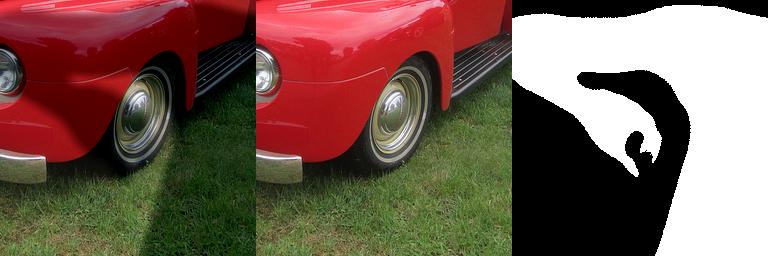}
    \hfill
    \includegraphics[width=.36\linewidth]{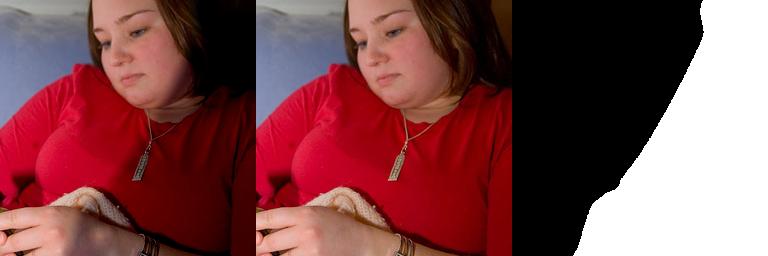}
    \includegraphics[width=.36\linewidth]{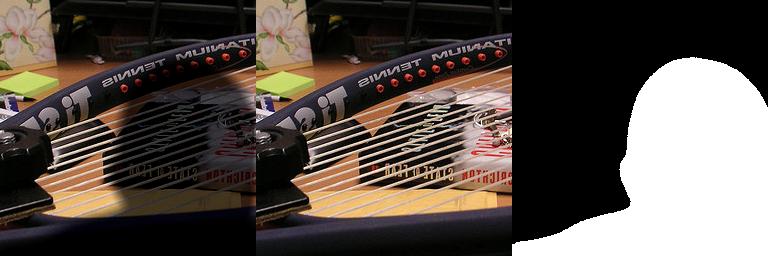}
    \vspace{-.5\baselineskip}    \caption{\textbf{Synthetic Images with a dark rate of $10$.}}
    \vspace{-1\baselineskip}
\end{figure}
\begin{figure}[htbp]
    \centering
    \includegraphics[width=.36\linewidth]{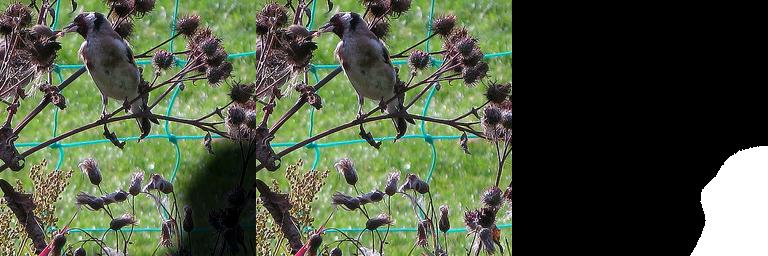}
    \includegraphics[width=.36\linewidth]{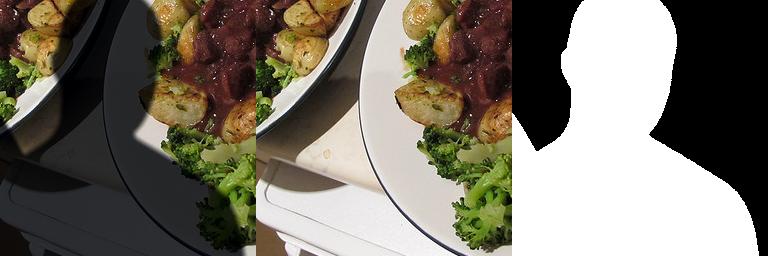}
    \hfill
    \includegraphics[width=.36\linewidth]{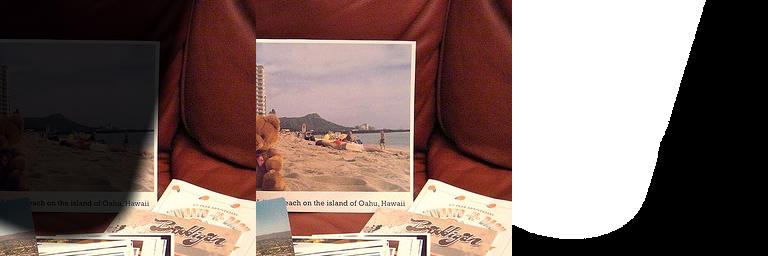}
    \includegraphics[width=.36\linewidth]{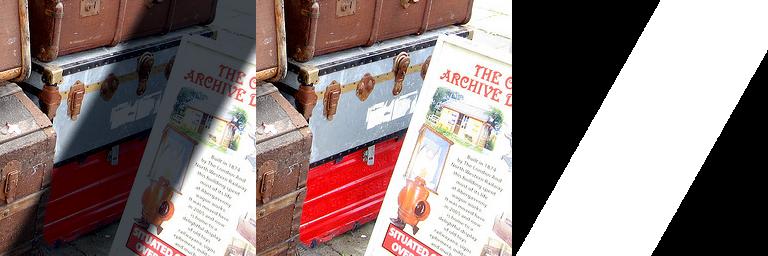}
    \vspace{-.5\baselineskip}
    \caption{\textbf{Synthetic Images with a dark rate of $20$.}}
\end{figure}
\newpage

\section{Additional Visual Comparisons of The \emph{AISTD} Dataset}
\begin{figure*}[htbp]
\centering
  \begin{subfigure}[t]{.12\linewidth}
    \captionsetup{justification=centering, labelformat=empty, font=scriptsize}

    \includegraphics[width=1\linewidth,height=.75\linewidth]{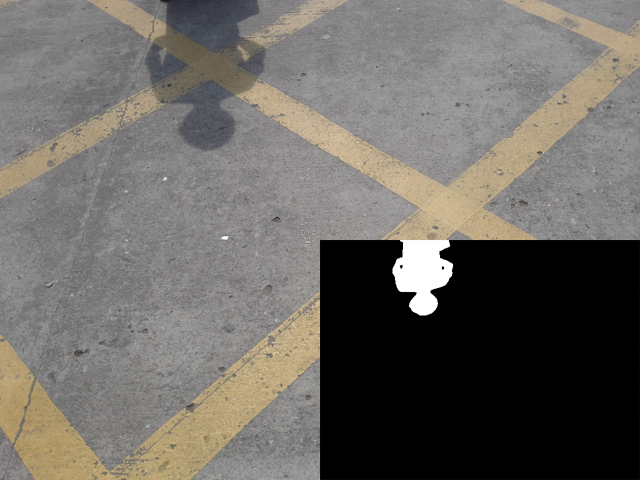}
    \includegraphics[width=1\linewidth,height=.75\linewidth]{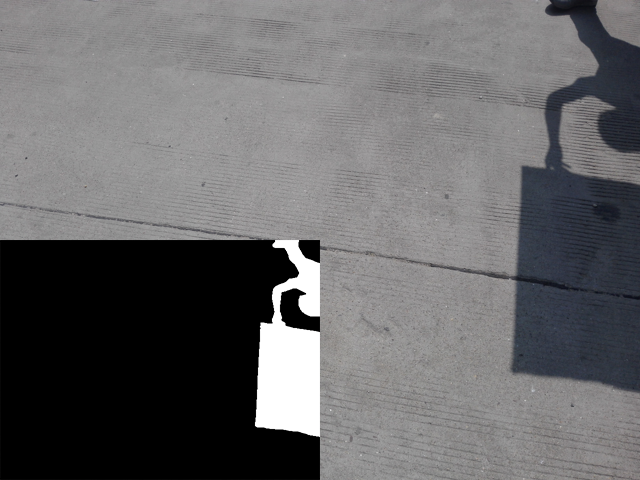}
    \includegraphics[width=1\linewidth,height=.75\linewidth]{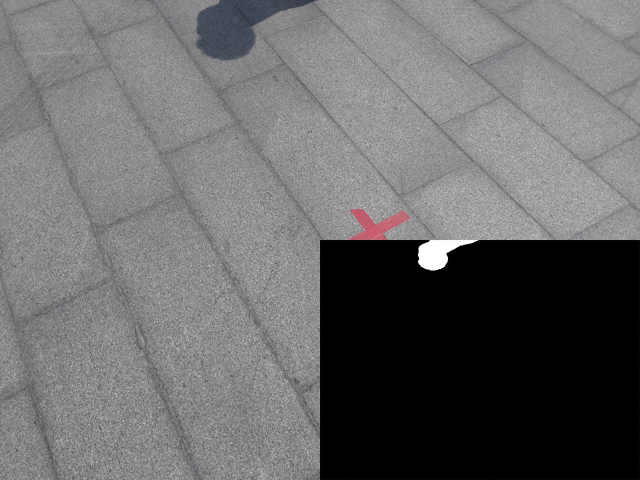}
    \includegraphics[width=1\linewidth,height=.75\linewidth]{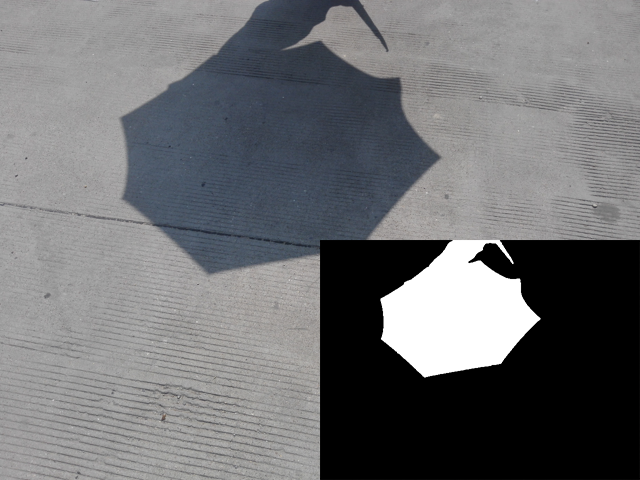}
    \includegraphics[width=1\linewidth,height=.75\linewidth]{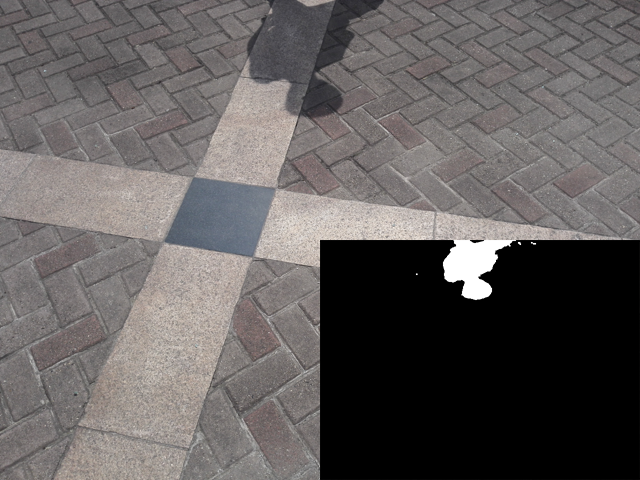}
    \includegraphics[width=1\linewidth,height=.75\linewidth]{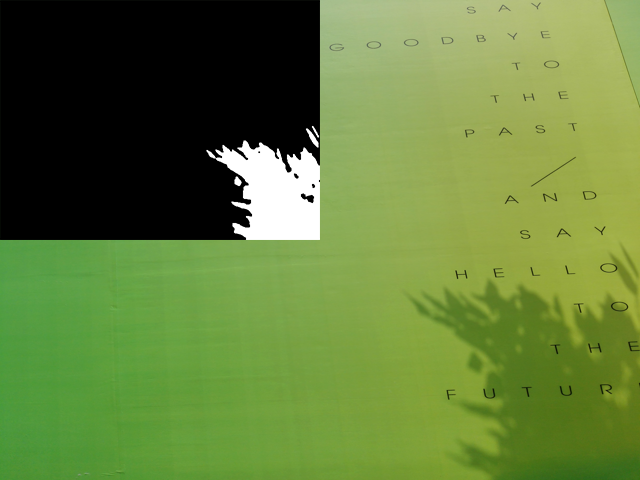}
    \includegraphics[width=1\linewidth,height=.75\linewidth]{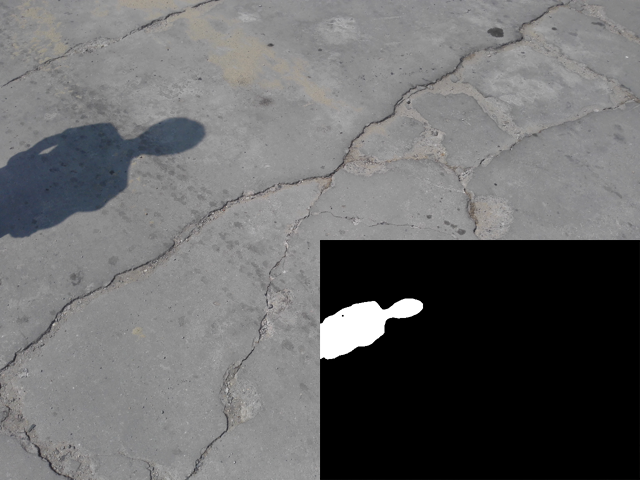}
    \includegraphics[width=1\linewidth,height=.75\linewidth]{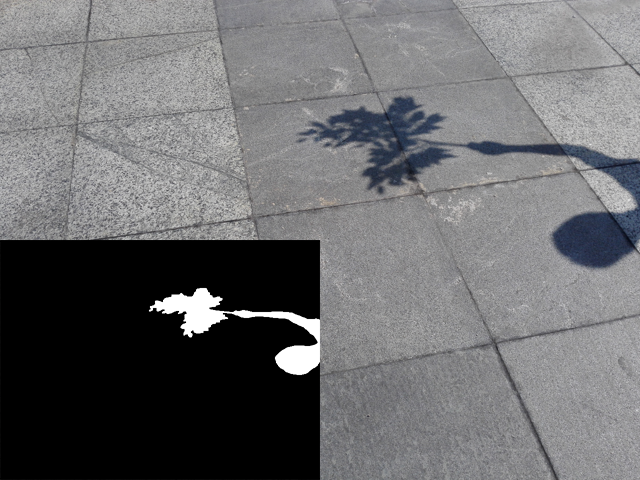}
    \caption{Shadow \& Mask}
  \end{subfigure}
  \begin{subfigure}[t]{.12\linewidth}
    \captionsetup{justification=centering, labelformat=empty, font=scriptsize}

    \includegraphics[width=1\linewidth,height=.75\linewidth]{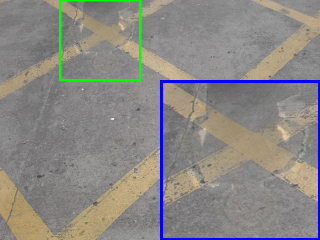}

    \includegraphics[width=1\linewidth,height=.75\linewidth]{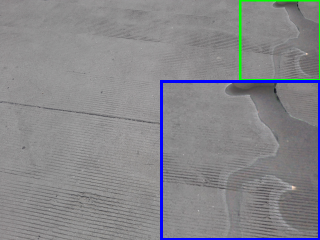}
    \includegraphics[width=1\linewidth,height=.75\linewidth]{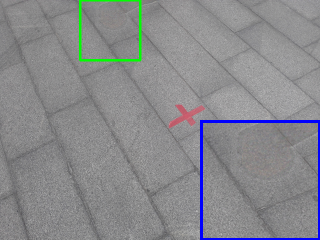}
    \includegraphics[width=1\linewidth,height=.75\linewidth]{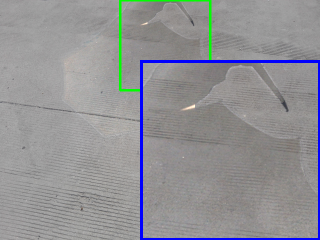}
    \includegraphics[width=1\linewidth,height=.75\linewidth]{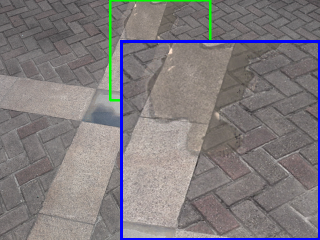}
    \includegraphics[width=1\linewidth,height=.75\linewidth]{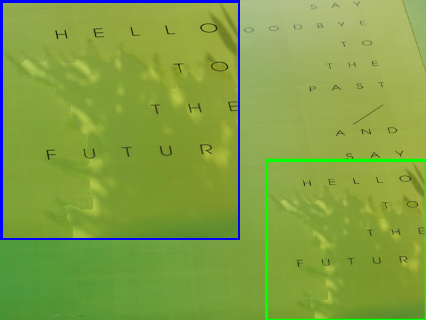}
    \includegraphics[width=1\linewidth,height=.75\linewidth]{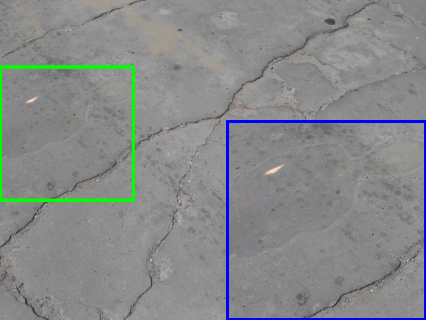}
    \includegraphics[width=1\linewidth,height=.75\linewidth]{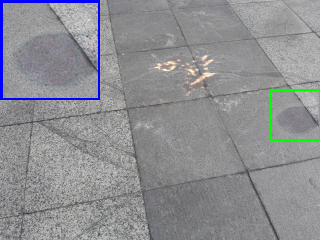}
    \caption{\tiny Param+M+D-Net~\cite{le2020shadow}}
  \end{subfigure}
  \begin{subfigure}[t]{.12\linewidth}
    \captionsetup{justification=centering, labelformat=empty, font=scriptsize}

    \includegraphics[width=1\linewidth,height=.75\linewidth]{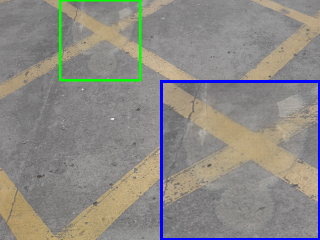}

    \includegraphics[width=1\linewidth,height=.75\linewidth]{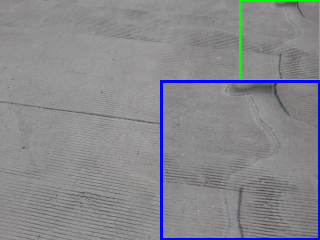}
    \includegraphics[width=1\linewidth,height=.75\linewidth]{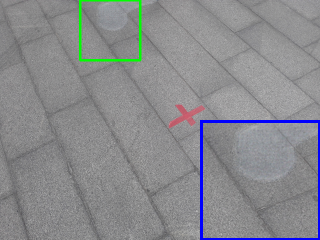}
    \includegraphics[width=1\linewidth,height=.75\linewidth]{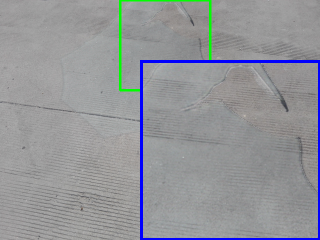}
    \includegraphics[width=1\linewidth,height=.75\linewidth]{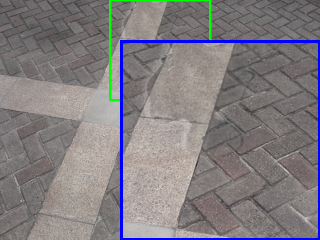}
    \includegraphics[width=1\linewidth,height=.75\linewidth]{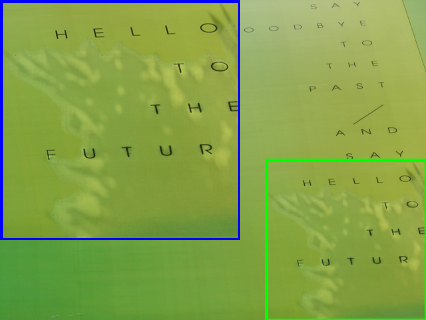}
    \includegraphics[width=1\linewidth,height=.75\linewidth]{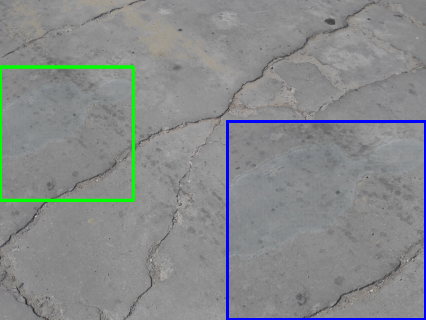}
    \includegraphics[width=1\linewidth,height=.75\linewidth]{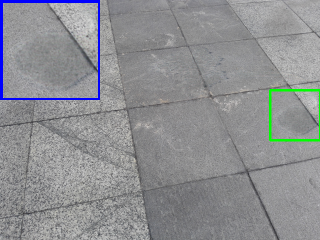}
    \caption{\tiny G2R-ShadowNet~\cite{liu_shadow_2021}}
  \end{subfigure}
  \begin{subfigure}[t]{.12\linewidth}
    \captionsetup{justification=centering, labelformat=empty, font=scriptsize}

    \includegraphics[width=1\linewidth,height=.75\linewidth]{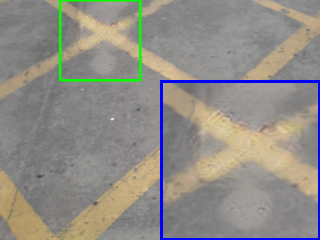}
    \includegraphics[width=1\linewidth,height=.75\linewidth]{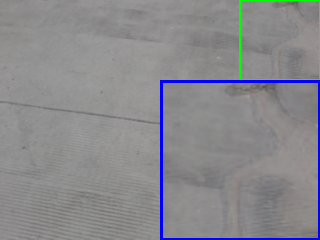}
    \includegraphics[width=1\linewidth,height=.75\linewidth]{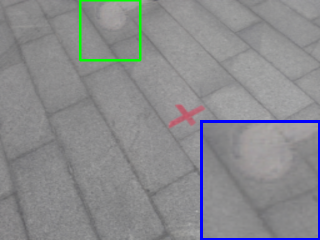}
    \includegraphics[width=1\linewidth,height=.75\linewidth]{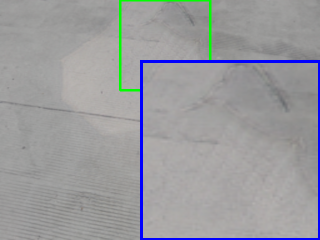}
    \includegraphics[width=1\linewidth,height=.75\linewidth]{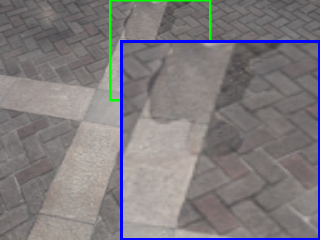}
    \includegraphics[width=1\linewidth,height=.75\linewidth]{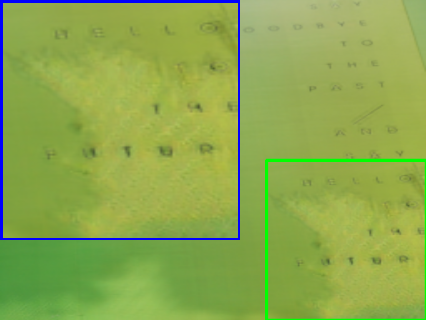}
    \includegraphics[width=1\linewidth,height=.75\linewidth]{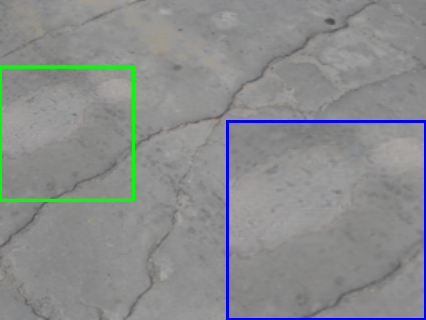}
    \includegraphics[width=1\linewidth,height=.75\linewidth]{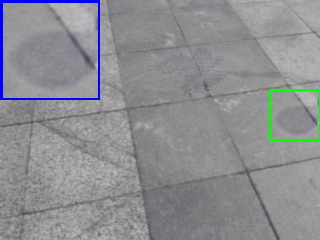}
    \caption{DC-ShadowNet~\cite{jin2021dc}}
  \end{subfigure}
  \begin{subfigure}[t]{.12\linewidth}
    \captionsetup{justification=centering, labelformat=empty, font=scriptsize}

    \includegraphics[width=1\linewidth,height=.75\linewidth]{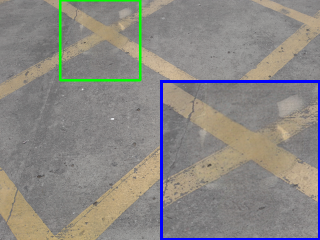}
    \includegraphics[width=1\linewidth,height=.75\linewidth]{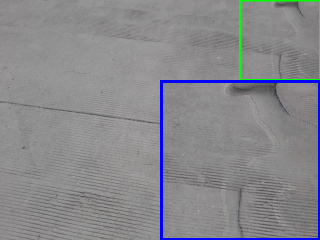}
    \includegraphics[width=1\linewidth,height=.75\linewidth]{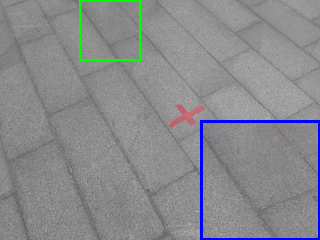}
    \includegraphics[width=1\linewidth,height=.75\linewidth]{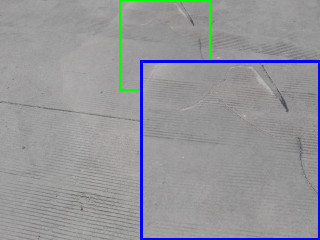}
    \includegraphics[width=1\linewidth,height=.75\linewidth]{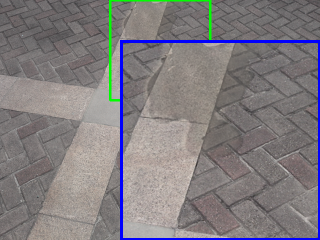}
    \includegraphics[width=1\linewidth,height=.75\linewidth]{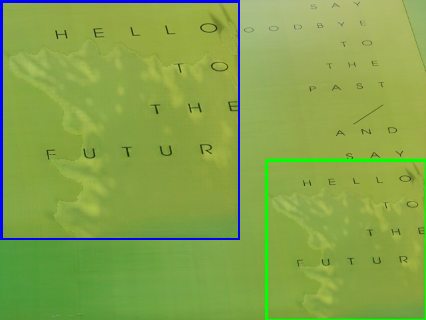}
    \includegraphics[width=1\linewidth,height=.75\linewidth]{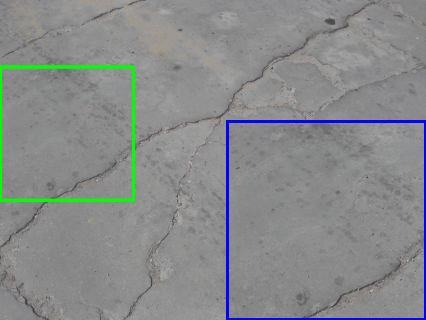}
    \includegraphics[width=1\linewidth,height=.75\linewidth]{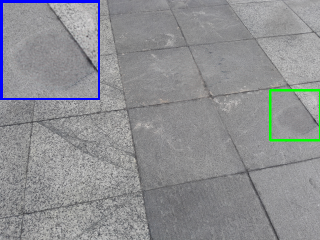}
    \caption{SG-ShadowNet~\cite{wan2022}}
  \end{subfigure}
  \begin{subfigure}[t]{.12\linewidth}
    \captionsetup{justification=centering, labelformat=empty, font=scriptsize}

    \includegraphics[width=1\linewidth,height=.75\linewidth]{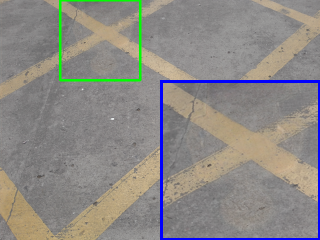}
    \includegraphics[width=1\linewidth,height=.75\linewidth]{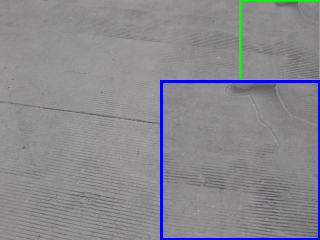}
    \includegraphics[width=1\linewidth,height=.75\linewidth]{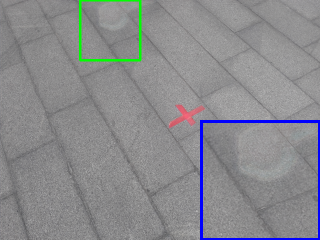}
    \includegraphics[width=1\linewidth,height=.75\linewidth]{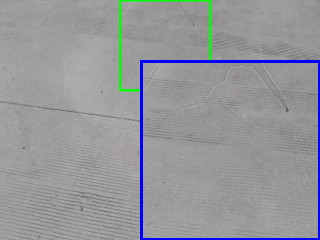}
    \includegraphics[width=1\linewidth,height=.75\linewidth]{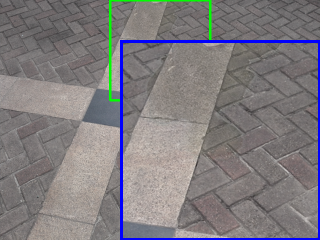}
    \includegraphics[width=1\linewidth,height=.75\linewidth]{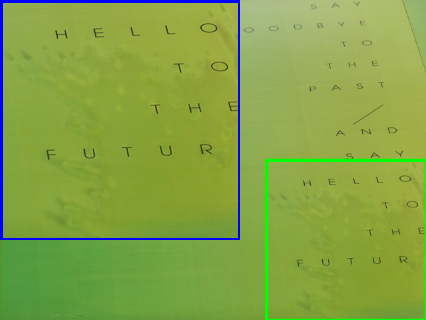}
    \includegraphics[width=1\linewidth,height=.75\linewidth]{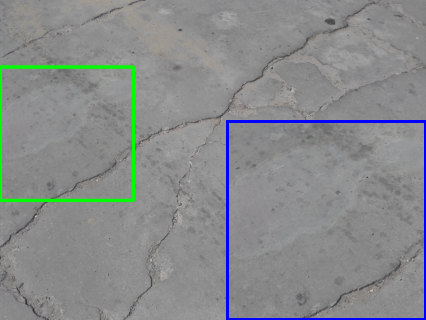}
    \includegraphics[width=1\linewidth,height=.75\linewidth]{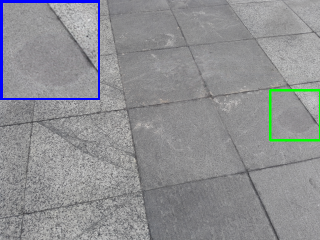}
    \caption{BMN~\cite{zhu_bijective_2022}}
  \end{subfigure}
  \begin{subfigure}[t]{.12\linewidth}
    \captionsetup{justification=centering, labelformat=empty, font=scriptsize}
    \includegraphics[width=1\linewidth,height=.75\linewidth]{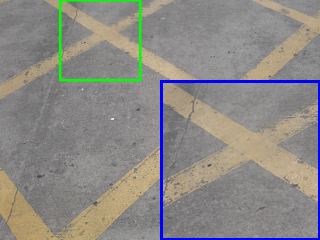}
    \includegraphics[width=1\linewidth,height=.75\linewidth]{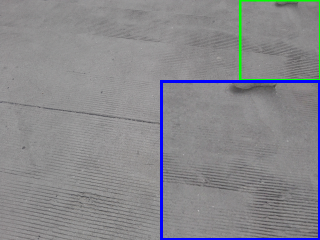}
    \includegraphics[width=1\linewidth,height=.75\linewidth]{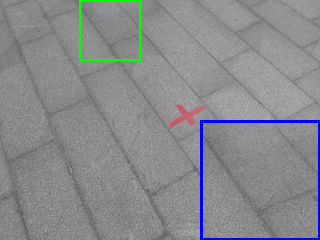}
    \includegraphics[width=1\linewidth,height=.75\linewidth]{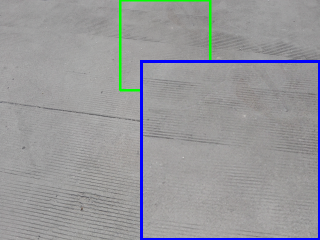}
    \includegraphics[width=1\linewidth,height=.75\linewidth]{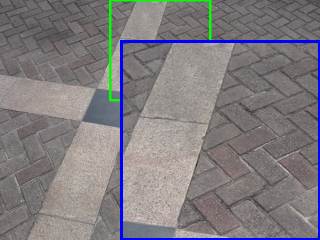}
    \includegraphics[width=1\linewidth,height=.75\linewidth]{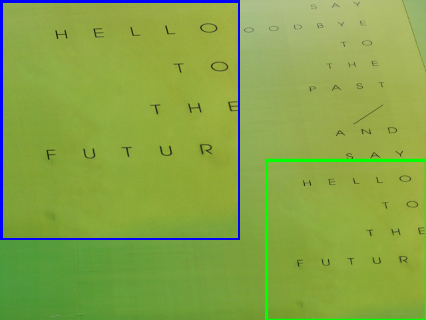}
    \includegraphics[width=1\linewidth,height=.75\linewidth]{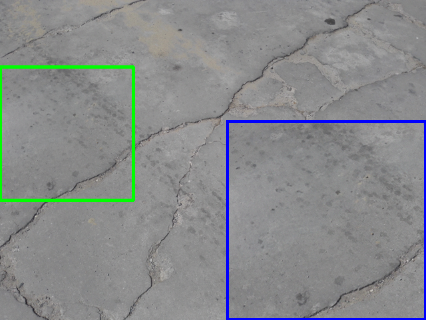}
    \includegraphics[width=1\linewidth,height=.75\linewidth]{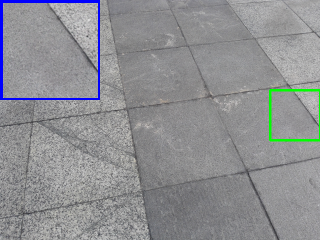}
    \caption{\tiny \ours~(Ours)}
  \end{subfigure}
  \begin{subfigure}[t]{.12\linewidth}
    \captionsetup{justification=centering, labelformat=empty, font=scriptsize}
    \includegraphics[width=1\linewidth,height=.75\linewidth]{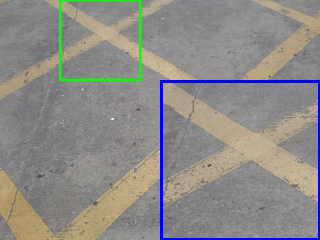}
    \includegraphics[width=1\linewidth,height=.75\linewidth]{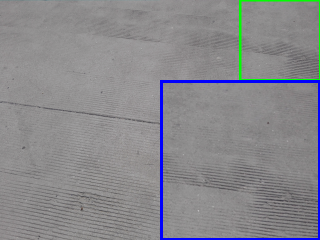}
    \includegraphics[width=1\linewidth,height=.75\linewidth]{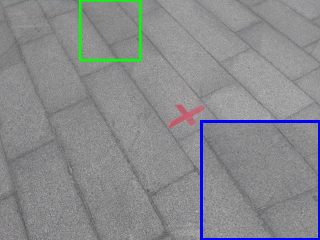}
    \includegraphics[width=1\linewidth,height=.75\linewidth]{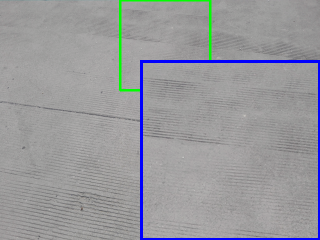}
    \includegraphics[width=1\linewidth,height=.75\linewidth]{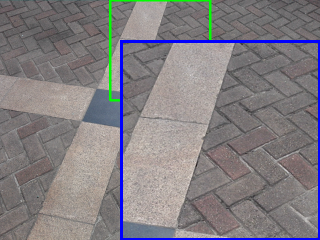}
    \includegraphics[width=1\linewidth,height=.75\linewidth]{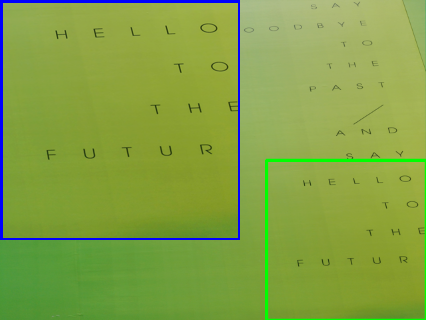}
    \includegraphics[width=1\linewidth,height=.75\linewidth]{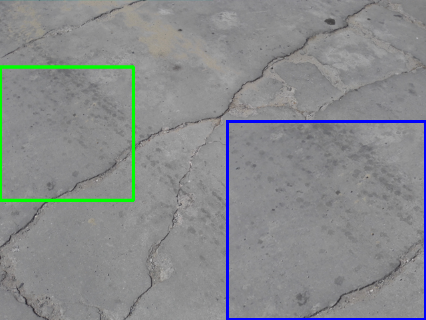}
    \includegraphics[width=1\linewidth,height=.75\linewidth]{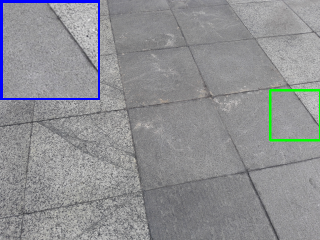}
    \caption{Ground Truth}
  \end{subfigure}
  \caption{\textbf{Visual comparisons of representative hard shadow removal results on the \emph{AISTD} dataset.} Here we highlight key details under the shadows (the green box points to the shadow region of the image and the blue box is a zoomed-in crop of the green box). Our model preserves details and removes other subtle shadow effects.}
  \vspace{-1\baselineskip}
  \label{fig:aistd}
\end{figure*}
\newpage

\section{Additional Visual Comparisons of The \emph{SRD} Dataset}
\begin{figure*}[ht]
\centering
  \begin{subfigure}[t]{.135\linewidth}
    \captionsetup{justification=centering, labelformat=empty, font=scriptsize}
    \includegraphics[width=1\linewidth,height=1\linewidth]{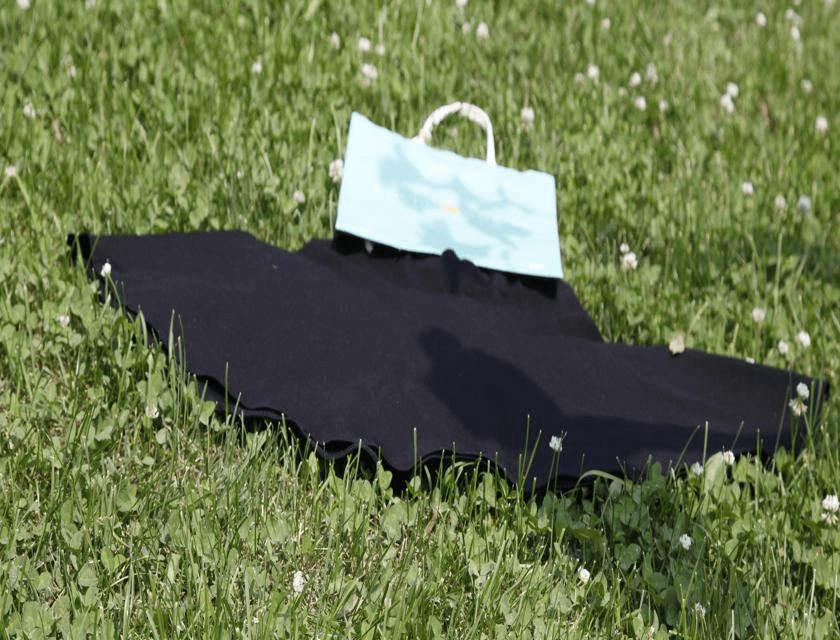}
    \includegraphics[width=1\linewidth,height=1\linewidth]{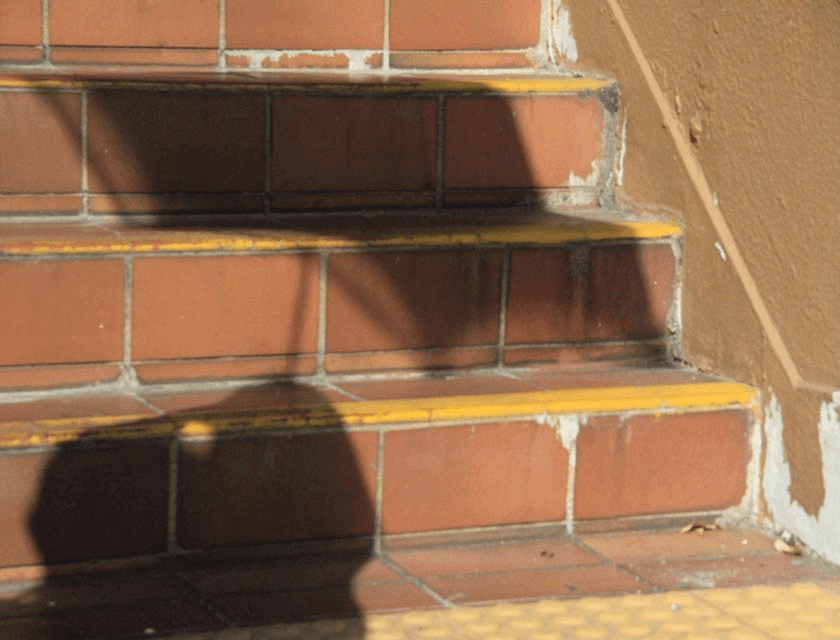}
    \includegraphics[width=1\linewidth,height=1\linewidth]{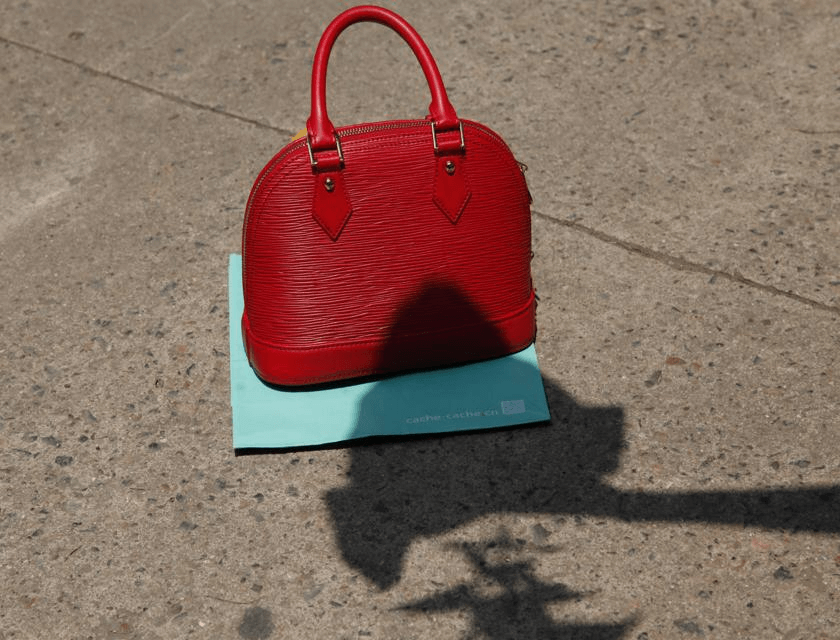}
    \includegraphics[width=1\linewidth,height=1\linewidth]{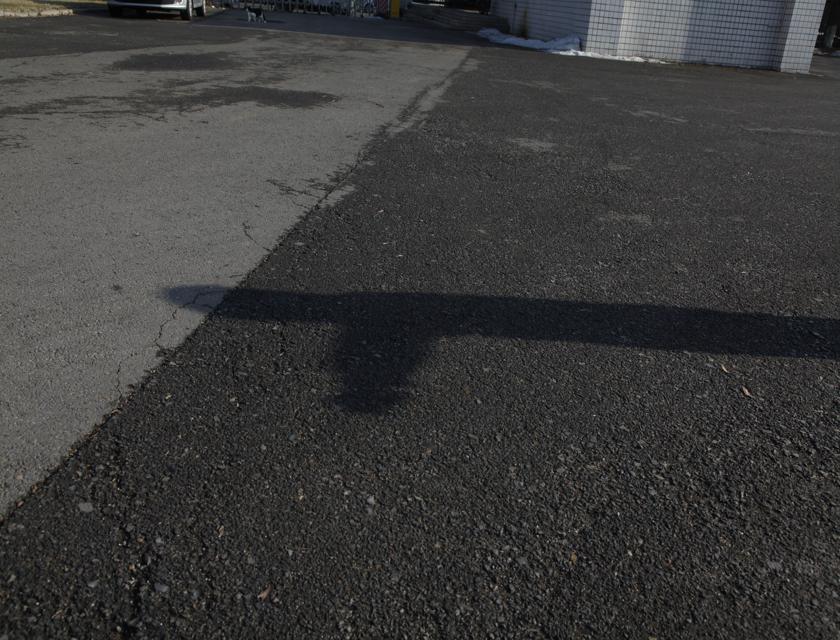}
    \includegraphics[width=1\linewidth,height=1\linewidth]{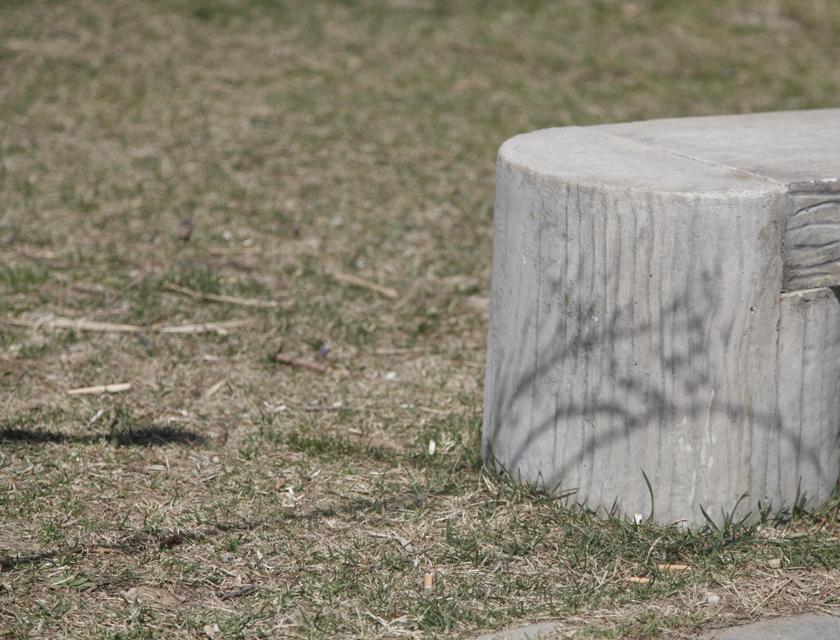}
    \includegraphics[width=1\linewidth,height=1\linewidth]{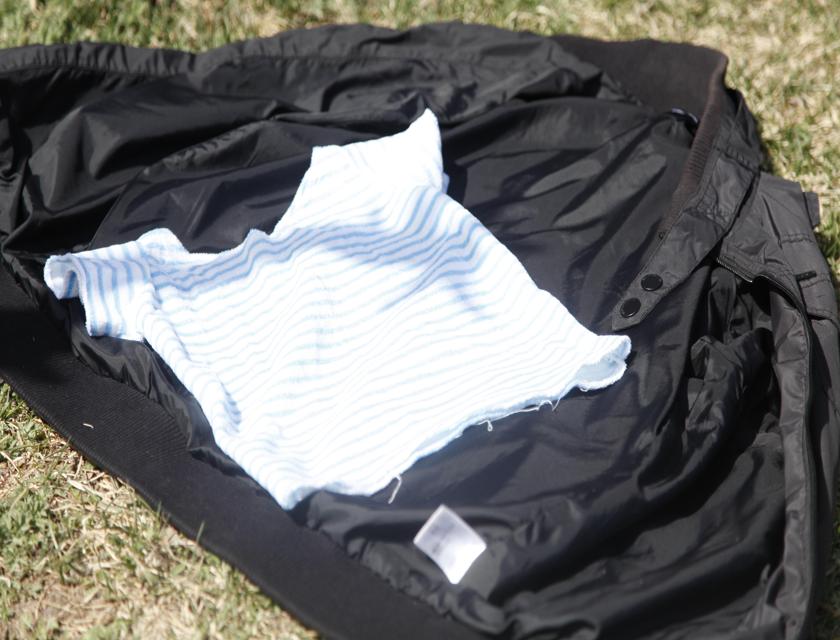}
    \includegraphics[width=1\linewidth,height=1\linewidth]{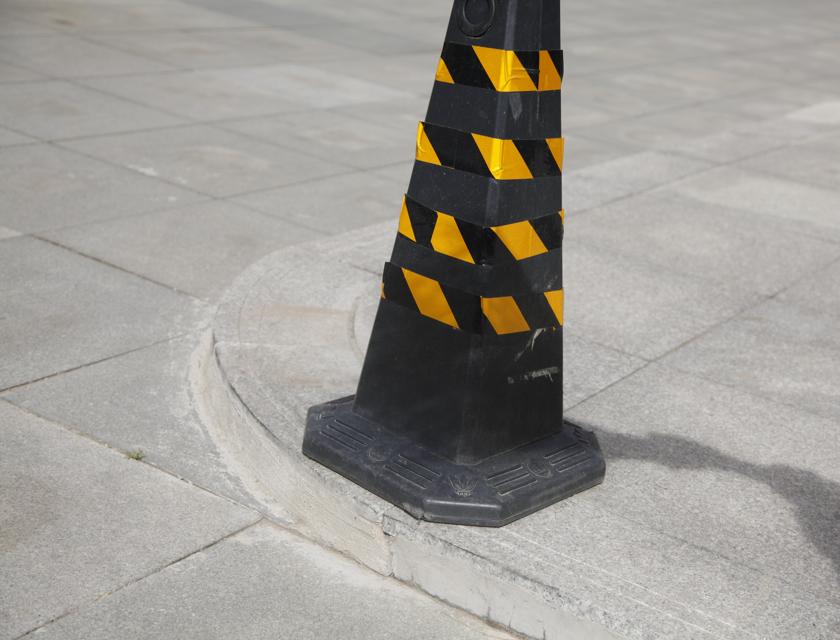}
    \caption{Shadows}
  \end{subfigure}
  \begin{subfigure}[t]{.135\linewidth}
    \captionsetup{justification=centering, labelformat=empty, font=scriptsize}
    \includegraphics[width=1\linewidth,height=1\linewidth]{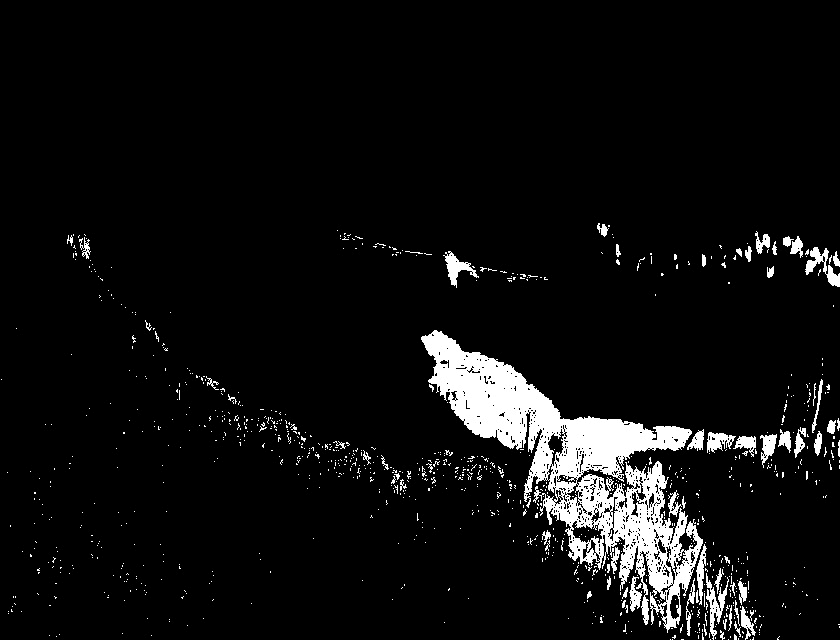}
    \includegraphics[width=1\linewidth,height=1\linewidth]{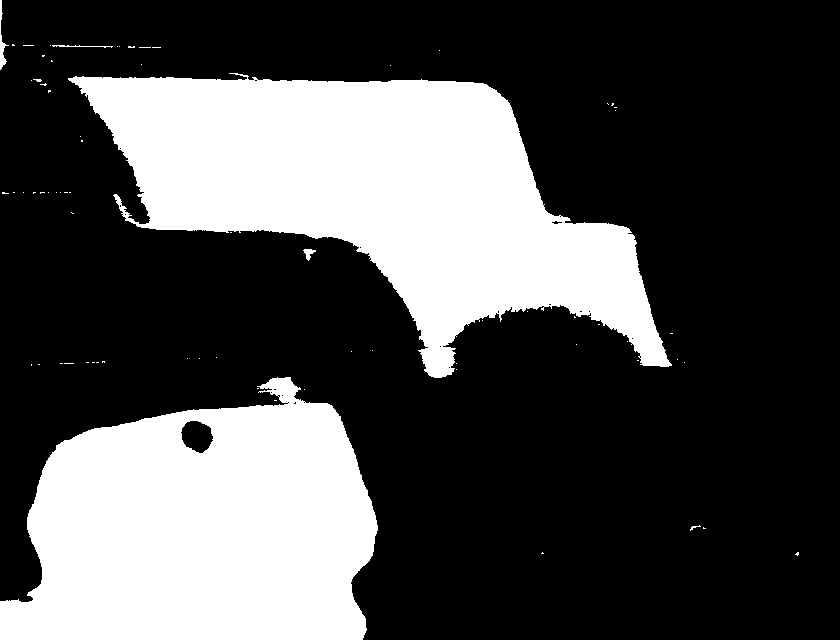}
    \includegraphics[width=1\linewidth,height=1\linewidth]{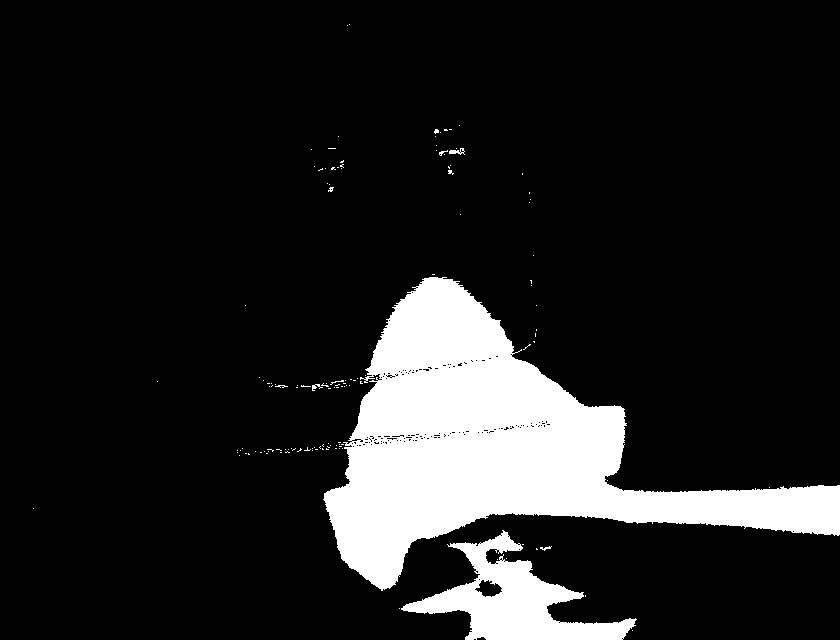}
    \includegraphics[width=1\linewidth,height=1\linewidth]{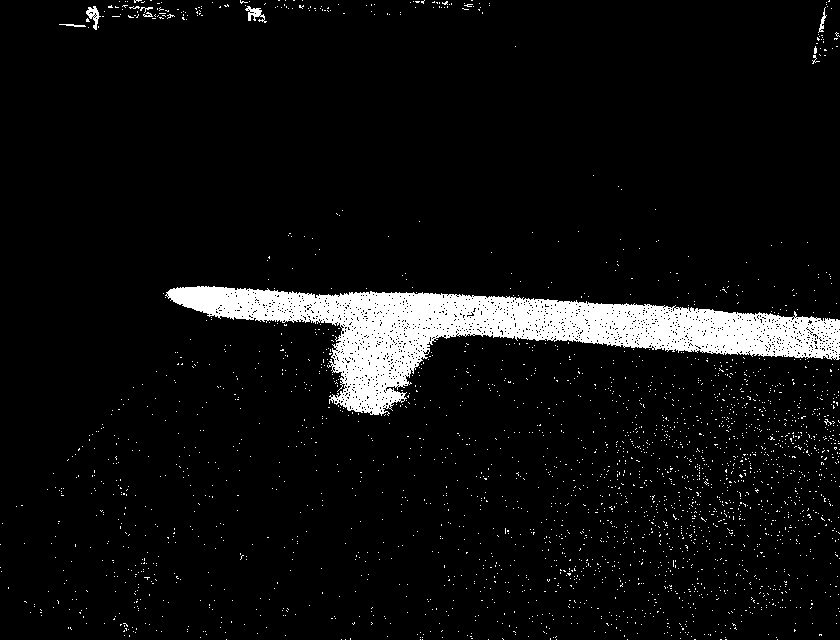}
    \includegraphics[width=1\linewidth,height=1\linewidth]{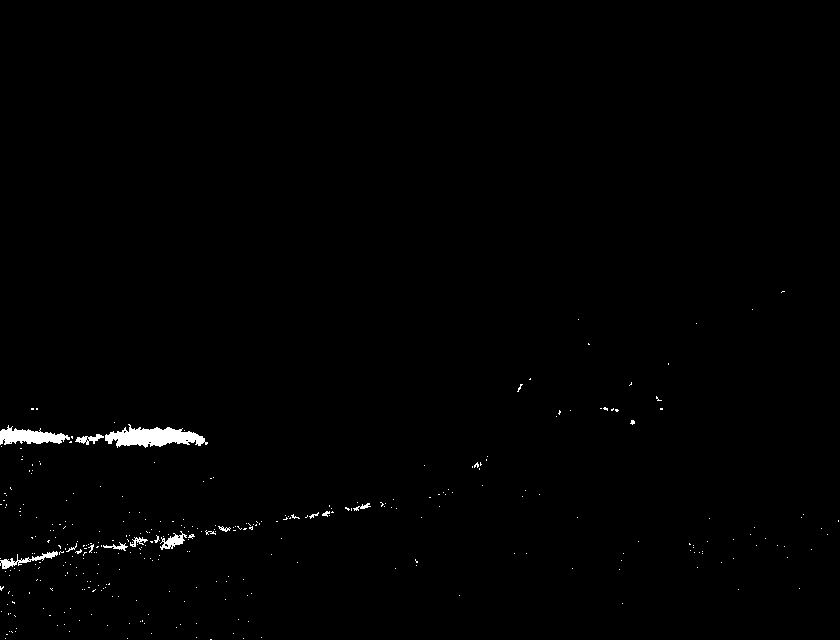}
    \includegraphics[width=1\linewidth,height=1\linewidth]{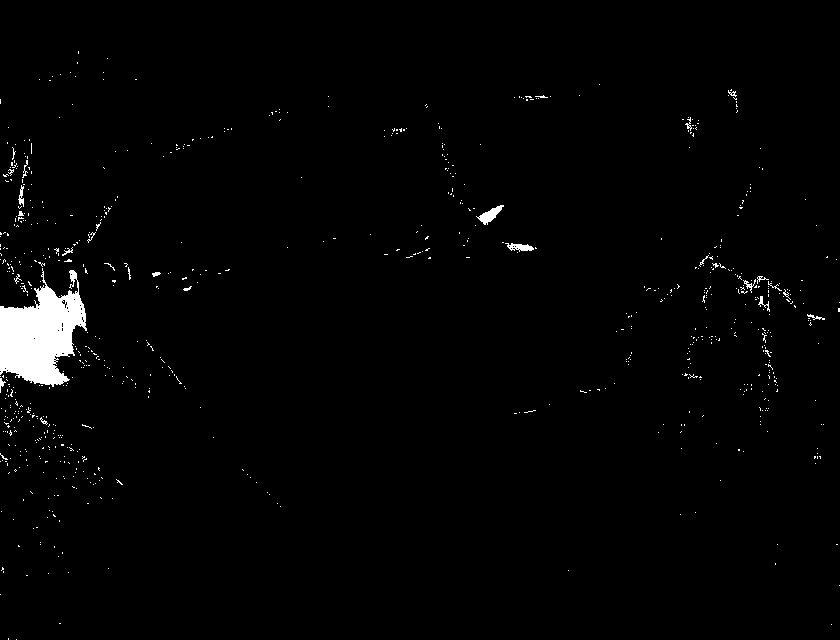}
    \includegraphics[width=1\linewidth,height=1\linewidth]{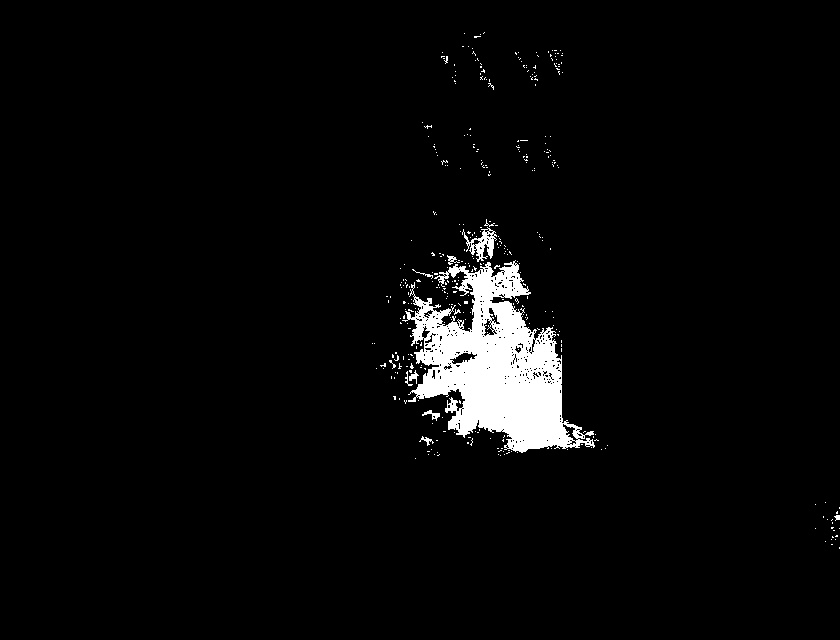}
    \caption{Shadow Mask}
  \end{subfigure}
  \begin{subfigure}[t]{.135\linewidth}
    \captionsetup{justification=centering, labelformat=empty, font=scriptsize}
    \includegraphics[width=1\linewidth,height=1\linewidth]{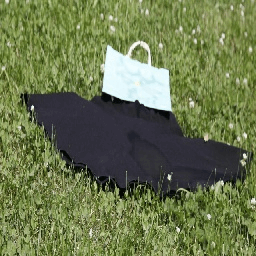}
    \includegraphics[width=1\linewidth,height=1\linewidth]{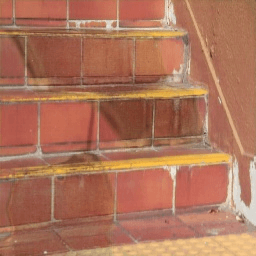}
    \includegraphics[width=1\linewidth,height=1\linewidth]{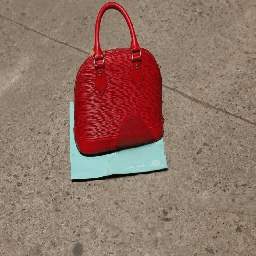}
    \includegraphics[width=1\linewidth,height=1\linewidth]{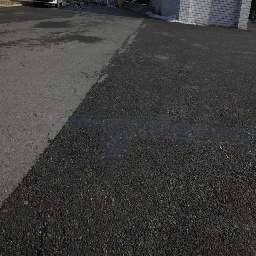}
    \includegraphics[width=1\linewidth,height=1\linewidth]{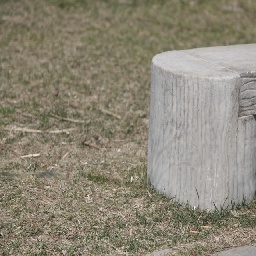}
    \includegraphics[width=1\linewidth,height=1\linewidth]{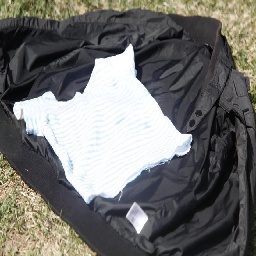}
    \includegraphics[width=1\linewidth,height=1\linewidth]{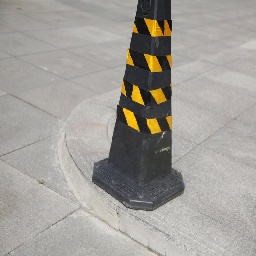}
    \caption{Auto-Exposure~\cite{fu2021auto}}
  \end{subfigure}
  \begin{subfigure}[t]{.135\linewidth}
    \captionsetup{justification=centering, labelformat=empty, font=scriptsize}
    \includegraphics[width=1\linewidth,height=1\linewidth]{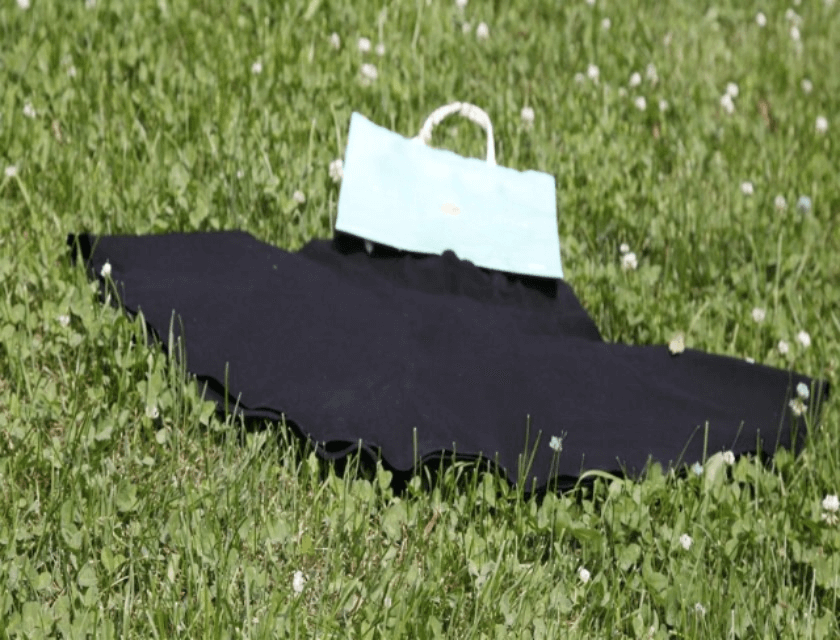}
    \includegraphics[width=1\linewidth,height=1\linewidth]{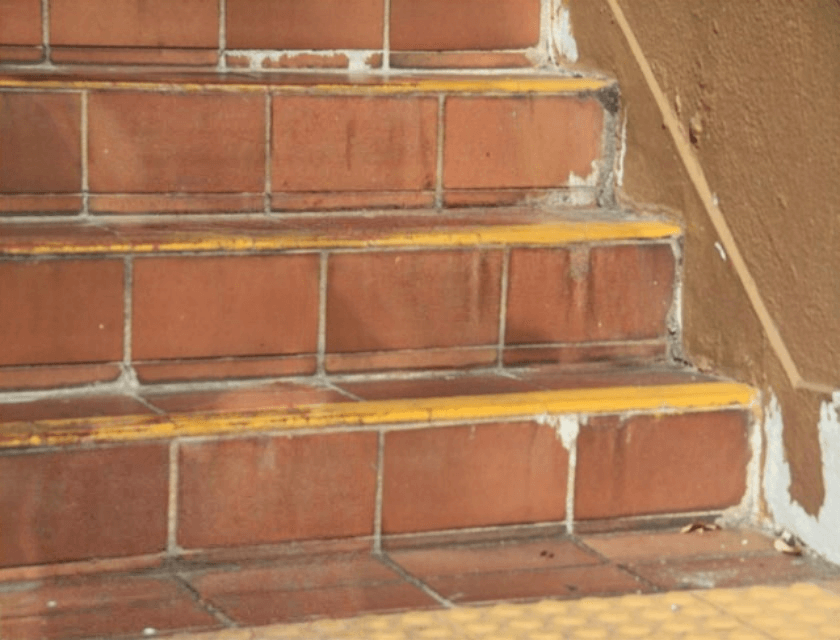}
    \includegraphics[width=1\linewidth,height=1\linewidth]{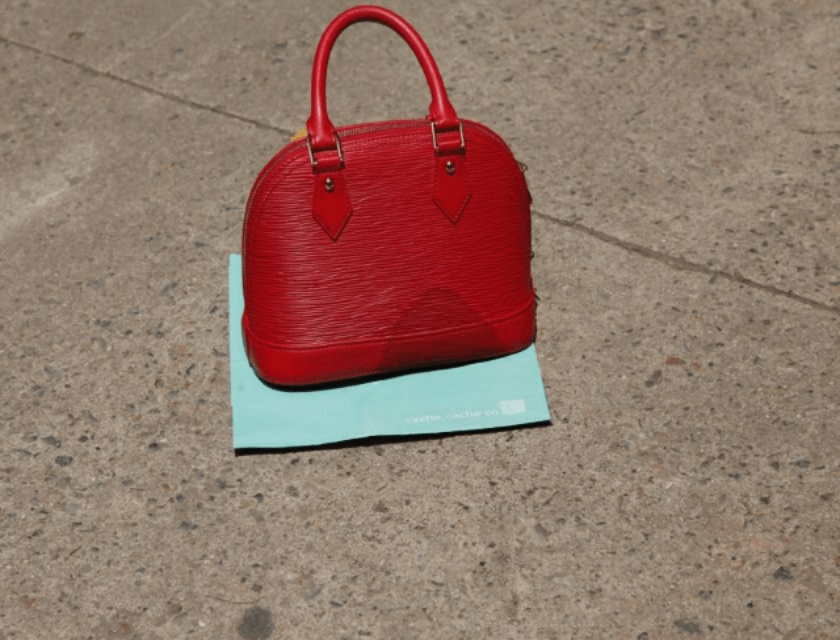}
    \includegraphics[width=1\linewidth,height=1\linewidth]{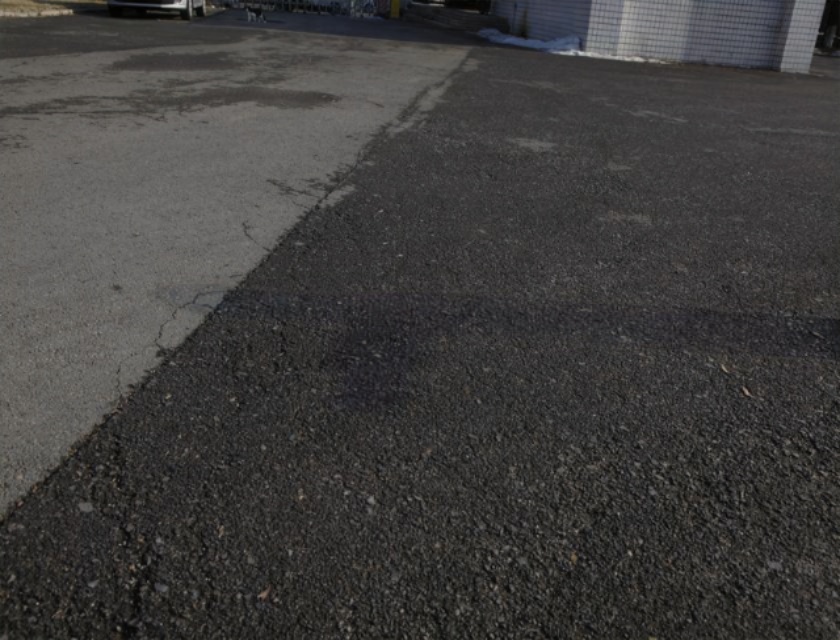}
    \includegraphics[width=1\linewidth,height=1\linewidth]{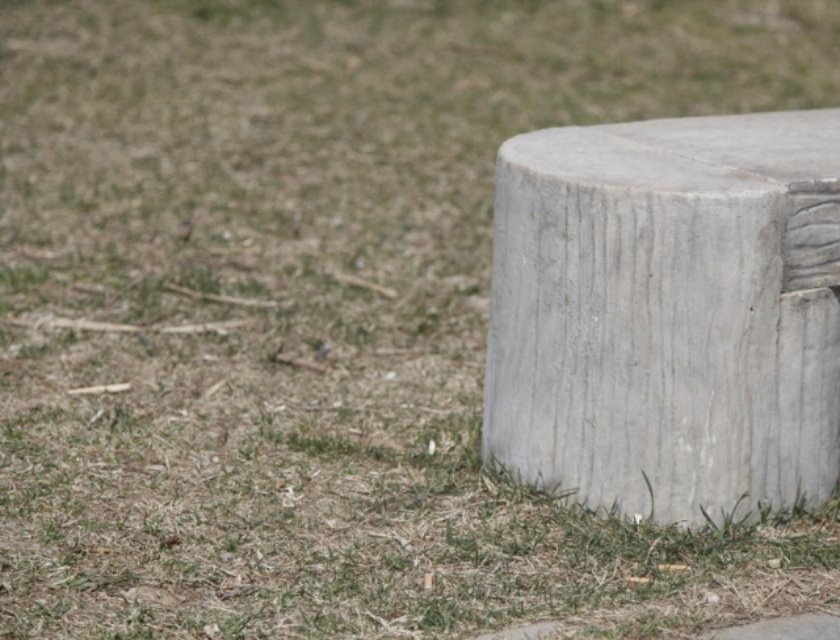}
    \includegraphics[width=1\linewidth,height=1\linewidth]{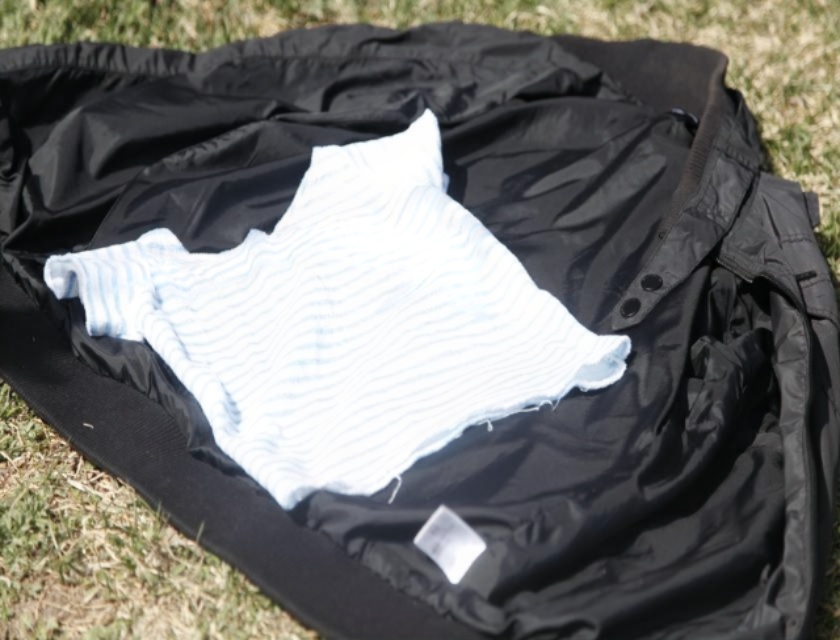}
    \includegraphics[width=1\linewidth,height=1\linewidth]{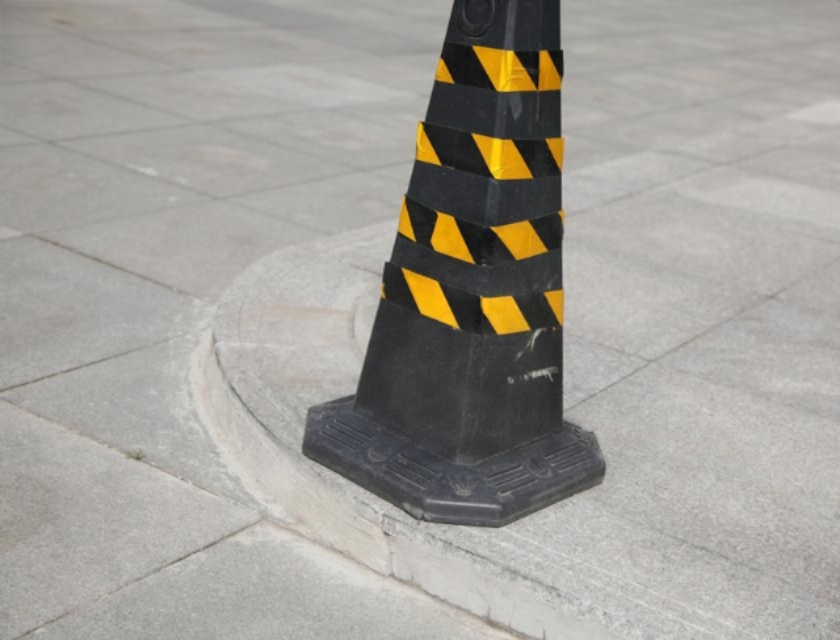}
    \caption{DHAN~\cite{cun_towards_2020}}
  \end{subfigure}
  \begin{subfigure}[t]{.135\linewidth}
    \captionsetup{justification=centering, labelformat=empty, font=scriptsize}
    \includegraphics[width=1\linewidth,height=1\linewidth]{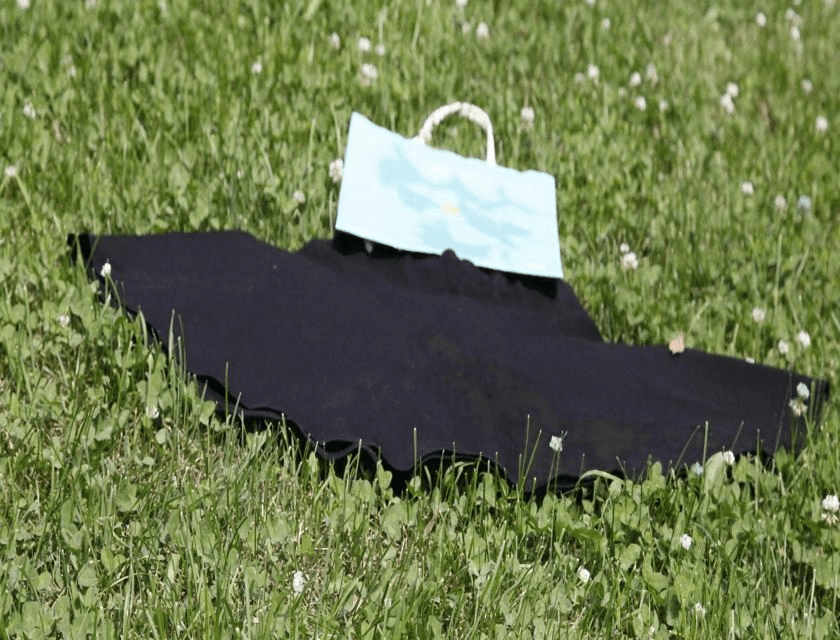}
    \includegraphics[width=1\linewidth,height=1\linewidth]{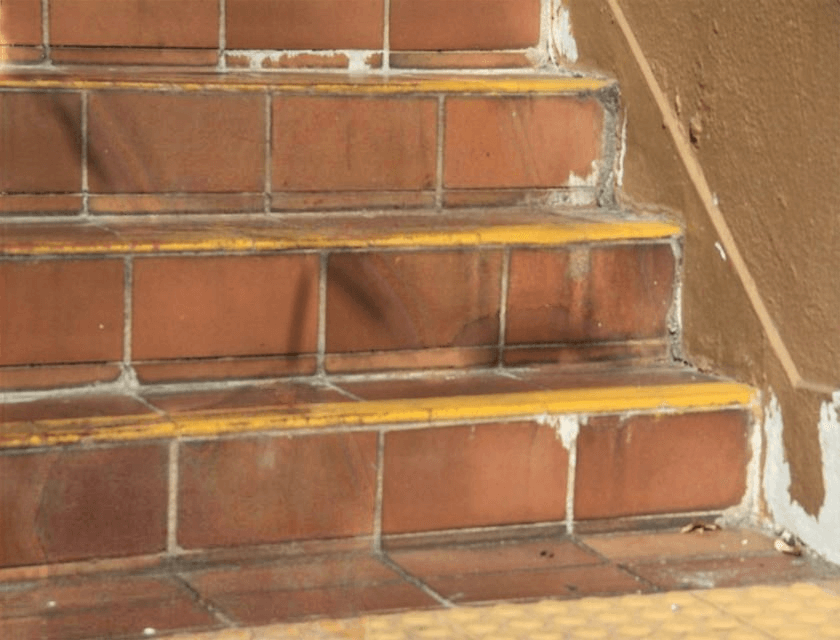}
    \includegraphics[width=1\linewidth,height=1\linewidth]{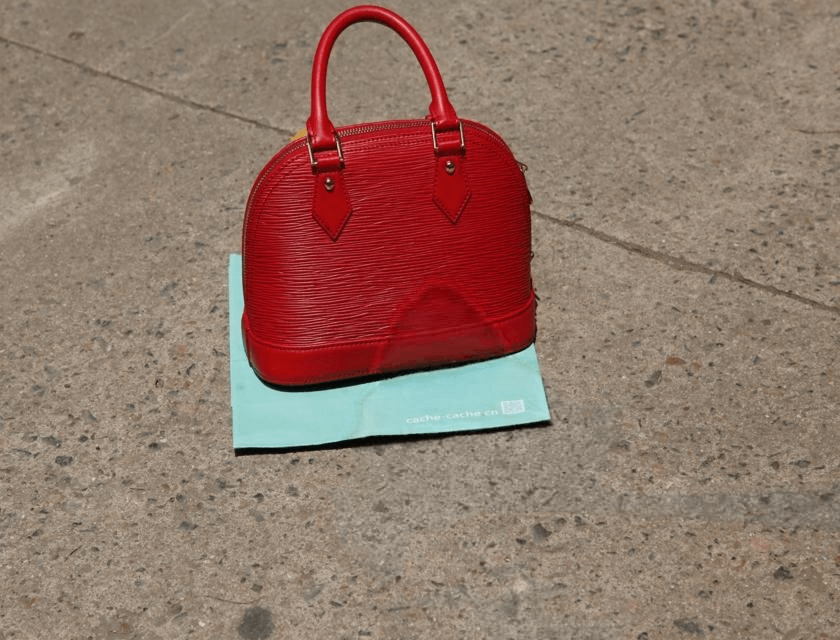}
    \includegraphics[width=1\linewidth,height=1\linewidth]{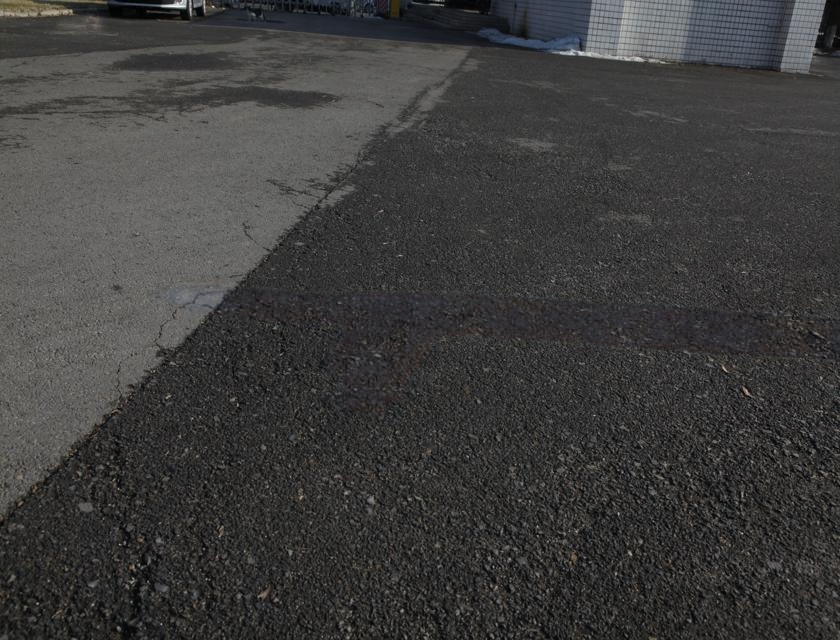}
    \includegraphics[width=1\linewidth,height=1\linewidth]{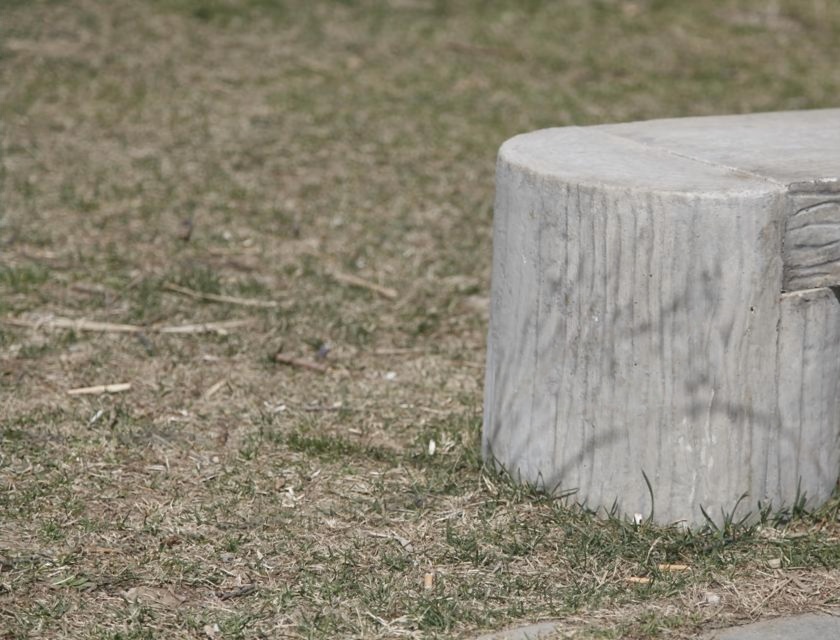}
    \includegraphics[width=1\linewidth,height=1\linewidth]{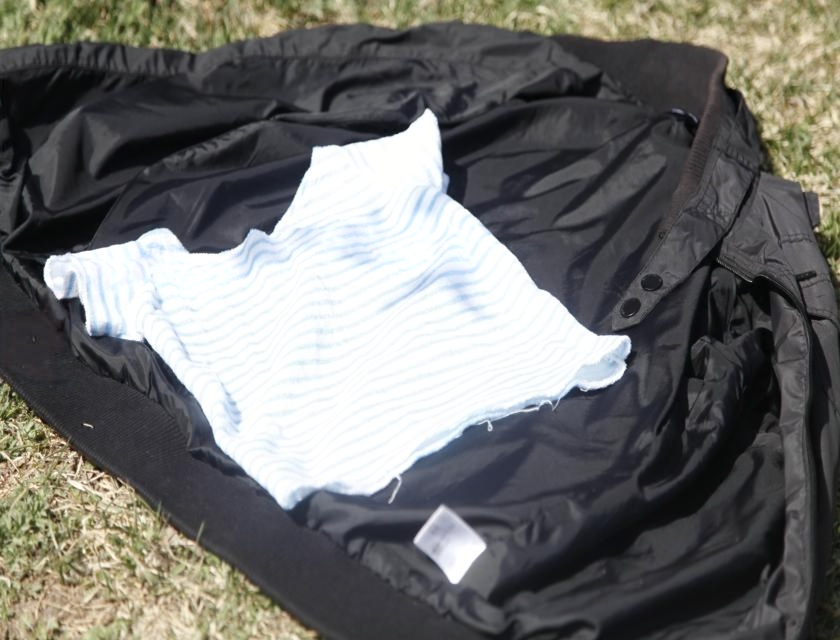}
    \includegraphics[width=1\linewidth,height=1\linewidth]{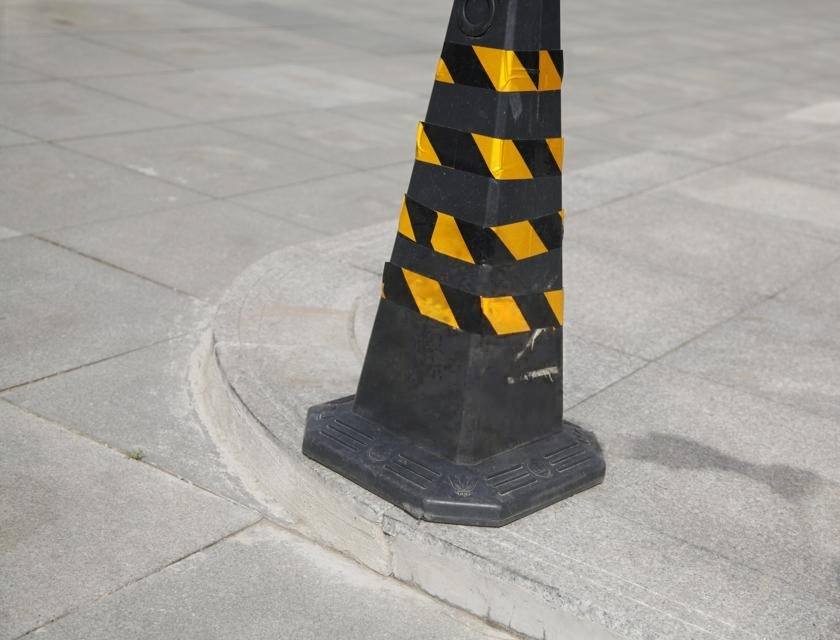}
    \caption{BMN~\cite{zhu_bijective_2022}}
  \end{subfigure}
  \begin{subfigure}[t]{.135\linewidth}
    \captionsetup{justification=centering, labelformat=empty, font=scriptsize}
    \includegraphics[width=1\linewidth,height=1\linewidth]{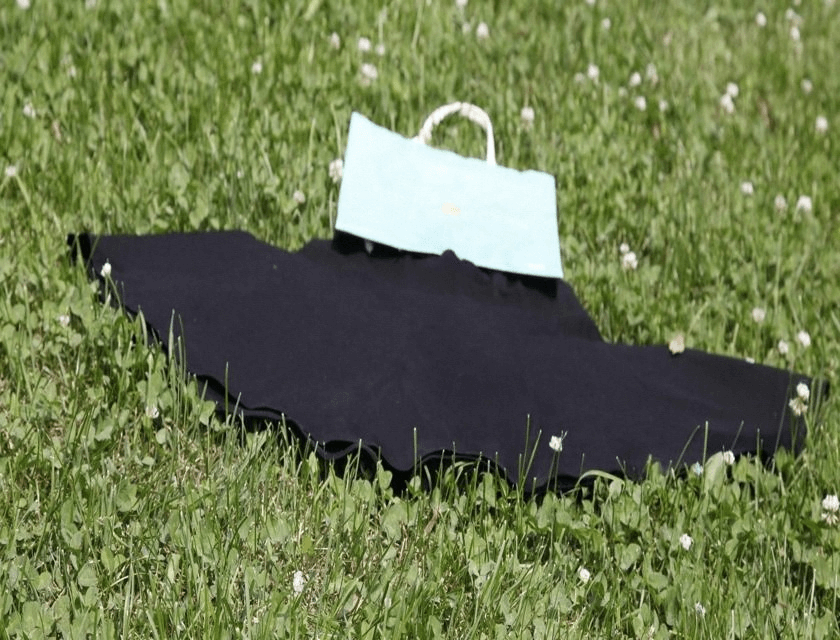}
    \includegraphics[width=1\linewidth,height=1\linewidth]{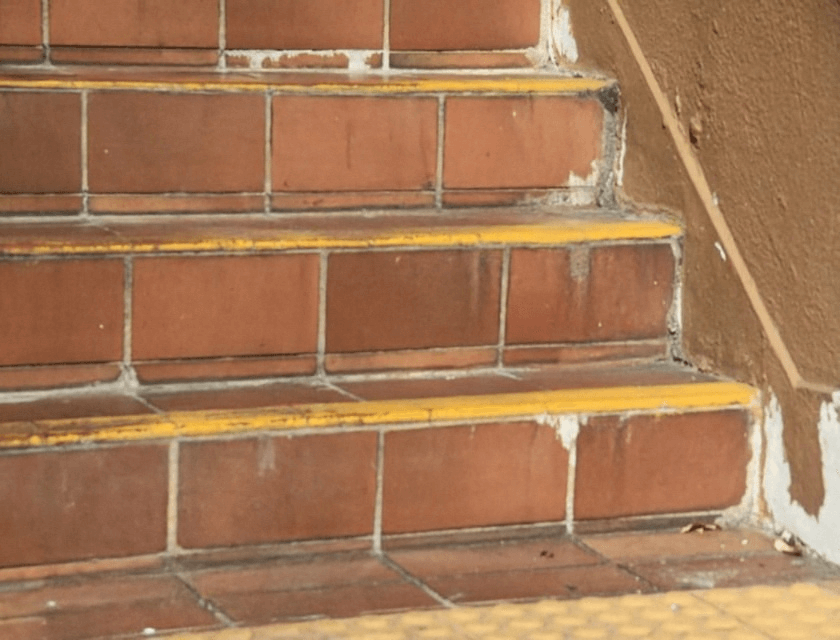}
    \includegraphics[width=1\linewidth,height=1\linewidth]{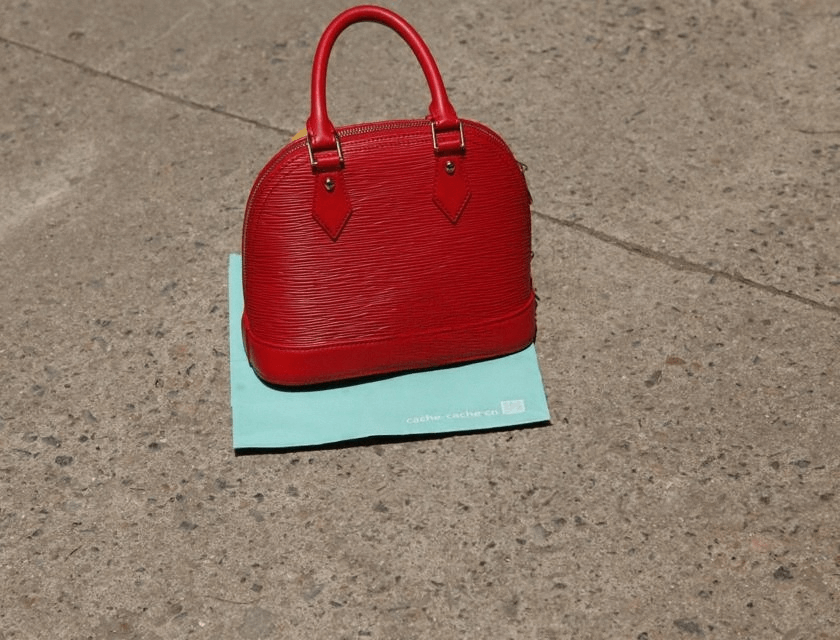}
    \includegraphics[width=1\linewidth,height=1\linewidth]{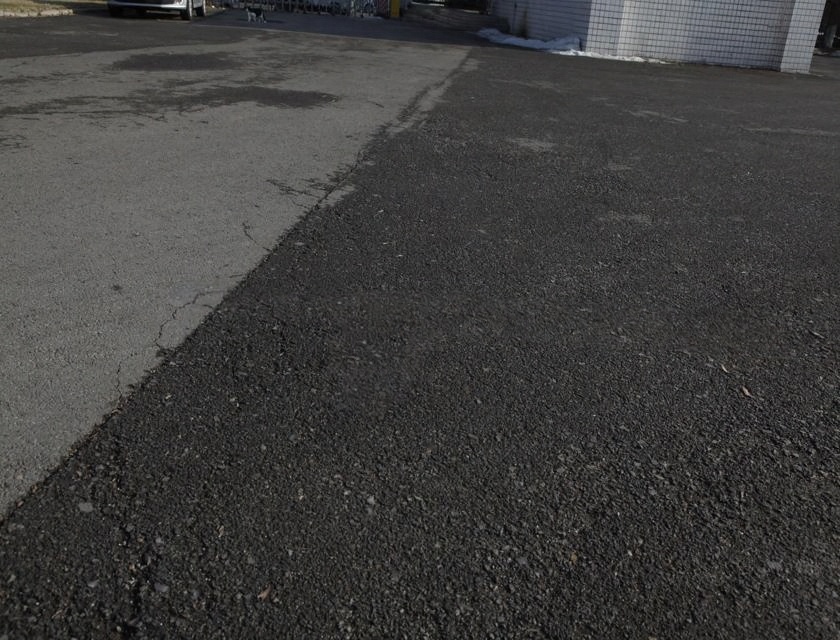}
    \includegraphics[width=1\linewidth,height=1\linewidth]{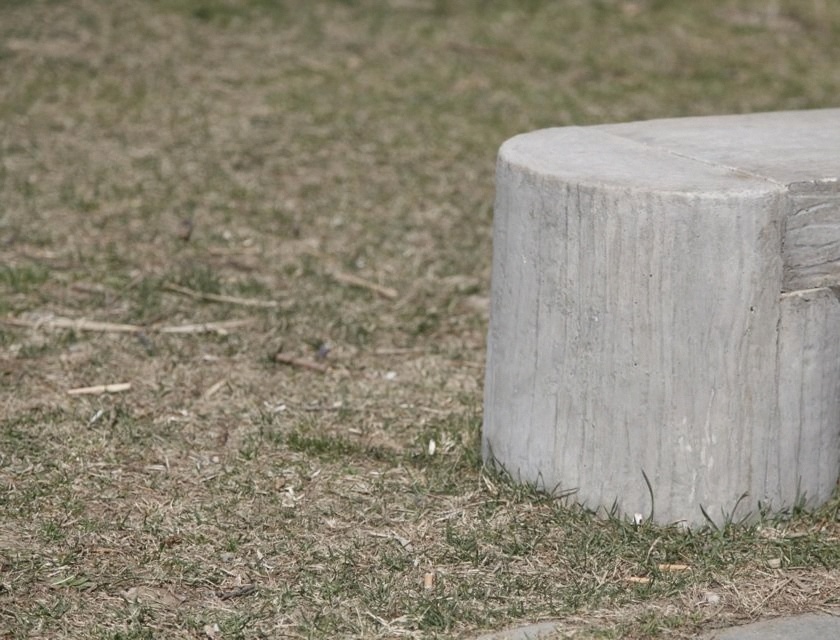}
    \includegraphics[width=1\linewidth,height=1\linewidth]{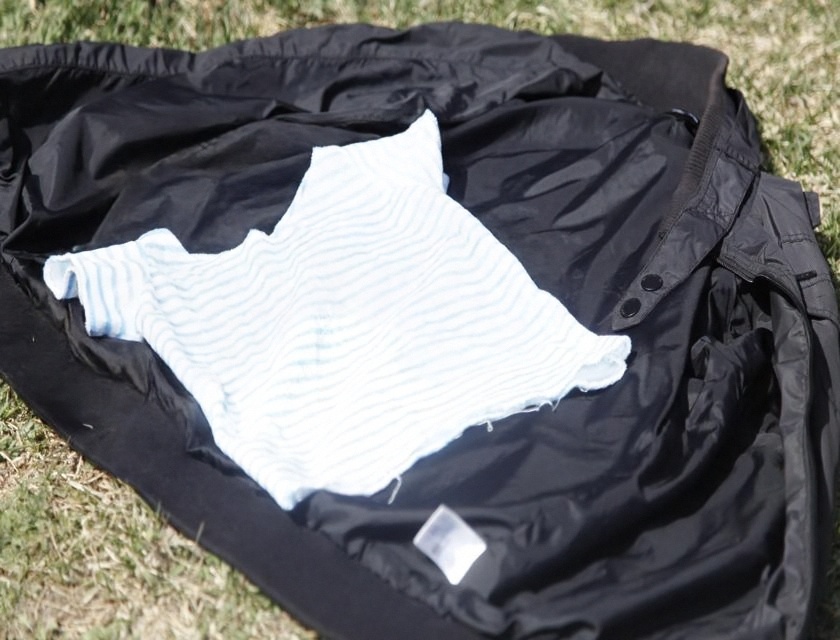}
    \includegraphics[width=1\linewidth,height=1\linewidth]{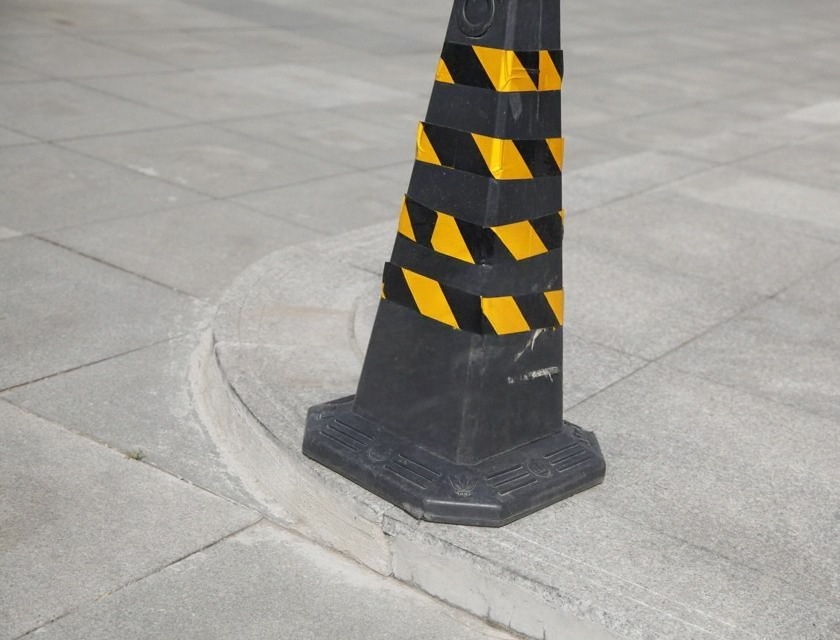}
    \caption{\tiny \ours~(Ours)}
  \end{subfigure}
  \begin{subfigure}[t]{.135\linewidth}
    \captionsetup{justification=centering, labelformat=empty, font=scriptsize}
    \includegraphics[width=1\linewidth,height=1\linewidth]{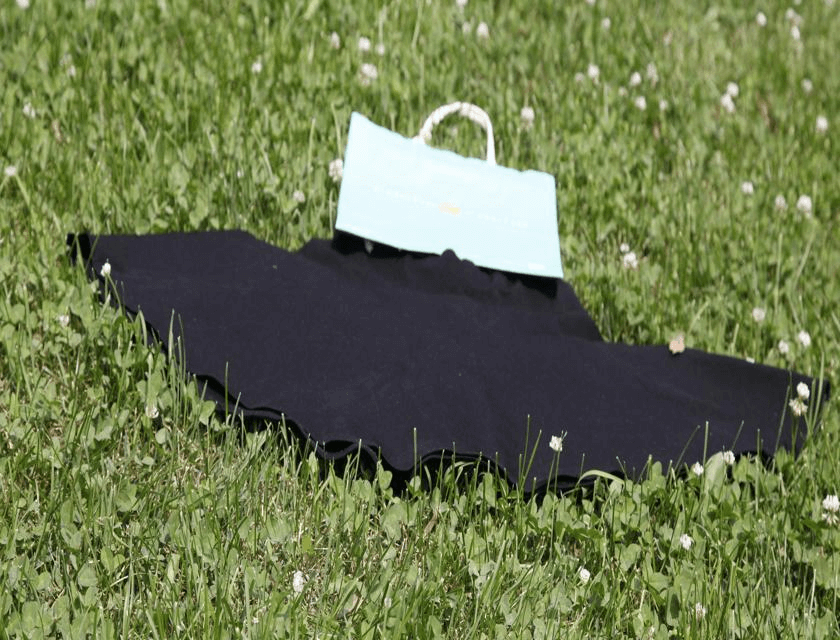}
    \includegraphics[width=1\linewidth,height=1\linewidth]{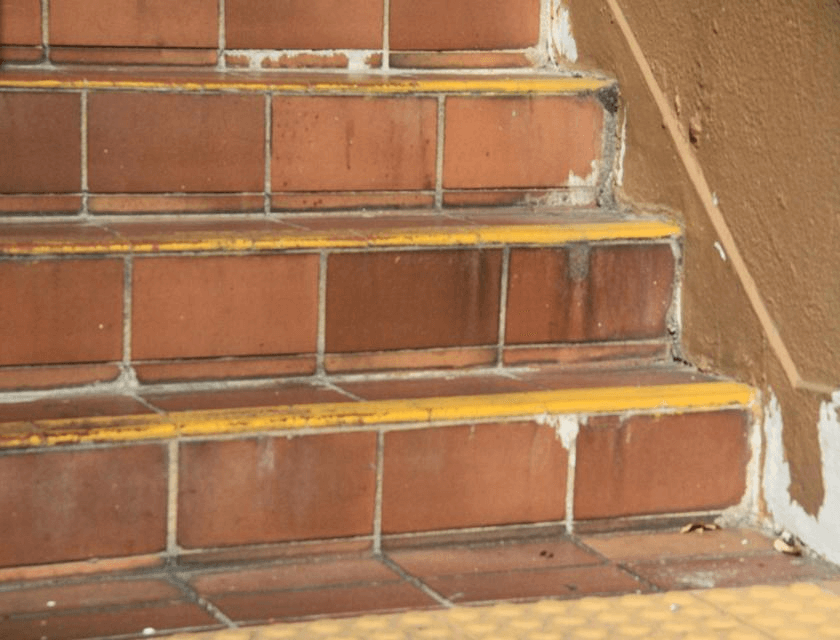}
    \includegraphics[width=1\linewidth,height=1\linewidth]{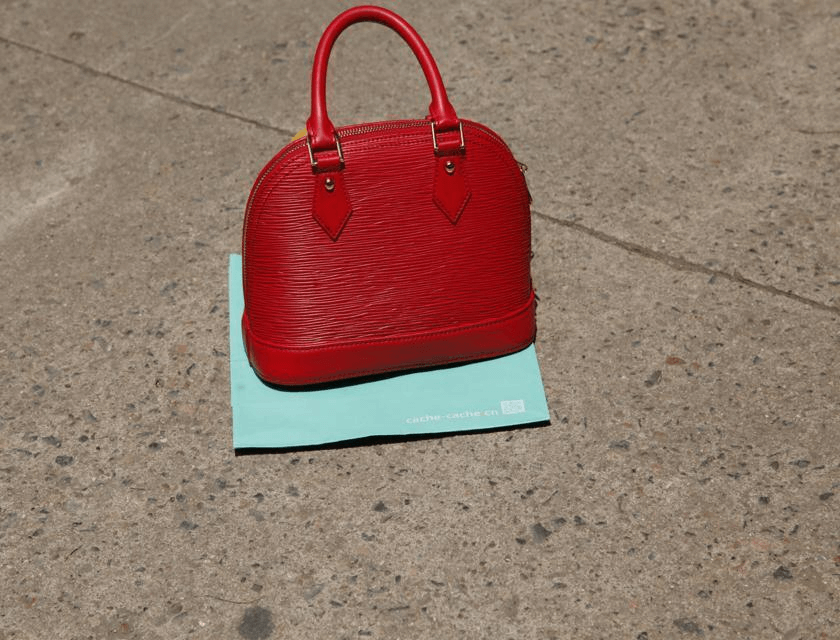}
    \includegraphics[width=1\linewidth,height=1\linewidth]{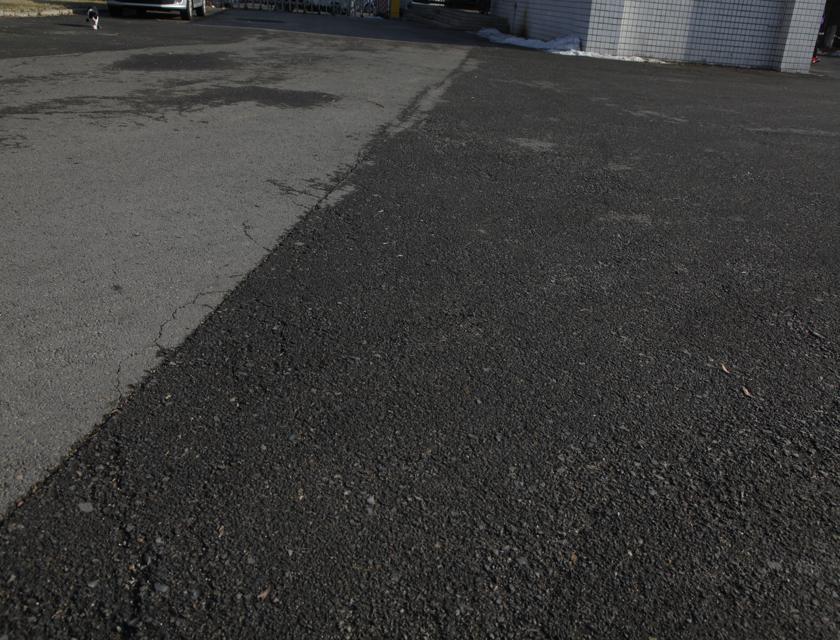}
    \includegraphics[width=1\linewidth,height=1\linewidth]{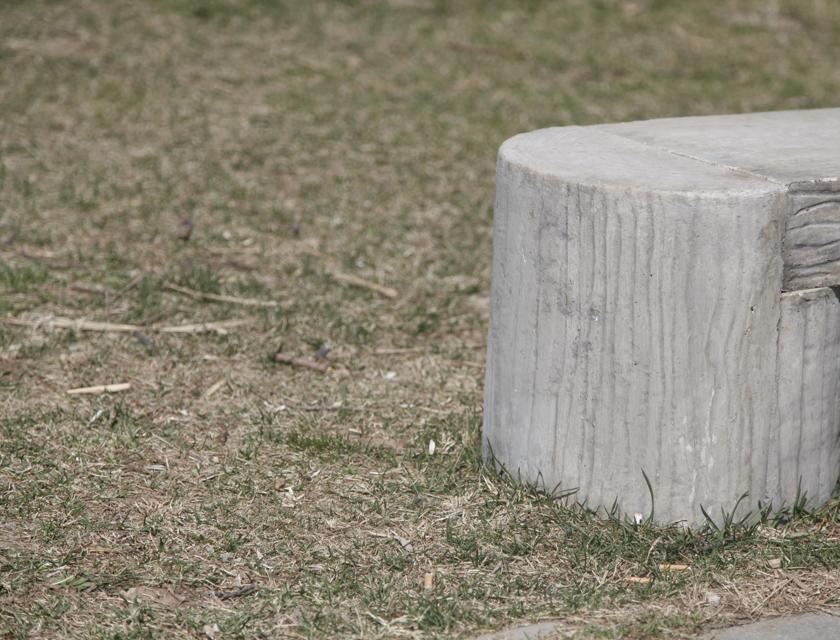}
    \includegraphics[width=1\linewidth,height=1\linewidth]{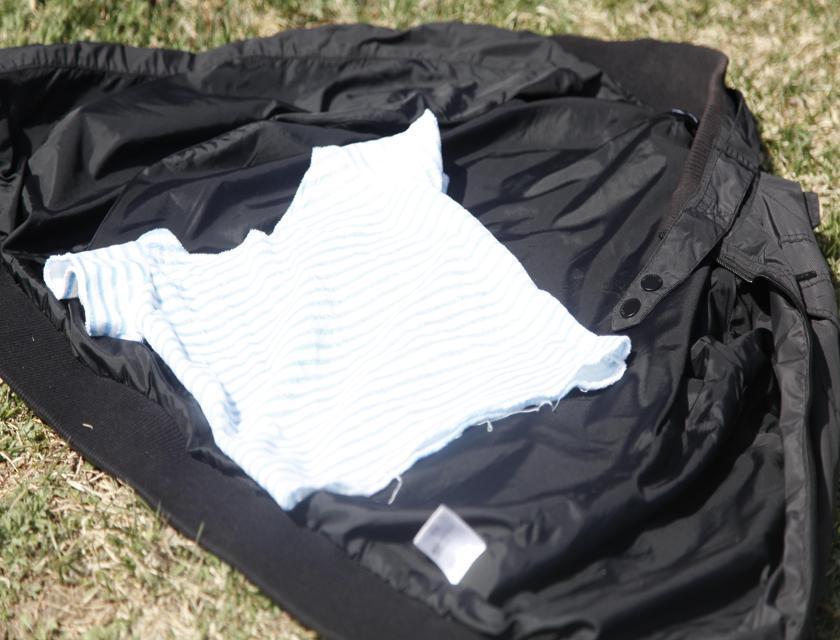}
    \includegraphics[width=1\linewidth,height=1\linewidth]{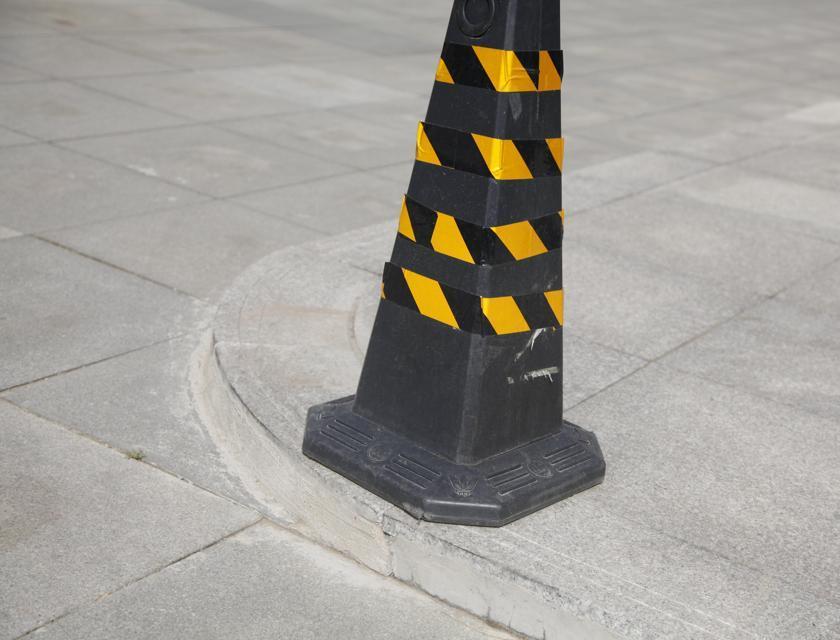}
    \caption{Ground Truth}
  \end{subfigure}
  \hfill
  \vspace{-.5\baselineskip}
  \caption{\textbf{Visual comparisons of representative hard shadow cases from the SRD dataset.}}
  \vspace{-1\baselineskip}
  \label{fig:srd}
\end{figure*}
\newpage

\section{Additional Visual Comparisons of Instance Shadow Removal}
Here we provide visual comparisons between our instance shadow removal results and the results of the most recent state-of-the-art shadow removal method, \ie, SG-ShadowNet~\cite{wan2022}.
While SG-ShadowNet can accept an instance shadow mask, it fails to preserve the details under the shadow and generate realistic results. 

\begin{figure*}[h]
\centering
  \begin{subfigure}[t]{.19\linewidth}
    \captionsetup{justification=centering, labelformat=empty, font=scriptsize}
    \includegraphics[width=1\linewidth]{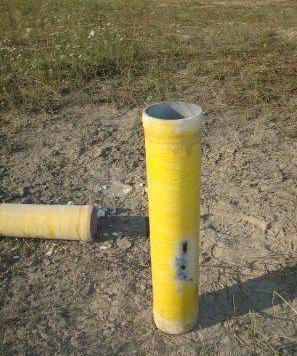}
    \includegraphics[width=1\linewidth]{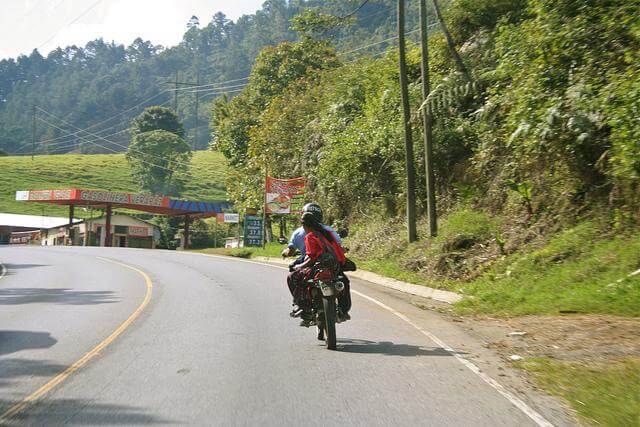}
    \includegraphics[width=1\linewidth]{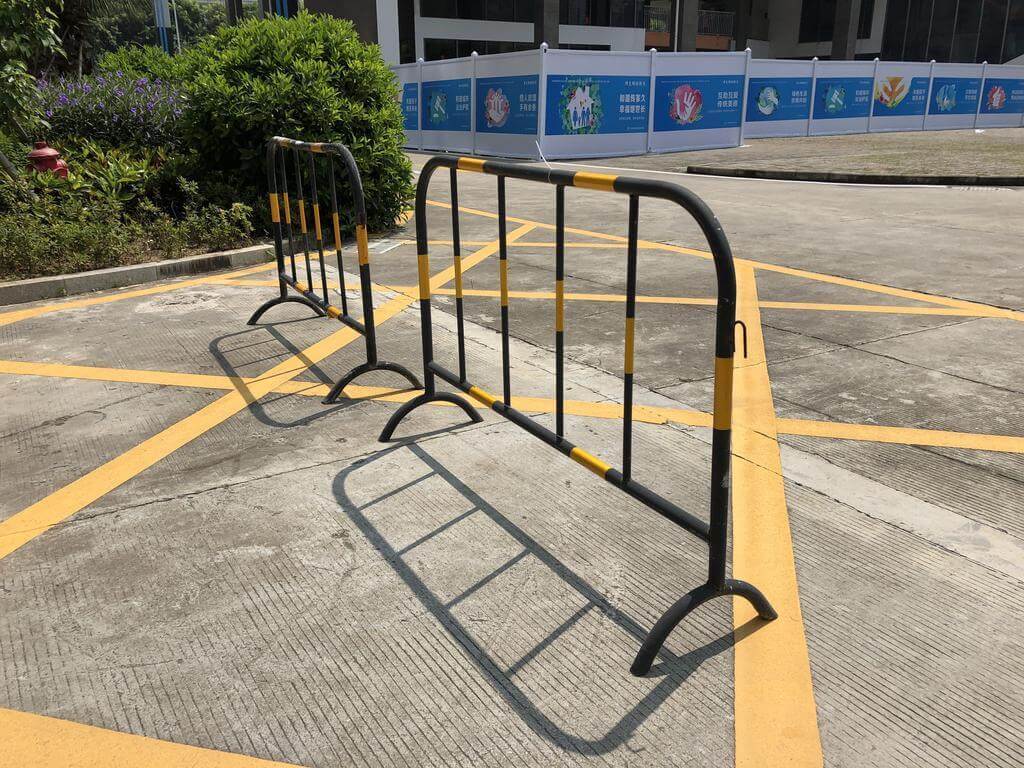}
    \includegraphics[width=1\linewidth]{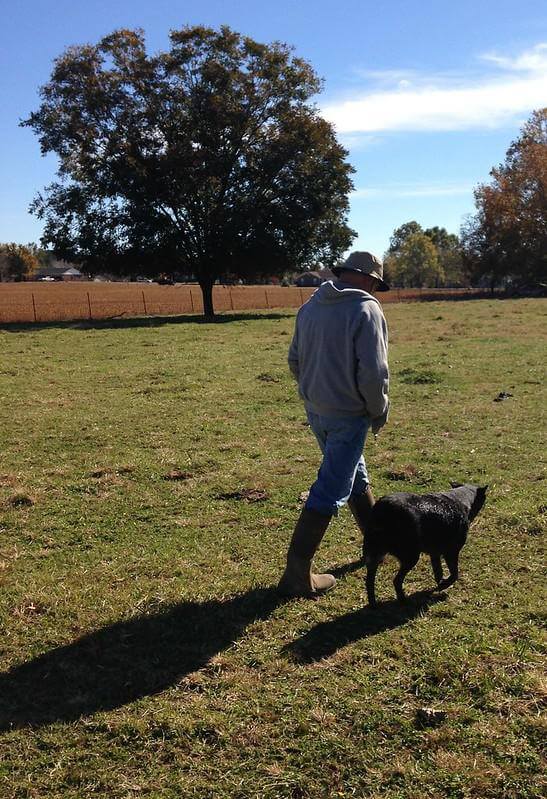}
    \caption{Shadows}
  \end{subfigure}
  \begin{subfigure}[t]{.19\linewidth}
    \captionsetup{justification=centering, labelformat=empty, font=scriptsize}
    \includegraphics[width=1\linewidth]{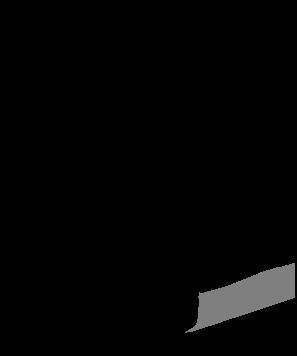}
    \includegraphics[width=1\linewidth]{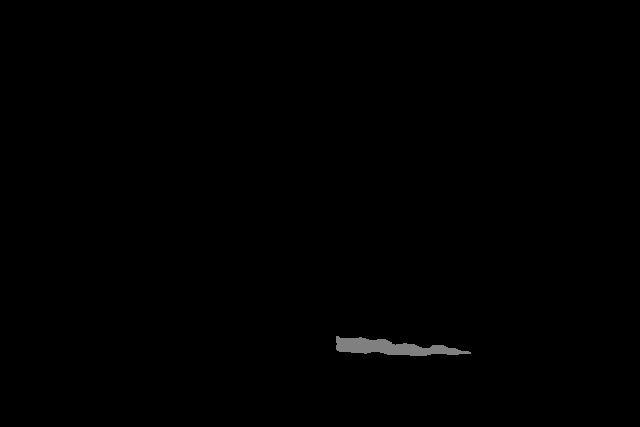}
    \includegraphics[width=1\linewidth]{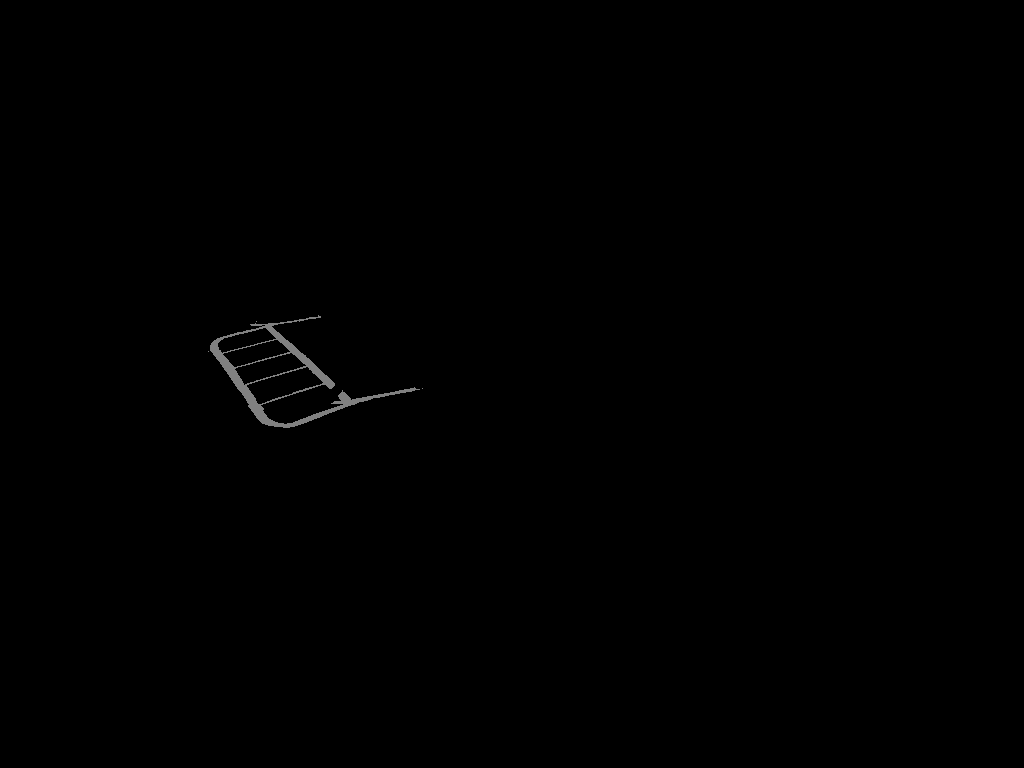}
    \includegraphics[width=1\linewidth]{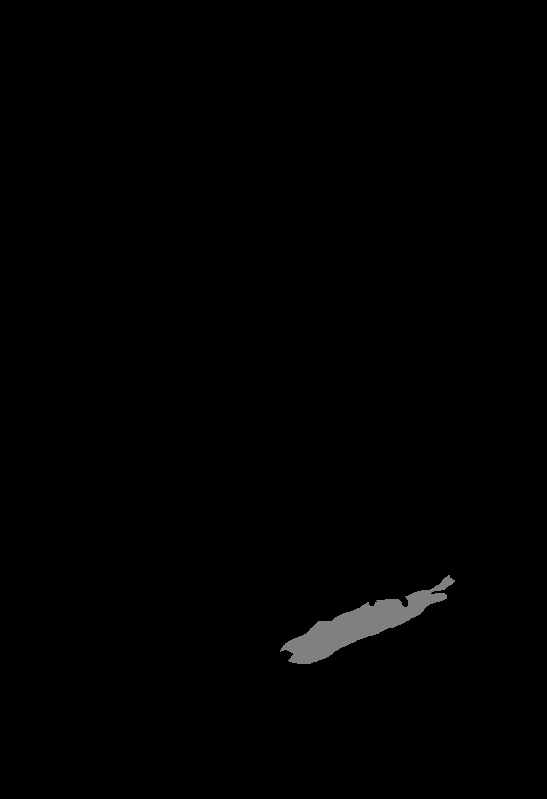}
    \caption{Shadow Mask}
  \end{subfigure}
  \begin{subfigure}[t]{.19\linewidth}
    \captionsetup{justification=centering, labelformat=empty, font=scriptsize}
    \includegraphics[width=1\linewidth]{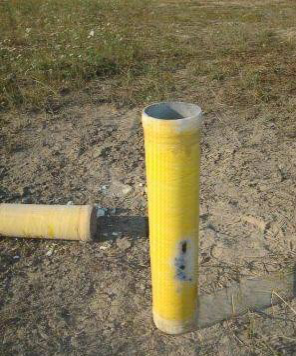}
    \includegraphics[width=1\linewidth]{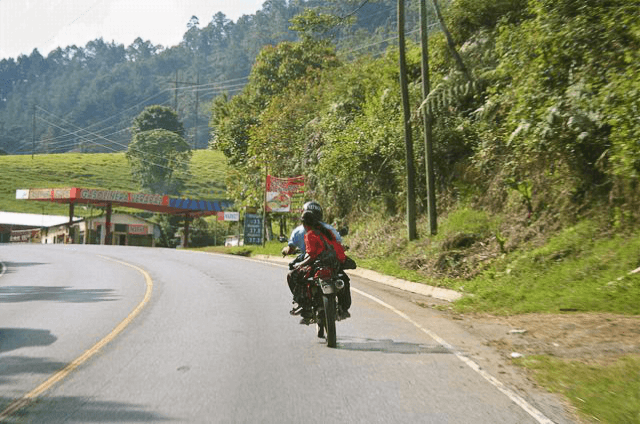}
    \includegraphics[width=1\linewidth]{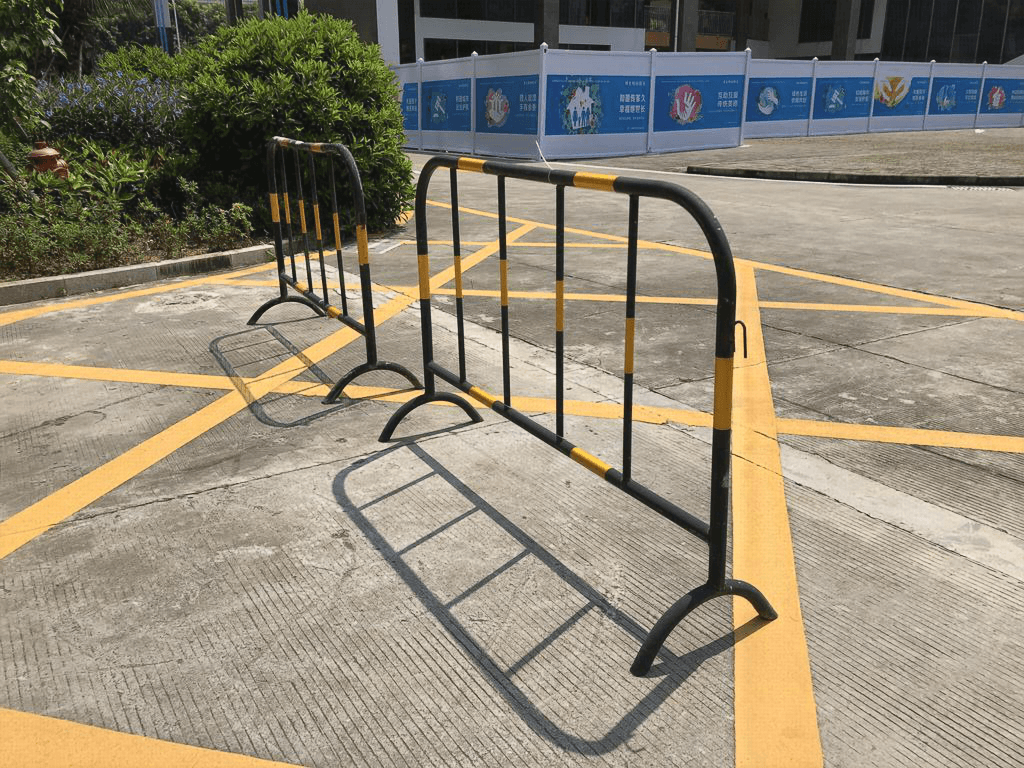}
    \includegraphics[width=1\linewidth]{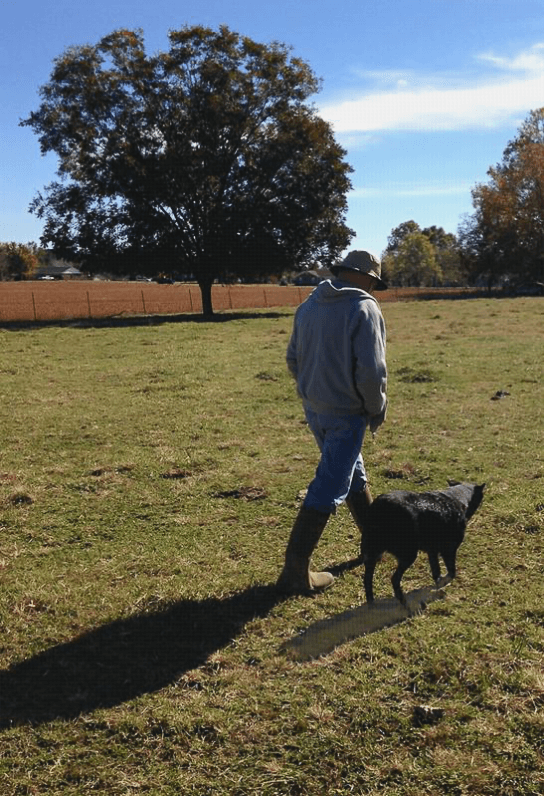}
    \caption{SG-ShadowNet~\cite{wan2022}}
  \end{subfigure}
  \begin{subfigure}[t]{.19\linewidth}
    \captionsetup{justification=centering, labelformat=empty, font=scriptsize}
    \includegraphics[width=1\linewidth]{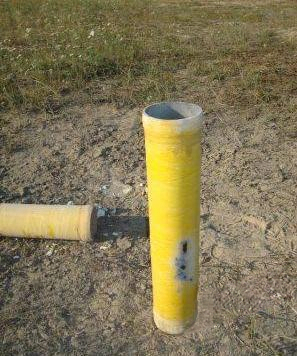}
    \includegraphics[width=1\linewidth]{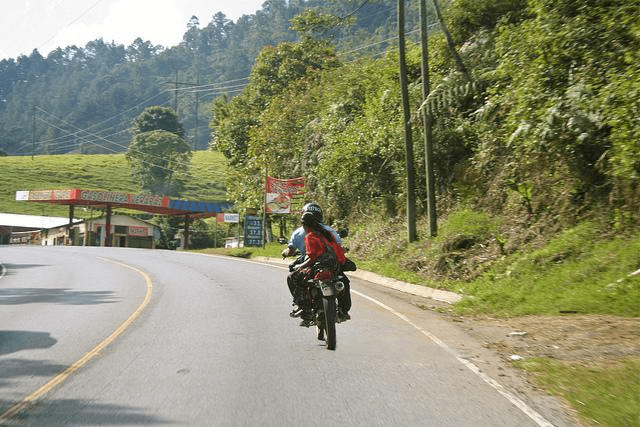}
    \includegraphics[width=1\linewidth]{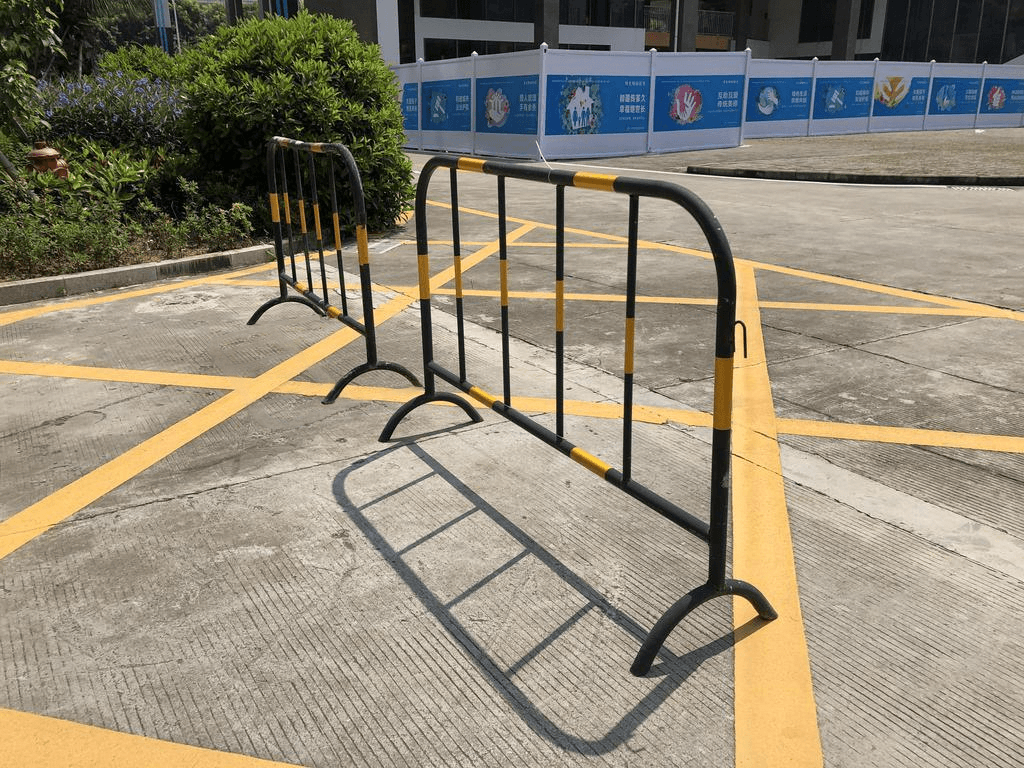}
    \includegraphics[width=1\linewidth]{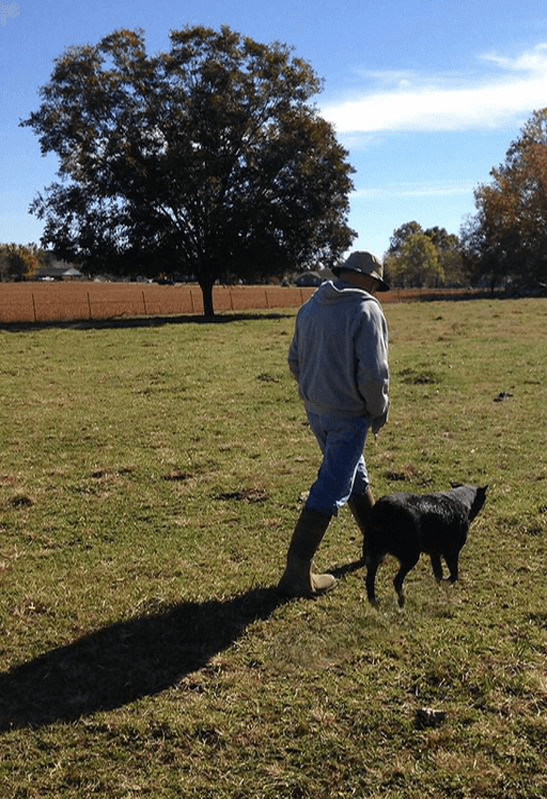}
    \caption{\ours~(Ours)}
  \end{subfigure}
  \begin{subfigure}[t]{.19\linewidth}
    \captionsetup{justification=centering, labelformat=empty, font=scriptsize}
    \includegraphics[width=1\linewidth]{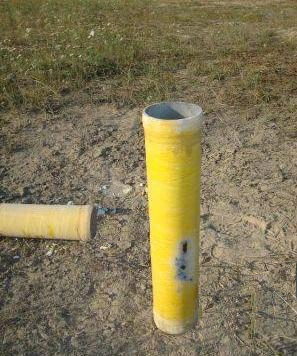}
    \includegraphics[width=1\linewidth]{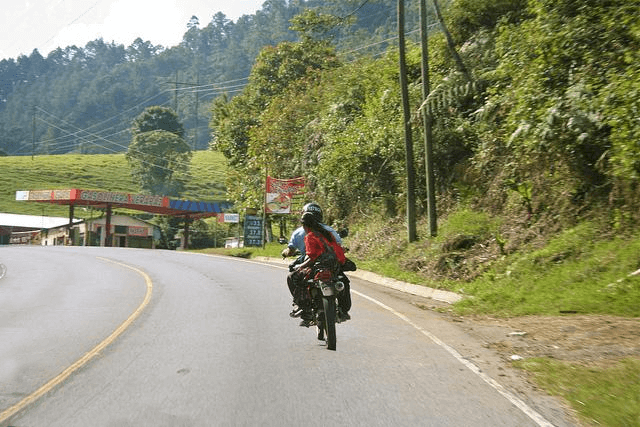}
    \includegraphics[width=1\linewidth]{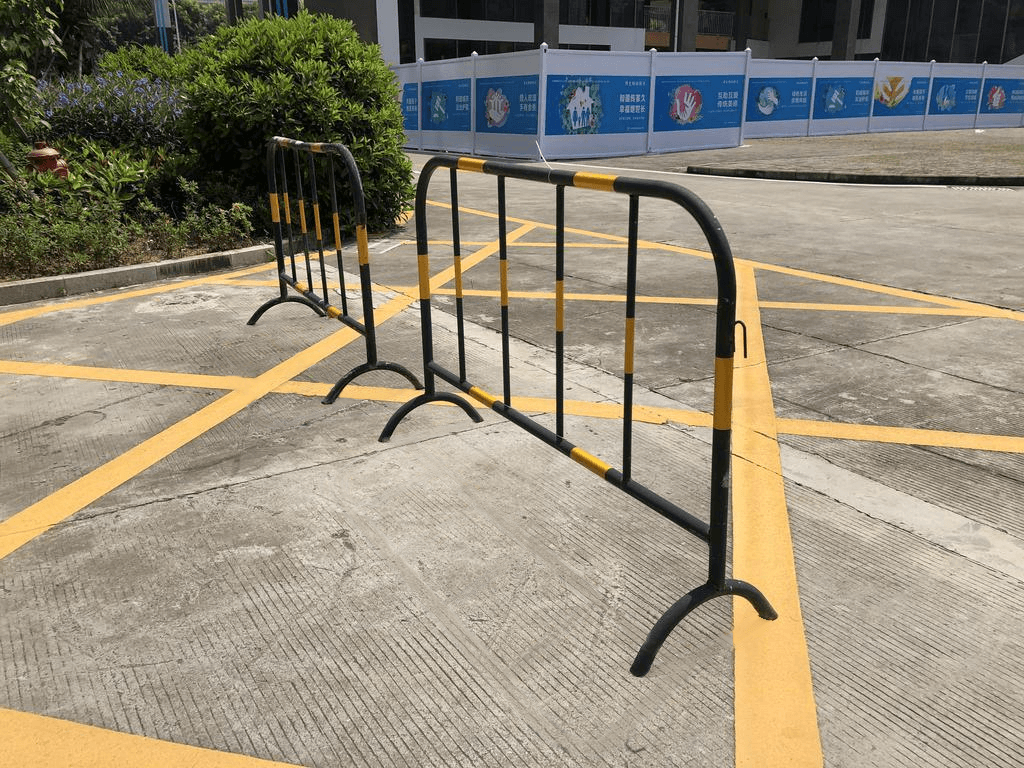}
    \includegraphics[width=1\linewidth]{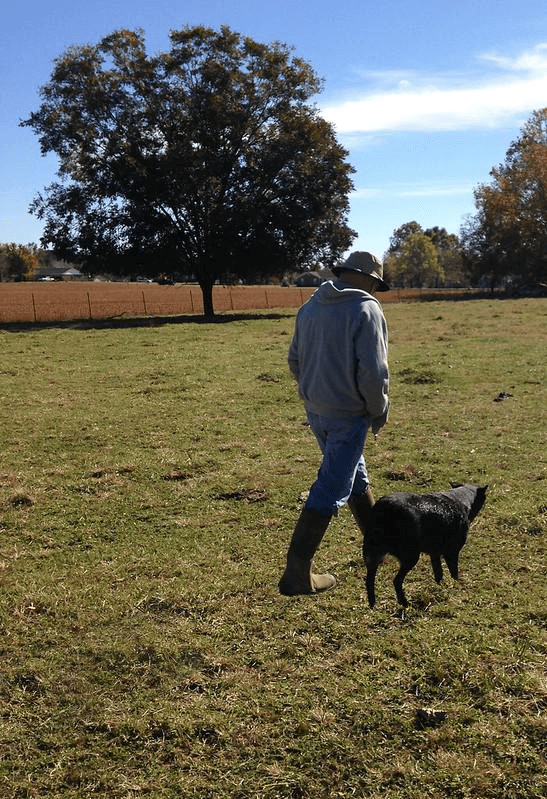}
    \caption{Reference}
  \end{subfigure}
  \hfill
  \vspace{-.5\baselineskip}
  \caption{\textbf{Visual comparisons of instance shadow removal cases.}}
  \vspace{-1\baselineskip}
  \label{fig:instancesrd}
\end{figure*}

\end{document}